\documentclass{article} 
\usepackage{iclr2027_conference,times}

\usepackage{amsmath,amsfonts,bm}

\def\eqref#1{equation~\ref{#1}}

\def\1{\bm{1}}

\DeclareMathAlphabet{\mathsfit}{\encodingdefault}{\sfdefault}{m}{sl}
\SetMathAlphabet{\mathsfit}{bold}{\encodingdefault}{\sfdefault}{bx}{n}

\newcommand{\E}{\mathbb{E}}

\newcommand{\KL}{D_{\mathrm{KL}}}

\usepackage{hyperref}
\usepackage{url}
\usepackage{tocloft}

\definecolor{sagegreen}{RGB}{235, 247, 240}
\newcommand{\green}{\cellcolor{sagegreen}}
\usepackage{times}
\usepackage{latexsym}
\usepackage[T1]{fontenc}
\usepackage[utf8]{inputenc}
\usepackage{microtype}
\usepackage{inconsolata}
\usepackage{subcaption}
\usepackage{graphicx}
\usepackage{algorithm}
\usepackage{algpseudocode}
\usepackage{amsmath,amssymb}
\usepackage{enumitem}
\usepackage{xspace}
\newcommand{\method}{{\textsc{OWLET}}\xspace}
\usepackage{amsmath}
\usepackage{bm}

\definecolor{amber}{HTML}{F39C12}

\definecolor{emerald}{HTML}{27AE60}

\usepackage{amsfonts}
\usepackage{multirow}
\usepackage{booktabs}
\usepackage{amsthm}

\usepackage{xcolor}
\usepackage{colortbl}
\definecolor{desc}{RGB}{99,178,238}
\definecolor{acc}{RGB}{118,218,145}
\definecolor{rej}{RGB}{248,149,136}
\definecolor{uclablue}{rgb}{0.15,0.45,0.68}
\definecolor{c0}{RGB}{0,120,212}
\definecolor{myblue}{RGB}{39, 116, 174}

\definecolor{uclablue}{rgb}{0.15, 0.45, 0.68}
\hypersetup{
    breaklinks,
    citecolor=uclablue,
    colorlinks=true,
    linkcolor=uclablue,
    pdftitle={OWLET: On-Policy Weighted Learning via Evidence Transfer for Video Large Language Model Post-Training},
    pdfauthor={Qitan Lv, Ke Fan, Hong Wang, Wen Wu, Yilin Wang, Bowen Zhou, Feng Wu, Chao Zhang}
}
\usepackage{wrapfig}
\usepackage{amssymb}

\title{
    \raisebox{-0.2\height}{\includegraphics[width=0.06\textwidth]{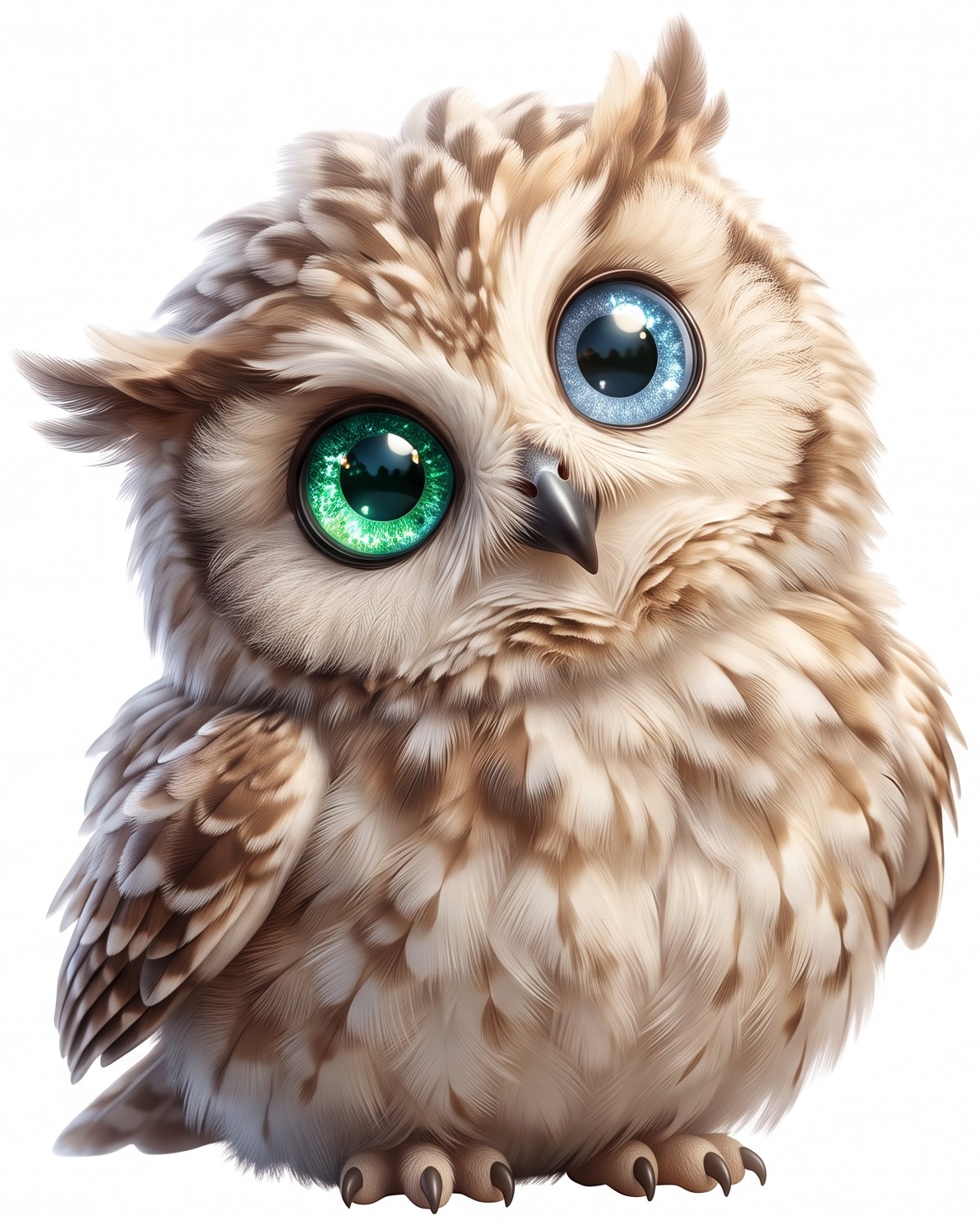}}\hspace{-0.25em} 
\method: \underline{O}n-Policy \underline{W}eighted \underline{L}earning via \underline{E}vidence \underline{T}ransfer for Video Large Language Model Post-Training}
\addtocontents{toc}{\protect\setcounter{tocdepth}{-1}}

\author{
    \textbf{Qitan Lv}$^{1,2}$\thanks{This work is done when Qitan Lv is an intern in Shanghai AI Laboratory.} \quad
    \textbf{Ke Fan}$^{2,3}$ \quad
    \textbf{Hong Wang}$^{1}$ \quad
    \textbf{Wen Wu}$^{2}$ \\
    \textbf{Yilin Wang}$^{2}$ \quad
    \textbf{Bowen Zhou}$^{1,2,3}$ \quad
    \textbf{Feng Wu}$^{1}$ \quad
    \textbf{Chao Zhang}$^{2,3}$\thanks{The corresponding author.} \\
    $^1$University of Science and Technology of China \quad
    $^2$Shanghai AI Laboratory \\
    $^3$Tsinghua University \\
    \texttt{\{qitanlv, wanghong1700\}@mail.ustc.edu.cn} \quad
    \texttt{fengwu@ustc.edu.cn} \\
    \texttt{\{fanke1, wuwen, wangyilin, zbw, zhangchao\}@pjlab.org.cn}
}

\definecolor{azure(colorwheel)}{HTML}{007FFF}

\iclrfinalcopy
\begin{document}

\maketitle
\fancyhead{}

\begin{abstract}
    Post-training has become a key stage for eliciting reasoning capabilities in Video Large Language Models (Video-LLMs). Existing post-training methods mainly follow two paradigms: reinforcement learning (RL) and on-policy distillation (OPD). However, they do not fully exploit visual evidence for effective training. RL supervises the policy with a single trajectory-level reward, which is too coarse-grained for video reasoning: the same response-level advantage is applied to every generated token, disregarding that the tokens of a video response are grounded in the visual stream to very different degrees. OPD provides token-level supervision through output-distribution matching, but lacks direct supervision of the spatiotemporal representations that produce those outputs. We observe across extensive experiments that the same model reasons more accurately when its video view combines global temporal context with question-relevant details. Comparing the same student-generated response under this view and uniform sampling reveals changes in token likelihoods and internal representations, providing a training signal for evidence use. Building on this observation, we propose \method, an evidence-privileged RL method that derives training signals from a richer teacher view. This view couples uniform global context with question-conditioned local windows, providing visual evidence without requiring a larger model. To better exploit this privileged visual evidence, \method introduces (i) \textbf{evidence-aware credit redistribution}, which compares token likelihoods across the two views and redistributes positive response-level advantages according to their relative evidence support; and (ii) \textbf{spatiotemporal state alignment}, which aligns the two views' hidden states on verifier-accepted responses. To facilitate training at scale, we release Video-Evidence-96K, containing 95,705 automatically curated question-conditioned temporal-evidence records over 21,832 videos. Experiments on the Qwen3.5 and Qwen3-VL families, ranging from 0.8B to 27B parameters, demonstrate the effectiveness of \method, yielding up to 10.9 point accuracy improvement over the vanilla model.
\end{abstract}

\section{Introduction}


Recent advances in video large language models (Video-LLMs)~\citep{qwen35, google2026gemini35, openai2025gpt5, openai2025gpt52} have extended large language models to the video domain, achieving strong downstream performance~\citep{videosalmonn2, zhang2023temporal}. However, answering a video question often requires more than recognizing what a clip contains. The evidence that determines the answer may be concentrated in a few key frames scattered among thousands~\citep{videoitg}, or span the video as a whole, requiring an understanding of event order, duration, and changes over time~\citep{xiao2021next}. The central post-training question is how to make the model use visual input more effectively.

To this end, two on-policy paradigms have been widely adopted for reasoning-oriented post-training~\citep{song2026survey}. Reinforcement learning (RL) samples responses from the current policy and reinforces those accepted by a verifier, keeping optimization on the distribution the model uses at test time~\citep{ shao2024deepseekmath}. On-policy distillation (OPD) preserves this sampling scheme but replaces the scalar reward with a teacher that re-scores the student's  responses~\citep{videoopd}. 
However, these paradigms do not fully exploit visual evidence for effective training, leaving two aspects of visual supervision underexplored: \textbf{(i)} RL supervises the policy with a single trajectory-level reward, which is too coarse-grained for video reasoning. The same response-level advantage is applied to every generated token, disregarding that the tokens are grounded in the visual stream to very different degrees. \textbf{(ii)} OPD provides token-level supervision through output-distribution matching, but lacks direct supervision of the spatiotemporal representations that produce those outputs. This omission is particularly vital for video reasoning, where cues from distant frames need to be integrated into an internal account of the  temporal structure: matching the output distributions alone does not ensure that the student acquires representations formed by the teacher (please refer to Section~\ref{sec:motivation} for details).

\begin{figure}[t]
    \centering
    \includegraphics[width=\linewidth]{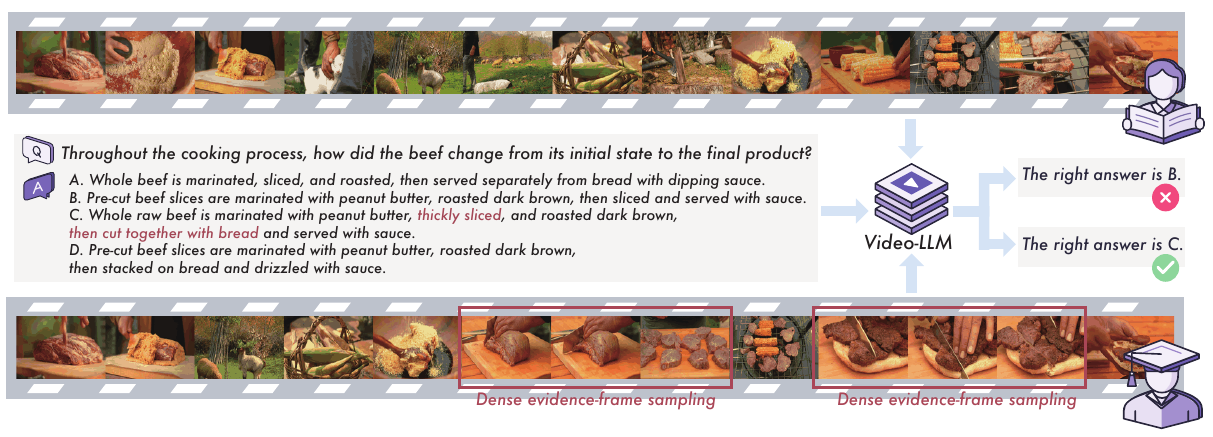}
    \caption{Illustration of how \method improves video reasoning through evidence-guided post-training. Uniform frame sampling misses the temporal transitions needed to answer the question, whereas \method learns from a correct evidence-privileged teacher.}
    \label{fig:framework}
\end{figure}

Addressing these limitations calls for supervision that reveals both which tokens benefit from visual evidence and how that evidence is organized into spatiotemporal representations. We therefore construct an evidence-privileged \textit{teacher view} that couples a uniform global context with question-conditioned local windows, preserving the video's overall structure while resolving relevant moments in greater detail as shown in Figure \ref{fig:framework}. Our motivated experiments in Section~\ref{sec:motivation} show that the same model reasons more accurately when it combines global temporal context with question-relevant details. Comparing the same student-generated response under this richer view and uniform sampling reveals changes in visual token support and internal representations. These differences arise from the contribution of visual evidence and provide supervision without requiring a larger model.

Building on this observation, we propose \method, an evidence-privileged on-policy RL framework to enhance video-LLM reasoning capabilities. The student samples responses from uniformly sampled frames, and the teacher provides supervision through two complementary components. \textbf{(i) Evidence-aware credit redistribution} compares token likelihoods under the two views and redistributes positive response-level advantages into token-level advantages according to their relative evidence support. The verifier thus determines the direction of learning, while the teacher view guides the allocation of positive credit within each response. \textbf{(ii) Spatiotemporal state alignment} transfers the spatiotemporal representations formed under the teacher view and aligns the response-end hidden states of the two views on verifier-accepted responses. Both components operate on responses sampled by the student itself, keeping supervision tied to its own reasoning contexts.
To facilitate training at scale, we release \textbf{Video-Evidence-96K}, containing 95,705 automatically curated question-conditioned temporal-evidence records over 21,832 videos. Our contributions are as follows:

\begin{itemize}[itemsep=0pt, leftmargin=*]

\item \textbf{A large-scale temporal-evidence resource.} We release Video-Evidence-96K, 95,705 automatically curated question-conditioned temporal-evidence records over 21,832 videos, which provides the global tracks and question-conditioned windows needed to construct the teacher view at scale.

\item \textbf{Evidence-privileged on-policy learning.} We propose \method, which turns a temporally richer teacher view into fine-grained supervision for video-LLM post-training, through evidence-aware credit redistribution at the token level and spatiotemporal state alignment at the response level.

\item \textbf{Consistent and versatile gains.} Extensive experiments demonstrate the effectiveness of \method, yielding up to 10.9 point accuracy improvement over the vanilla model.

\end{itemize}


\section{Preliminary}
\label{sec:pre}

\paragraph{Video-LLM Post-Training.}
A video-LLM answers a question $q$ about a video $\mathcal{V}$. Since $\mathcal{V}$ contains more frames than the model can process, it is reduced to a \textit{view}: a sequence of frames that is encoded into visual tokens and prepended to $q$. We use $v$ for the view the student receives and $\tilde{v}$ for the teacher. Conditioned on a view, the policy $\pi_\theta$ generates a response $y=(y_1,\dots,y_{|y|})$ autoregressively.

\paragraph{Reinforcement Learning with Verifiable Rewards.}
RL has become the dominant route to reasoning ability, from preference optimization with policy-gradient algorithms~\citep{ouyang2022training, schulman2017proximal} to verifiable rewards, where a rule checks the final answer and no learned critic is required~\citep{shao2024deepseekmath}. For each question, a group of $G$ responses $\{y^i\}_{i=1}^{G}$ is sampled from the current policy under the student view, each receives a scalar reward $R(q,y^i)\in\{0,1\}$ from the verifier, and the advantage is estimated from the relative rewards within the group:
\begin{equation}
\label{eq:rl}
\mathcal{J}_{\mathrm{RL}}(\theta)
= \E_{\substack{q\sim P(Q),\\ \{y^i\}\sim\pi_{\theta_{\mathrm{old}}}(\cdot\mid q,v)}}
\left[
\frac{1}{G}\sum_{i=1}^{G}\frac{1}{|y^i|}\sum_{t=1}^{|y^i|}
\min\!\left(\rho^i_t(\theta)\,A^i,\ \mathrm{clip}\!\left(\rho^i_t(\theta),1-\epsilon,1+\epsilon\right)A^i\right)
\right],
\end{equation}
where $\rho^i_t(\theta)=\pi_\theta(y^i_t\mid q,v,y^i_{<t})\,/\,\pi_{\theta_{\mathrm{old}}}(y^i_t\mid q,v,y^i_{<t})$ is the importance ratio and $A^i=\bigl(R(q,y^i)-\mathrm{mean}(\{R(q,y^j)\})\bigr)/\mathrm{std}(\{R(q,y^j)\})$. Since sampling follows the evolving policy, training stays on the distribution the model uses at test time. However, the supervision is a single scalar per trajectory: every token is reinforced by the same reward regardless of how much it depends on the video.

\paragraph{On-Policy Self-Distillation.}
On-policy distillation replaces the scalar reward with a teacher that re-scores the student's own responses~\citep{agarwal2024policy, lu2025onpolicydistillation}. OPSD removes the external teacher: the two policies are initialized from the same weights, and what separates them is privileged information withheld from the student, so that any advantage the teacher holds cannot be attributed to extra capacity~\citep{hubotter2026reinforcement, shenfeld2026self, zhang2026opsdlonpolicyselfdistillationlongcontext}. The response is still sampled from the student, so the supervision is applied exactly on the states the student visits, but the objective becomes a token-level divergence to the teacher rather than a reward:
\begin{equation}
\label{eq:opsd}
\mathcal{L}_{\mathrm{OPSD}}(\theta)
= \E_{\substack{q\sim P(Q),\\ y\sim\pi_{\mathrm{stu}}(\cdot\mid q,v)}}
\left[
\frac{1}{|y|}\sum_{t=1}^{|y|}
\KL\!\left(
\pi_{\mathrm{stu}}(\cdot\mid q,v,y_{<t})
\,\Vert\,
\pi_{\mathrm{tea}}(\cdot\mid {q},\tilde{v},y_{<t})
\right)
\right].
\end{equation}
 This yields a dense signal at every position, yet it remains an output-space objective: Equation~\ref{eq:opsd} constrains only the next-token distributions, and weights all positions equally through the uniform factor $1/|y|$. Section~\ref{sec:motivation-gaps} shows that both properties are consequential for video, and \method addresses them by reshaping the per-token weight in Equation~\ref{eq:rl} and by adding a term that acts upstream of the distributions compared in Equation~\ref{eq:opsd}.

\section{Motivated Experiments}
\label{sec:motivation}

\subsection{Constructing the Evidence-Privileged Teacher View}
\label{sec:motivation-teacher-view}

\begin{figure*}[t]
    \centering
    \includegraphics[width=\linewidth]{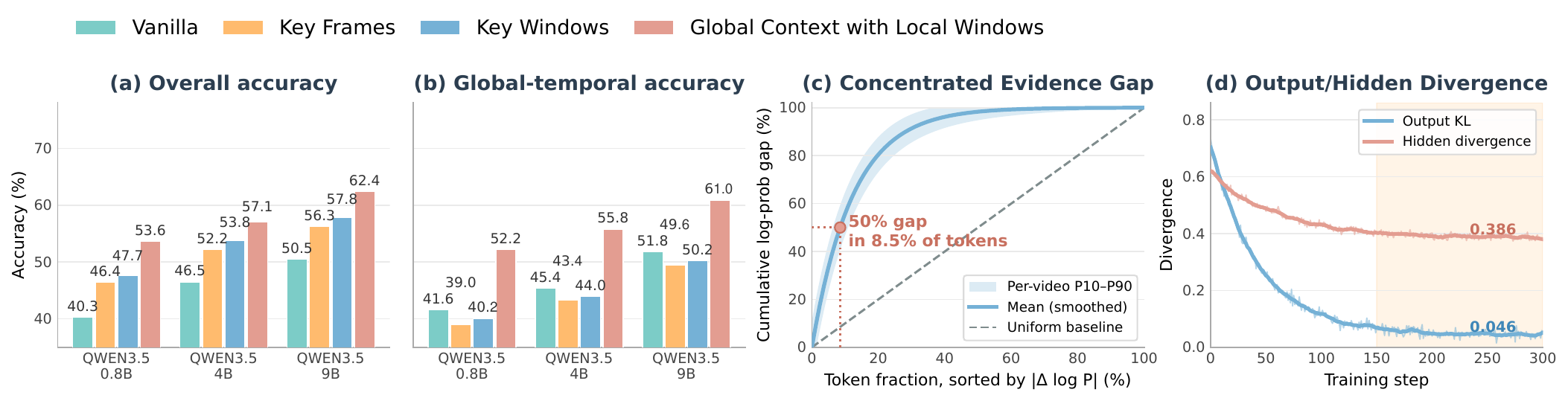}
    \caption{\textbf{(a, b)} Evidence-only views improve overall accuracy but lose global-temporal accuracy; only pairing global context with local windows improves both. \textbf{(c)} Half of the teacher--student log-probability gap concentrates on 8.5\% of response tokens, so spreading training credit uniformly dilutes the evidence signal. \textbf{(d)} Output-only distillation collapses the output KL, yet the hidden divergence plateaus: output supervision alone does not effectively align spatiotemporal representations.}
    \label{fig:motivation-combined}
\end{figure*}

Our design of \method is motivated by a simple question: \textit{how should a visually privileged view be constructed so that it actually improves reasoning?} A naive answer is a view assembled purely from key frames. However, as shown in Figure~\ref{fig:motivation-combined}, a view that contains only the annotated key frames may focus better on the local event that answers the question, but it also loses the surrounding context needed to understand the video as a whole. We therefore compare four input views on $1{,}000$ held-out samples from Video-Evidence-96K, each assembled from exactly $64$ frames so that the views differ only in which moments are shown: the \textit{vanilla} samples $64$ frames uniformly over the whole video; \textit{key frames} takes the $64$ frames ranked highest by question relevance; \textit{key windows} takes the $16$ highest-ranked frames and expands each into a disjoint four-frame window; and \textit{global context with local windows} combines $32$ uniform frames with disjoint four-frame windows around the $8$ highest-ranked frames. We evaluate both overall accuracy and a subset of \textit{global-temporal questions}, which require information spread across the video. These are questions about the overall purpose of a video, or about the causal relation between events (e.g., how a state evolves from beginning to end). Detailed categorization and composition  of Video-Evidence-96K are in Appendix~\ref{app:video-evidence-96k}.

Figure~\ref{fig:motivation-combined}(a,b) illustrates the trend across Qwen3.5-0.8B, 4B, and 9B. Using only key frames or windows improves the overall score over the vanilla, suggesting that focused visual evidence is generally beneficial. However, these focused views perform worse on global-temporal questions. The global context with local windows view performs best on both metrics: global frames preserve the broad temporal structure, while local windows provide details around question-relevant moments.

\subsection{Supervision Gaps Unique to Video Post-Training}
\label{sec:motivation-gaps}

Constructing the teacher view settles what the teacher should see, and we further need to design how the resulting advantage is transferred to the student. Since the teacher and the student are initialized with the same weight, differences in token log-probabilities for the same response under the two views reflect differences in visual input. Using Qwen3.5-9B with a 64-frame student view as an example, we identify two supervision gaps that existing methods for video reasoning leave underexplored. Results for other models and frame budgets are in Appendix~\ref{app:motivation-generalization}.

\paragraph{(i) Token credit is uniformly distributed, yet video grounding is not.}
Compared with text-only reasoning, video reasoning can exhibit greater variation in how strongly response tokens depend on input evidence. Tokens naming a temporal order, an event location, or a cross-frame causal relation often depend strongly on visual evidence, while connectives and narrative scaffolding tend to depend less on it~\citep{liu2026visualadvantageonpolicydistillationvisionlanguage}.  Figure~\ref{fig:motivation-combined}(c) quantifies the resulting structure. For student responses collected on $1{,}000$ held-out samples from Video-Evidence-96K, we re-score each one under the teacher view and rank its tokens by $|\Delta\log P|$ between the two views. The two passes share the same question, and the same generated prefix, and differ only in the frames shown; a large gap at a token therefore signals that predicting it relies on the additional visual evidence. Half of the total shift falls on \textbf{8.5\%} of the tokens, far from the uniform baseline, and the 10th--90th percentile band shows the concentration is consistent across videos. Existing post-training does not effectively exploit this gap: RL copies one scalar reward to every token, and OPD matches the two output distributions position by position without asking which positions the video decides. Both dilute supervision exactly where evidence matters most. This motivates \textbf{evidence-aware credit redistribution}, which reads the per-token gap as a diagnostic of evidence-dependence and redistributes training credit accordingly.

\paragraph{(ii) Output-level alignment leaves spatiotemporal representations unsupervised.}
Before its first answer token, a video-LLM must compress thousands of frames into hidden states carrying its internal account of the video's temporal structure. That integration happens during visual encoding and cross-modal attention, upstream of any token generation and therefore beyond the reach of an objective defined on the token distribution. Figure~\ref{fig:motivation-combined}(d) confirms the gap: once on-policy distillation converges, the output KL reduces to $0.046$, yet the response-end hidden states of the two views stay far apart, with $1-\mathrm{CKA}$~\citep{kornblith2019similarity}, a divergence index invariant to rotation and scaling of the representations, plateauing at $0.386$ (please refer to Appendix~\ref{app:ssa-protocol} for details). The student produces similar tokens without having internalized the spatiotemporal context the teacher view assembles. This motivates \textbf{spatiotemporal state alignment}, which aligns response-end hidden states at matched depths to transfer the representations output-level supervision cannot reach.

\section{Method}
\label{sec:method}

\begin{figure}[t]
    \centering
    \includegraphics[width=\linewidth]{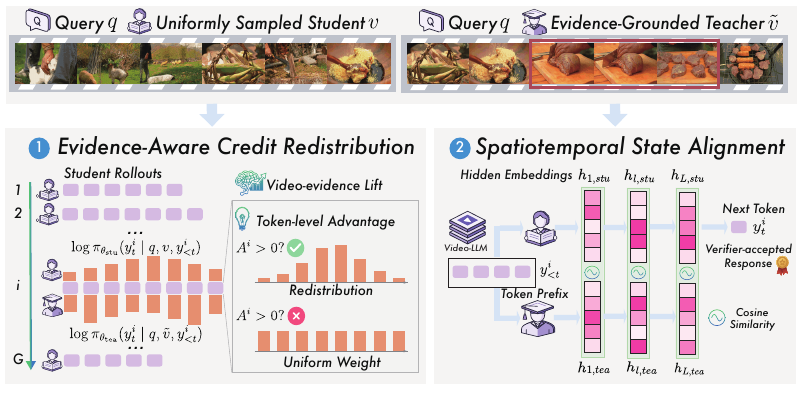}
    \caption{\textbf{Overview of \method.} The student samples responses from uniformly sampled frames, while the teacher scores the same responses using a view that combines global temporal coverage with question-conditioned local windows (red box). \method consists of two components: \textbf{(1)} Evidence-aware credit redistribution uses teacher--student token log-probability contrasts to redistribute positive advantages within each response; non-positive advantages remain unchanged; \textbf{(2)} Spatiotemporal state alignment aligns hidden states through cosine similarity on verifier-accepted responses.}
    \label{fig:method-overview}

\end{figure}

\subsection{Question-Conditioned Video-Evidence Data}
\label{sec:method-data}

We construct Video-Evidence-96K from 104,257 video--question--answer records drawn from LLaVA-Video-178K~\citep{zhang2024llava}, NExT-QA~\citep{xiao2021next}, Perception~\citep{patraucean2023perception}, and STAR~\citep{wu2024starbenchmarksituatedreasoning}. We validate non-empty questions and answers, well-formed multiple-choice options with unique labels, valid video paths, and sufficient temporal candidates for evidence sampling. An automatic question-conditioned frame-relevance annotator~\citep{videoitg} identifies temporally localized evidence; we form local windows around the selected frames and combine them with global frames. The resulting training set contains 95,705 records over 21,832 videos. We checked for video-level overlap between the Video-Evidence-96K and evaluation benchmarks in Section~\ref{sec:experiments}, ensuring that no evaluation videos were included in our training data. The complete construction pipeline and dataset composition are in Appendix~\ref{app:video-evidence-96k}.

\subsection{Evidence-Aware Credit Redistribution}
\label{sec:method-ecr}

\paragraph{Video-evidence lift.}
Different parts of a video response refer to different temporal evidence. For example, a token describing a state change may depend on a short local transition, whereas a token used mainly to connect statements may depend little on any particular frame. A trajectory-level reward cannot express this distinction. ECR converts the teacher--student contrast into token-level weights that emphasize relative support within each response.
Let $y^i=(y^i_1,\ldots,y^i_{T_i})$ be the $i$-th response in a group of $G$ student rollouts, sampled from $\pi_{\theta_{\mathrm{old}}}(\cdot\mid q,v)$. Here $T_i$ is the number of valid response tokens and $y^i_{<t}$ is the prefix before position $t$. We run the same response through both views and define the \emph{token-level evidence contrast} as
\begin{equation}
\ell^i_t
=\log\pi_{\bar\theta}(y^i_t\mid q,\tilde v,y^i_{<t})
 -\log\pi_{\theta_{\mathrm{old}}}(y^i_t\mid q,v,y^i_{<t}),
\qquad t=1,\ldots,T_i ,
\label{eq:ecr-gap}
\end{equation}
where $\bar\theta$ denotes the teacher parameters. A positive $\ell^i_t$ indicates that the global-plus-local view assigns more support to the observed token than the uniform view under the same text prefix. Tokens that express an evidence-sensitive transition or cross-frame relation can therefore exhibit a larger positive contrast, while purely discourse-oriented tokens can exhibit a smaller contrast. The quantity is used as a view-conditioned training signal; it does not require a teacher-generated trajectory. Appendix~\ref{app:advantage-token-stats} reports the proportions of rollouts with positive, negative, and zero advantages across training stages.

\paragraph{Within-response evidence profile.}
The evidence can vary across questions and responses, so ECR redistributes credit according to the relative evidence profile of each response. We first bound extreme contrasts and center them by their response mean:
\begin{equation}
\begin{aligned}
z^i_t
&=\operatorname{clip}\!\left(\ell^i_t,-\log c,\log c\right),&
\mu^i&=\frac{1}{T_i}\sum_{s=1}^{T_i}z^i_s,\\
\widetilde A^i_t
&=A^i_-+A^i_+\exp\!\left(z^i_t-\mu^i\right),&
A^i_+&=\max(A^i,0),\quad A^i_-=\min(A^i,0).
\end{aligned}
\label{eq:ecr-profile}
\end{equation}
The clipping threshold $c$ bounds the teacher-to-student likelihood ratio before centering; we use $c=5$. The residual $z^i_t-\mu^i$ compares position $t$ with the response's own evidence baseline. Equation~\ref{eq:ecr-profile} converts a single response-level advantage into a sequence of token-level advantages: each position $t$ receives its own $\widetilde A^i_t$ according to the evidence contrast at that position. For $A^i>0$, positions with above-baseline evidence receive larger advantages and positions with below-baseline evidence receive smaller ones. For $A^i\leq0$, $A^i_+=0$ and $\widetilde A^i_t=A^i_-$, preserving the ordinary RL penalty. This design effectively lets the privileged view redistribute positive trajectory credit.

\paragraph{Evidence-weighted policy update.}
Let $R(q,y^i)$ be the verifier reward and $A^i$ be the normalized $R(q,y^i)$ within $G$ responses. The current policy's importance ratio for each sampled token is
\[
\rho^i_t(\theta)
=\frac{\pi_\theta(y^i_t\mid q,v,y^i_{<t})}
        {\pi_{\theta_{\mathrm{old}}}(y^i_t\mid q,v,y^i_{<t})}.
\]
 ECR then optimizes the following clipped objective, where $\epsilon$ controls the clipping range:
\begin{equation}
\mathcal{J}_{\mathrm{ECR}}(\theta)
=\mathbb{E}\!\left[
\frac{1}{G}\sum_{i=1}^{G}\frac{1}{T_i}
\sum_{t=1}^{T_i}
\min\!\left\{
\rho^i_t(\theta)\widetilde A^i_t,\,
\operatorname{clip}\!\left(\rho^i_t(\theta),1-\epsilon,1+\epsilon\right)\widetilde A^i_t
\right\}
\right].
\label{eq:ecr-objective}
\end{equation}

\subsection{Spatiotemporal State Alignment}
\label{sec:method-ssa}

Token-level weighting does not explicitly constrain the hidden representations produced by the visual computation. SSA addresses this gap by aligning hidden states produced from the student and teacher views for the same sampled response. Let $h^i_{\ell,v}$ and $h^i_{\ell,\tilde v}$ denote the response-end hidden states for response $y^i$ after decoder layer $\ell$ under views $v$ and $\tilde v$, respectively. The response-end state is the hidden state at the final valid response token after layer $\ell$. Let $\mathcal{S}$ denote the decoder layers, and let $C=\{i:R(q,y^i)=1\}$ be the verifier-accepted response indices, with cardinality $|C|$. We define
\begin{equation}
\mathcal{L}_{\mathrm{SSA}}
=
\frac{1}{\max(1,|C|)\,|\mathcal{S}|}
\sum_{i\in C}\sum_{\ell\in\mathcal{S}}
\left(
1-
\frac{\left\langle h^i_{\ell,v},h^i_{\ell,\tilde v}\right\rangle}
{\left\|h^i_{\ell,v}\right\|_2
 \left\|h^i_{\ell,\tilde v}\right\|_2}
\right).
\label{eq:ssa}
\end{equation}
When $C$ is empty, we set $\mathcal{L}_{\mathrm{SSA}}=0$. SSA provides a shared anchor after both models have processed the question and the same generated text. They can incorporate local appearance changes together with relations among distant events.  Because the targets are computed on student-generated responses, SSA acts on the student's own reasoning contexts. It encourages the student to integrate uniformly sampled observations consistently with the local evidence and global temporal structure.

\subsection{Joint Training and Inference}
\label{sec:method-joint}

The two components are trained jointly by minimizing
\begin{equation}
\mathcal{L}_{\mathrm{OWLET}}(\theta)
=-\mathcal{J}_{\mathrm{ECR}}(\theta)
+\lambda_{\mathrm{SSA}}\mathcal{L}_{\mathrm{SSA}}(\theta),
\label{eq:owlet-objective}
\end{equation}
where the negative sign converts policy maximization into loss minimization and $\lambda_{\mathrm{SSA}}$ controls the strength of state alignment. We set $\lambda_{\mathrm{SSA}}$ as $0.05$. Both terms use responses sampled on-policy from the student view, while the teacher view supplies only token scores and representation targets.

\begin{table}[t]
\centering
\caption{Accuracy (\%) results on five video-understanding benchmarks for Qwen3.5 at 64 input frames. We \textbf{bold} the best result for each backbone and benchmark.}
\label{tab:main}
\resizebox{\textwidth}{!}{
\setlength{\tabcolsep}{4pt}
\renewcommand{\arraystretch}{1.1}
\begin{tabular}{llcccccc}
\toprule
\textbf{Backbone} & \textbf{Method} & \textbf{Video-MME} & \textbf{MVBench} & \textbf{LongVideoBench} & \textbf{Video-MME 2} & \textbf{MLVU} & \textbf{Avg.} \\
\midrule
\multirow{5}{*}{Qwen3.5-0.8B} & Vanilla & 49.4 & 51.8 & 48.9 & 15.8 & 41.9 & 41.6 \\
 & SFT & 49.5 & 52.2 & 49.5 & 15.7 & 42.1 & 41.8 \\
 & GRPO & 50.8 & 53.9 & 50.4 & 18.5 & 43.3 & 43.4 \\
 & OPSD & 50.9 & 53.9 & 51.4 & 17.2 & 44.1 & 43.5 \\
 & \green \method & \green \textbf{54.9} & \green \textbf{58.6} & \green \textbf{55.9} & \green \textbf{19.9} & \green \textbf{45.2} & \green \textbf{46.9} \\
\midrule
\multirow{5}{*}{Qwen3.5-4B} & Vanilla & 64.7 & 67.4 & 58.0 & 24.7 & 50.3 & 53.0 \\
 & SFT & 65.1 & 68.0 & 58.4 & 25.6 & 50.7 & 53.6 \\
 & GRPO & 65.2 & 68.3 & 59.8 & 26.2 & 50.8 & 54.1 \\
 & OPSD & 65.5 & 68.9 & 58.3 & 25.6 & 51.1 & 53.9 \\
 & \green \method & \green \textbf{68.3} & \green \textbf{72.2} & \green \textbf{63.4} & \green \textbf{28.5} & \green \textbf{58.8} & \green \textbf{58.2} \\
\midrule
\multirow{5}{*}{Qwen3.5-9B} & Vanilla & 65.0 & 67.1 & 61.9 & 25.6 & 52.7 & 54.4 \\
 & SFT & 65.9 & 67.2 & 62.5 & 26.5 & 53.7 & 55.2 \\
 & GRPO & 66.3 & 67.7 & 63.4 & 26.8 & 53.5 & 55.5 \\
 & OPSD & 66.1 & 68.7 & 62.9 & 26.7 & 53.3 & 55.5 \\
 & \green \method & \green \textbf{70.7} & \green \textbf{72.8} & \green \textbf{65.2} & \green \textbf{28.9} & \green \textbf{61.6} & \green \textbf{59.8} \\
\midrule
\multirow{5}{*}{Qwen3.5-27B} & Vanilla & 71.4 & 70.2 & 64.2 & 30.4 & 61.5 & 59.5 \\
 & SFT & 71.8 & 70.5 & 64.2 & 30.6 & 61.8 & 59.8 \\
 & GRPO & 72.0 & 71.1 & 64.5 & 32.1 & 61.1 & 60.2 \\
 & OPSD & 71.4 & 70.9 & 65.1 & 30.7 & 62.9 & 60.2 \\
 & \green \method & \green \textbf{76.1} & \green \textbf{76.2} & \green \textbf{68.8} & \green \textbf{34.6} & \green \textbf{66.7} & \green \textbf{64.5} \\
\bottomrule
\end{tabular}
}
\end{table}

\section{Experiments}\label{sec:experiments}

We evaluate \method on:
(1)~main comparisons with post-training baselines across four Qwen3.5 scales and five video benchmarks;
(2)~robustness across student frame budgets;
(3)~ablations of the teacher view, ECR, and SSA;
(4)~training dynamics case studies; and
(5)~generalization to Qwen3-VL in non-thinking and thinking modes (Appendix~\ref{app:qwen3vl-results}).
Appendices~\ref{app:opd-variants}--\ref{app:longer-training} additionally include: comparison with on-policy distillation variants, different teacher parameterization settings, advantage and token-weight statistics, SSA-weight sensitivity, response examples of different post-trained models, training-time comparisons, response-length statistics, global-temporal accuracy after post-training, and extended training schedules.

\subsection{Experiment Setups} \label{sec:set_up}

\paragraph{Implementations}
We post-train Qwen3.5-0.8B/4B/9B/27B~\citep{qwen35} in non-thinking mode, each with a student frame budget $N\in\{16,32,64\}$. We also evaluate Qwen3-VL 2B/4B/8B in both non-thinking and thinking modes in Appendix~\ref{app:qwen3vl-results} to demonstrate the generalization of \method. We report accuracy on Video-MME~\citep{fu2025video}, MVBench~\citep{li2024mvbench}, LongVideoBench~\citep{wu2024longvideobench}, Video-MME-2~\citep{fu2026videommev2}, and MLVU~\citep{zhou2025mlvu}. We use eight NVIDIA H200 (140GB) GPUs and a learning rate of $1\times 10^{-6}$. Details are in Appendices~\ref{app:baselines-benchmarks} and~\ref{app:training-hparams}.

\subsection{Main Results} \label{sec:main_res}

As shown in Table~\ref{tab:main}, \method outperforms SFT, GRPO, and OPSD on all five benchmarks for the four measured Qwen3.5 scales. We also observe that the largest per-benchmark gains typically appear on MVBench and MLVU, which stress fine-grained temporal understanding and long-video reasoning. This pattern matches the design of the teacher view: local windows supply the temporal detail that evidence-sensitive questions require, while the global track preserves the long-range structure that long-video benchmarks demand in Section~\ref{sec:motivation-teacher-view}.

Table~\ref{tab:budget} reports the macro-average at 16, 32, and 64 input frames. \method achieves the best result in all twelve combinations of model size and frame budget, demonstrating that its advantage is robust to the amount of visual evidence available to the student. Qwen3.5-4B with \method also exceeds the strongest 9B baseline at the same budget, indicating that the proposed evidence-guided supervision can improve the effectiveness of a smaller model. Appendix~\ref{app:main-results-analysis} provides per-benchmark results at 16 and 32 frames, detailed question-group results for Video-MME~2, and the Video-MME duration splits. Beyond the Qwen3.5 family, we also post-train Qwen3-VL-2B/4B/8B in both non-thinking and thinking modes; \method again attains the highest macro-average in all settings. The detailed results and analyses are provided in Appendix~\ref{app:qwen3vl-results}.

\begin{table}[t]
\centering
\caption{Macro-average accuracy (\%) over the five benchmarks for Qwen3.5, at $16$, $32$, and $64$ input frames. We \textbf{bold} the best result for each backbone and frame budget.}
\label{tab:budget}
\footnotesize
\setlength{\tabcolsep}{6pt}
\renewcommand{\arraystretch}{1.1}
\begin{tabular}{lcccccccccccc}
\toprule
\multirow{2}{*}{\textbf{Method}} & \multicolumn{3}{c}{\textbf{Qwen3.5-0.8B}} & \multicolumn{3}{c}{\textbf{Qwen3.5-4B}} & \multicolumn{3}{c}{\textbf{Qwen3.5-9B}} & \multicolumn{3}{c}{\textbf{Qwen3.5-27B}} \\
\cmidrule(lr){2-4} \cmidrule(lr){5-7} \cmidrule(lr){8-10} \cmidrule(lr){11-13}
 & 16 & 32 & 64 & 16 & 32 & 64 & 16 & 32 & 64 & 16 & 32 & 64 \\
\midrule
Vanilla & 37.8 & 40.1 & 41.6 & 48.0 & 49.8 & 53.0 & 49.0 & 51.3 & 54.4 & 54.3 & 56.8 & 59.5 \\
SFT & 38.6 & 40.9 & 41.8 & 48.5 & 50.7 & 53.6 & 49.8 & 52.2 & 55.2 & 54.8 & 56.1 & 59.8 \\
GRPO & 39.6 & 41.9 & 43.4 & 49.8 & 51.8 & 54.1 & 50.5 & 52.9 & 55.5 & 55.7 & 57.9 & 60.2 \\
OPSD & 39.1 & 41.5 & 43.5 & 49.1 & 51.0 & 53.9 & 50.3 & 52.6 & 55.5 & 55.7 & 58.1 & 60.2 \\
\green \method & \green \textbf{42.9} & \green \textbf{45.2} & \green \textbf{46.9} & \green \textbf{53.3} & \green \textbf{54.6} & \green \textbf{58.2} & \green \textbf{54.2} & \green \textbf{57.1} & \green \textbf{59.8} & \green \textbf{57.9} & \green \textbf{60.4} & \green \textbf{64.5} \\
\bottomrule
\end{tabular}
\end{table}

\subsection{Ablation Study} \label{sec:ab}

\begin{figure*}[t]
    \centering
    \begin{subfigure}[t]{0.32\textwidth}
        \centering
        \includegraphics[width=\linewidth]{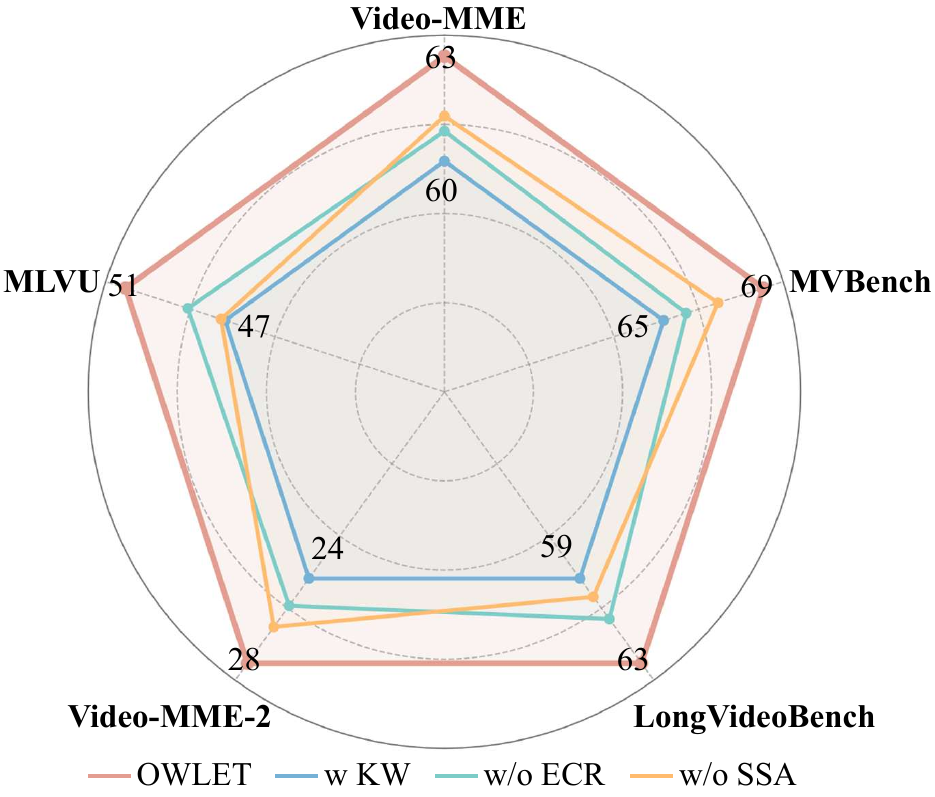}
        \caption{$16$ frames}
    \end{subfigure}\hfill
    \begin{subfigure}[t]{0.32\textwidth}
        \centering
        \includegraphics[width=\linewidth]{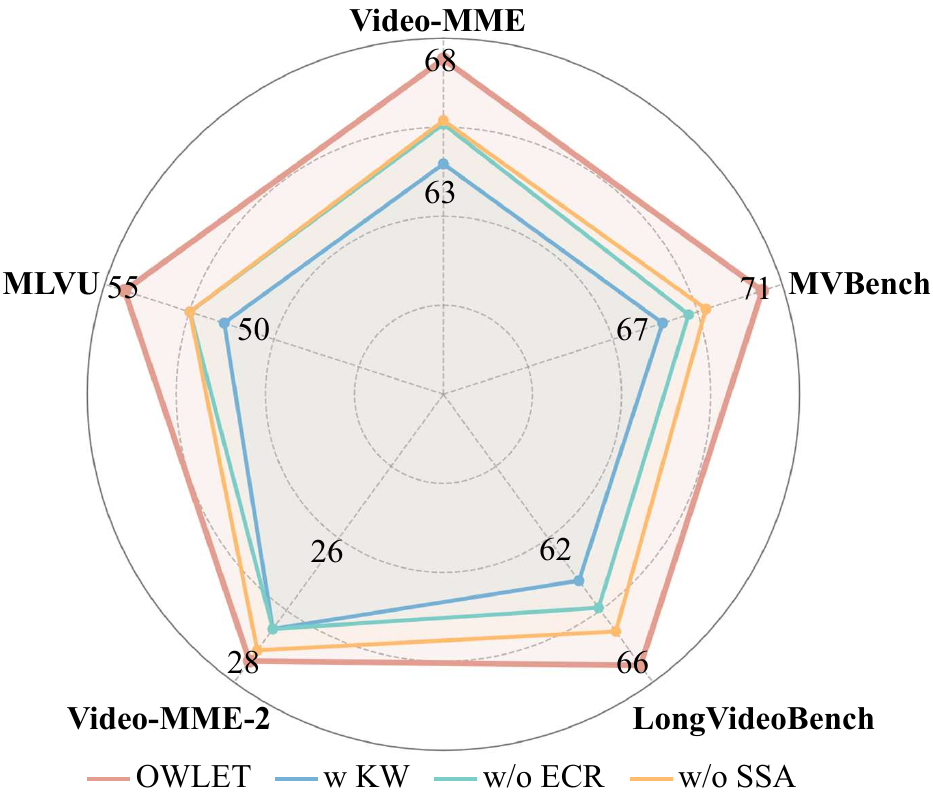}
        \caption{$32$ frames}
    \end{subfigure}\hfill
    \begin{subfigure}[t]{0.32\textwidth}
        \centering
        \includegraphics[width=\linewidth]{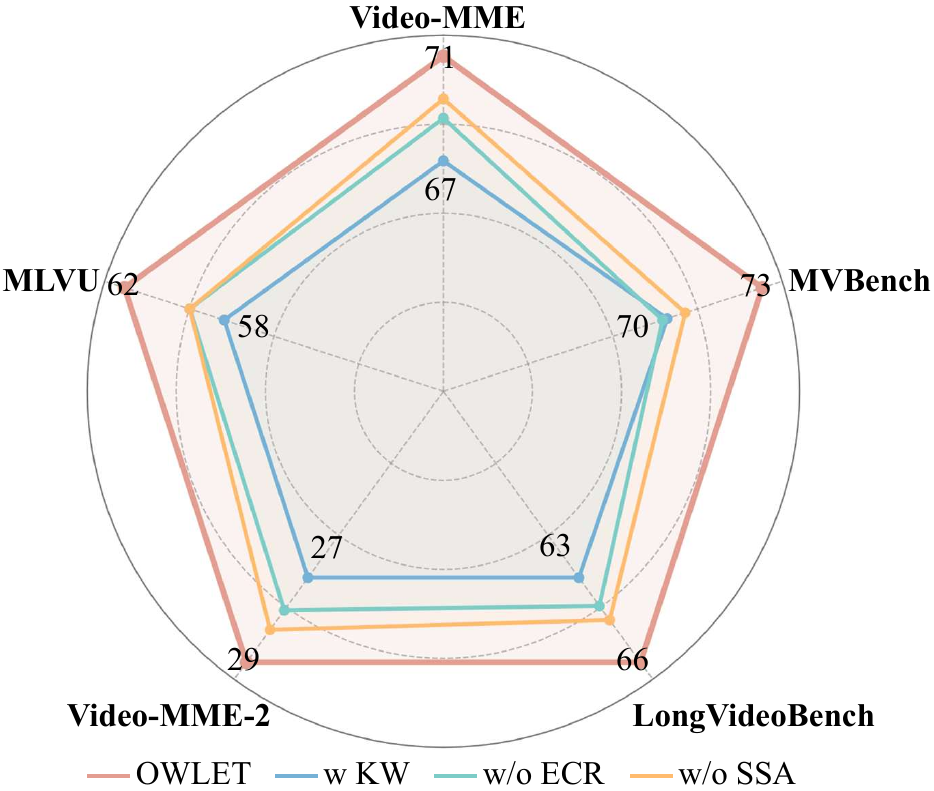}
        \caption{$64$ frames}
    \end{subfigure}
    \caption{Ablation results for Qwen3.5-9B across different frame budgets. }
    \label{fig:ablation-radar}
\end{figure*}

To investigate the contribution of the proposed components, we conduct an ablation study on Qwen3.5-9B across 16, 32, and 64 input frames. We consider three variants: \method with a key-window-only teacher view (w/ Key Windows), \method without evidence-aware credit redistribution (w/o ECR), and \method without spatiotemporal state alignment (w/o SSA). All variants share the training and evaluation protocol with the full model. As shown in Figure~\ref{fig:ablation-radar}, removing any component degrades performance across the five benchmarks and frame budgets. Replacing the global-plus-local teacher view with key windows produces the largest average decrease, indicating that global temporal coverage is complementary to localized evidence. The full \method remains strongest at every budget, while the key-window variant remains the weakest on average, showing that additional local frames do not replace the benefit of combining global coverage with local evidence.
Additional ablations across the remaining Qwen3.5 scales and Qwen3-VL modes are provided in Appendix~\ref{app:ablation-results}.

\subsection{Case Study}\label{sec:cs}

To understand where the improvements of \method come from, we conduct case studies on Qwen3.5-9B with 64 frames, tracking how the supervision gaps identified in Section~\ref{sec:motivation-gaps} evolve over training.

\begin{figure}[t]
    \centering
    \includegraphics[width=\linewidth]{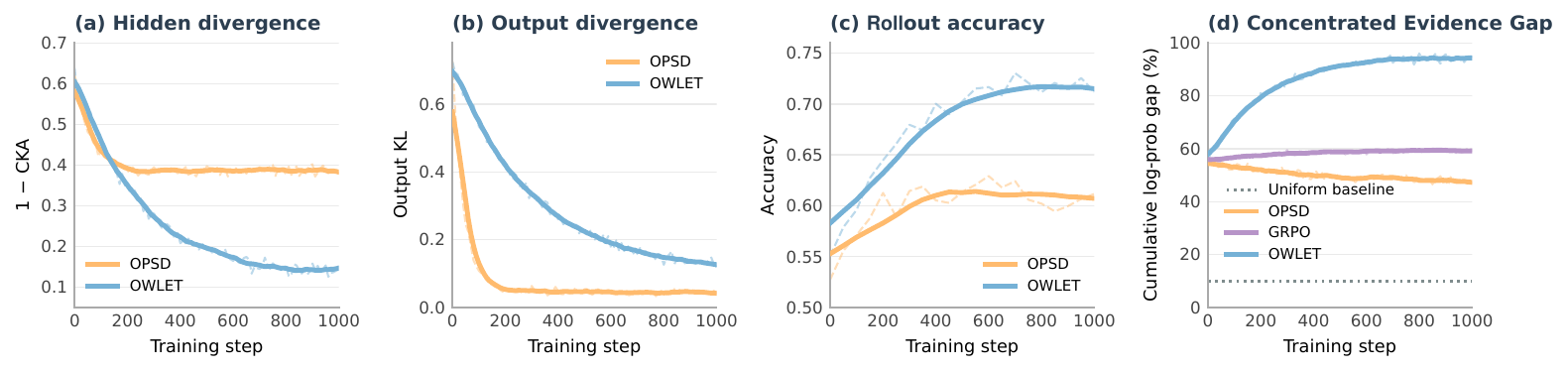}
    \caption{\textbf{Training dynamics.} Qwen3.5-9B with 64 student frames. \textbf{(a)} Hidden divergence, measured by response-end $1-\mathrm{CKA}$, for OPSD and OWLET. \textbf{(b)} Output KL between student and teacher views for OPSD and OWLET. \textbf{(c)} Rollout accuracy on the training set for OPSD and OWLET. \textbf{(d)} Cumulative absolute log-probability gap at the top 10\% of response tokens for OPSD, GRPO, and OWLET; the dotted line marks the uniform baseline. Diagnostic samples are spaced 10 steps apart and accuracy samples 50 steps apart. Thin dashed lines connect sampled trajectories; thick lines show rolling means.}
    \label{fig:case-ssa-dynamics}
\end{figure}

\paragraph{Exploration of Representation Dynamics.}
Figure~\ref{fig:case-ssa-dynamics}(a--c) tracks the response-end hidden divergence ($1-\mathrm{CKA}$), the output KL between the two views, and the rollout accuracy on the training set for OPSD and \method. We observe a clear dissociation between the two interfaces. OPSD rapidly minimizes its own objective---the output KL falls below $0.05$---yet its hidden divergence plateaus near $0.39$, the level of the diagnostic run in Figure~\ref{fig:motivation-combined}(d), and its accuracy saturates near $61\%$. In contrast, \method steadily drives the hidden divergence from the shared starting point of $0.62$ down to about $0.13$ while its accuracy climbs from $54\%$ to $73\%$; notably, its output KL also decreases to about $0.11$ without ever being optimized directly. This dissociation confirms the gap identified in Section~\ref{sec:motivation-gaps}: matching output distributions alone leaves the upstream spatiotemporal representation misaligned, whereas explicitly aligning that representation through SSA closes both divergences and accompanies substantially higher rollout accuracy.

\paragraph{Exploration of Evidence-Gap Concentration.}
Figure~\ref{fig:case-ssa-dynamics}(d) examines how each objective reshapes the structure of the evidence gap. We report the share of the total per-response gap $\sum_t|\Delta\log p_t|$ carried by its top 10\% of tokens for post-trained methods; at initialization every method inherits the base model's concentration of about $56\%$. The three objectives then diverge. Under \method, the share climbs steeply over $90\%$: as ECR keeps routing positive credit toward evidence-dependent positions, the two views come to agree on scaffolding tokens, and nearly the entire residual gap collapses onto the small evidence-bearing set. GRPO, which never sees the teacher view during training, rises mildly to about $60\%$: reward-driven training strengthens visual grounding and thereby makes evidence tokens somewhat more view-sensitive, but nothing in a scalar reward concentrates credit further. OPSD moves in the opposite direction, easing down to about $48\%$: its KL gradient is largest precisely at the high-gap positions, so it preferentially erodes the concentrated head---the flattening that Section~\ref{sec:motivation-gaps} shows discards the most informative part of the signal---while the erosion remains mild because the objective converges (panel b) well before the relative structure is gone. Together with panel (c), these trends indicate that \method converts the concentrated evidence signal into higher rollout accuracy rather than averaging it away.

\section{Conclusion}
\label{sec:conclusion}

In this paper, we introduce \method, a novel evidence-privileged reinforcement-learning framework that performs \textbf{o}n-policy \textbf{w}eighted \textbf{l}earning via \textbf{e}vidence \textbf{t}ransfer for video-LLM post-training. \method supervises each student rollout with a teacher view of the same video that pairs {global temporal context} with {question-conditioned local windows}, and closes the two supervision gaps identified by our motivated experiments through two key innovations: (i) evidence-aware credit redistribution, which converts the teacher--student token contrast into \textbf{token-level credit} while the verifier retains control of the learning direction; and (ii) spatiotemporal state alignment, which transfers the \textbf{spatiotemporal representations} that output-level supervision cannot reach. Trained on the proposed Video-Evidence-96K corpus, \method adds no model capacity and no test-time cost: the student answers from uniformly sampled frames in a single pass, with no evidence annotation or teacher computation. Experiments on Qwen3.5 and Qwen3-VL backbones from 0.8B to 27B parameters across five benchmarks demonstrate the effectiveness of \method, yielding up to $10.9$ accuracy improvement over the vanilla model.


\section*{AI use statement}




In this work, we used LLM-based assistants to polish the wording of text. We have not used generative AI tools to produce research ideas or experimental designs. We also used generative AIs to assign global-temporal and local-evidence question labels, assess the quality of question-conditioned evidence annotations, and to generate the chain-of-thought traces for the SFT baseline.  No generative AI tool had access to non-public data. We have reviewed all AI-assisted work: all AI-assisted text was checked, corrected, and approved by the authors, and all AI-assisted code was verified for correctness, with every reported result produced by the authors' own runs. We take responsibility for the final content of this work, including text, claims, or artifacts produced with the aid of generative AI.

\section*{Ethics statement}



This work post-trains and evaluates video language models exclusively on publicly available academic datasets and benchmarks, used under their respective licenses; no private or personally collected data is involved, and no human-subject study was conducted beyond the authors' own verification of automatic annotations. Video-Evidence-96K is derived from existing public corpora, and we will document its provenance and filtering rules upon release so that upstream restrictions remain traceable. The proposed method improves how models use visual evidence and, to our knowledge, does not introduce risks beyond those already present in the underlying video-LLMs, such as biases inherited from public video corpora. As with any advance in video understanding, stronger temporal reasoning could in principle be misused in surveillance settings; our training recipe neither targets nor specifically facilitates such uses.

\section*{Reproducibility statement}



We document everything needed to reproduce our results in the paper and its appendix. The construction pipeline of Video-Evidence-96K, including source datasets, validation rules, evidence annotation, and splits, is described in Appendix~\ref{app:video-evidence-96k}; the protocols behind the motivated experiments are specified in Appendices~\ref{app:motivated-protocol}--\ref{app:ssa-protocol}; baselines, benchmarks, and the complete set of training and evaluation hyper-parameters are listed in Appendices~\ref{app:baselines-benchmarks} and~\ref{app:training-hparams}; the method itself is fully specified by Equations~\ref{eq:ecr-gap}--\ref{eq:owlet-objective}. We will release the training code, the dataset construction scripts, and the evaluation process upon publication.


\clearpage


\appendix

\clearpage

\addtocontents{toc}{\protect\setcounter{tocdepth}{2}}

\renewcommand{\contentsname}{Appendix Table of Contents}

\tableofcontents

\section{Limitations}
\label{app:limitations}

We consider several limitations and directions for future work. \textbf{(i) Reliance on automatically localized evidence.} Video-Evidence-96K is built with an automatic question-conditioned frame-relevance annotator, which the privileged teacher view uses to construct local windows. Although the global track provides broad temporal coverage, inaccurate or noisy relevance estimates can still expose the teacher to misleading evidence and affect both ECR and SSA. Future work could incorporate confidence-aware or jointly learned evidence selection, and multiple evidence proposals. \textbf{(ii) Dependence on verifiable outcome supervision.} Our training objective uses binary answer verification, and our experiments focus on multiple-choice video benchmarks. ECR preserves the verifier's response-level learning direction, while SSA is applied to verifier-accepted responses; for open-ended, ambiguous, or subjective questions, an equally reliable verifier may be unavailable. Extending the framework to free-form video QA and richer temporal reasoning will require semantic or process-level verifiers, preference signals, or other forms of outcome supervision. \textbf{(iii) Training cost and fixed design choices.} Inference of \method has no additional teacher computation, but the training process requires an extra forward pass through a frozen teacher under the global-plus-local view. Scaling \method to much longer videos, larger model scales, or different non-Qwen model families remains to be studied. Adaptive teacher views, competence-aware teacher weighting, and layer- or token-level state alignment are promising directions. \textbf{(iv) Indirect evidence attribution.} ECR uses the teacher--student likelihood contrast as an indirect measure of how strongly each response token is supported by visual evidence; it does not provide an explicit causal attribution of the underlying visual cues. This lightweight signal nevertheless remains effective across model scales, frame budgets, and backbone families in our experiments, while future work could explore its interpretability and combine it with explicit grounding or process-level supervision.

\section{Formalized Algorithms of \method}
Algorithm~\ref{alg:owlet} summarizes the training procedure of \method{}. For each training instance, the student first samples a group of on-policy responses from a uniformly sampled video view $v$. The same student-generated responses and prefixes are then evaluated by the rollout policy under $v$ and by a fixed teacher initialized at $\theta_0$ under a global-local view $\tilde{v}$. A verifier assigns binary rewards, from which group-normalized advantages are computed. Evidence-aware credit redistribution (ECR) contrasts the teacher and rollout token log-likelihoods, clips and centers the contrasts within each response, and redistributes positive advantage toward tokens better supported by the privileged view. In parallel, spatiotemporal state alignment (SSA) aligns response-end hidden states from the student and teacher on verifier-accepted responses, with gradients propagated only through the student. The student is optimized using the joint objective $(-\mathcal{J}_{\mathrm{ECR}}+\lambda_{\mathrm{SSA}}\mathcal{L}_{\mathrm{SSA}})$, while the teacher remains frozen throughout training. At inference time, only the trained student, the uniform video view, and the question are required.
\begin{algorithm}[t]
\caption{Training Procedure of \method}
\label{alg:owlet}
\begin{algorithmic}[1]
\Require Training records \(\mathcal D=\{(\mathcal V,q,a,\mathcal W)\}\); initial checkpoint \(\theta_0\); frame budget \(N\); group size \(G\); training steps \(K\);  clipping range \(\epsilon\); ECR threshold \(c\); SSA weight \(\lambda_{\mathrm{SSA}}\); decoder layers \(\mathcal S\)
\Ensure Trained student parameters \(\theta\)
\State \(\theta\gets\theta_0\), \(\bar\theta\gets\theta_0\) \Comment{the teacher \(\pi_{\bar\theta}\) stays fixed}
\For{$k=1,\ldots,K$}
    \State \(\theta_{\mathrm{old}}\gets\theta\)
    \State Sample a batch \(\mathcal B\subset\mathcal D\)
    \For{each $(\mathcal V,q,a,\mathcal W)\in\mathcal B$}
        \State \(v\gets\operatorname{UniformSample}(\mathcal V,N)\)
        \State \(\tilde v\gets\operatorname{GlobalLocalView}(\mathcal V,\mathcal W;N)\)
        \State Sample \(\{y^i\}_{i=1}^{G}\sim\pi_{\theta_{\mathrm{old}}}(\cdot\mid q,v)\); let \(T_i\) be the valid length of \(y^i\)
        \State \(R(q,y^i)\gets\operatorname{Verify}(q,y^i;a)\in\{0,1\}\) for \(i=1,\ldots,G\)
        \State \(A^i\gets\dfrac{R(q,y^i)-\operatorname{mean}_{j=1}^{G}R(q,y^j)}{\operatorname{std}_{j=1}^{G}R(q,y^j)}\) for \(i=1,\ldots,G\)
        \State \(C\gets\{i:R(q,y^i)=1\}\)
        \For{$i=1,\ldots,G$}
            \For{$t=1,\ldots,T_i$}
                \State \(\ell_t^i\gets\log\pi_{\bar\theta}(y_t^i\mid q,\tilde v,y_{<t}^i)-\log\pi_{\theta_{\mathrm{old}}}(y_t^i\mid q,v,y_{<t}^i)\)
                \State \(z_t^i\gets\operatorname{clip}(\ell_t^i,-\log c,\log c)\)
            \EndFor
            \State \(\mu^i\gets T_i^{-1}\sum_{s=1}^{T_i}z_s^i\), \quad \(A_+^i\gets\max(A^i,0)\), \quad \(A_-^i\gets\min(A^i,0)\)
            \For{$t=1,\ldots,T_i$}
                \State \(\widetilde A_t^i\gets A_-^i+A_+^i\exp(z_t^i-\mu^i)\)
                \State \(\rho_t^i(\theta)\gets\dfrac{\pi_\theta(y_t^i\mid q,v,y_{<t}^i)}{\pi_{\theta_{\mathrm{old}}}(y_t^i\mid q,v,y_{<t}^i)}\)
            \EndFor
            \State Compute \(\{h_{\ell,v}^i\}_{\ell\in\mathcal S}\) with \(\pi_\theta\) on \((q,v,y^i)\)
            \State Compute \(\{h_{\ell,\tilde v}^i\}_{\ell\in\mathcal S}\) with frozen \(\pi_{\bar\theta}\) on \((q,\tilde v,y^i)\), without gradients
        \EndFor
        \State \(\mathcal J_{\mathrm{ECR}}(q)\gets\dfrac{1}{G}\sum_{i=1}^{G}\dfrac{1}{T_i}\sum_{t=1}^{T_i}\min\!\left(\rho_t^i(\theta)\widetilde A_t^i,\operatorname{clip}(\rho_t^i(\theta),1-\epsilon,1+\epsilon)\widetilde A_t^i\right)\)
        \State \(\mathcal L_{\mathrm{SSA}}(q)\gets\dfrac{1}{\max(1,|C|)|\mathcal S|}\sum_{i\in C}\sum_{\ell\in\mathcal S}\left[1-\cos\!\left(h_{\ell,v}^i,\operatorname{sg}(h_{\ell,\tilde v}^i)\right)\right]\)
        \State \(\mathcal L_{\mathrm{SSA}}(q)\gets0\) when \(C=\varnothing\)
    \EndFor
    \State \(\mathcal J_{\mathrm{ECR}}\gets|\mathcal B|^{-1}\sum_{(\mathcal V,q,a,\mathcal W)\in\mathcal B}\mathcal J_{\mathrm{ECR}}(q)\)
    \State \(\mathcal L_{\mathrm{SSA}}\gets|\mathcal B|^{-1}\sum_{(\mathcal V,q,a,\mathcal W)\in\mathcal B}\mathcal L_{\mathrm{SSA}}(q)\)
    \State \(\mathcal L_{\mathrm{OWLET}}(\theta)\gets-\mathcal J_{\mathrm{ECR}}(\theta)+\lambda_{\mathrm{SSA}}\mathcal L_{\mathrm{SSA}}(\theta)\)
    \State Update only \(\theta\) by back-propagating \(\mathcal L_{\mathrm{OWLET}}\); keep \(\bar\theta=\theta_0\) fixed
\EndFor

\State \Return \(\theta\)
\end{algorithmic}
\end{algorithm}
\clearpage

\section{Related Work}
\label{sec:related_work}

\subsection{Long-Video Understanding}

Video-LLMs extend image-language models to the temporal domain. A fixed number of frames is sampled from the video, turned into visual tokens, and placed before the question in a single input sequence~\citep{qwen3vl, qwen35, zhu2025internvl3, chen2024expanding, videosalmonn2}. This is the standard pipeline for video question answering, and benchmarks now test it on longer videos and harder question types~\citep{fu2025video, wu2024longvideobench, wang2025lvbench, zhou2025mlvu, chen2025cg}. Its central constraint is that context length and compute cap how many frames the model can read, while the videos themselves keep growing. An hour of footage reaches the model as a few dozen frames, so the frame budget decides how much of the video is actually seen.

One line of work enlarges this budget by making frames cheaper. Token compression reduces the tokens spent per frame~\citep{li2026videochat, shen2024longvu, llamavid, lv2026hippo}, memory mechanisms summarize past content into a compact state that is carried forward~\citep{song2024moviechatdensetokensparse, zhang2025flash, shu2025video}, and redundancy-aware encoders drop content that repeats across neighboring frames~\citep{wang2025retakereducingtemporalknowledge}. Training recipes for hour-scale material push in the same direction~\citep{ranasinghe2025understanding, lin2026unleashing}. These methods change how much video reaches the model, but the supervision that teaches it what to do with the video is still the answer alone.

A second line asks which frames to keep. The evidence that decides an answer usually sits in a short span, and the rest of the video contributes little or actively distracts~\citep{zou2024secondshoursreviewingmultimodal, xiao2024itrustanswervisually}. Query-agnostic selectors score frames by coverage or diversity~\citep{mdp3_iccv2025, videotree_cvpr2025}, while question-conditioned selectors and grounding models locate the span the question refers to~\citep{videoitg, frame_voyager_iclr2025, mllm_frame_selection_cvpr2025, ffs_cvpr2025, zhang2023temporal}. Some datasets annotate this span directly, giving clue intervals or evidence frames alongside each question~\citep{chen2025cg, wu2024starbenchmarksituatedreasoning, cheng2025videoholmesmllmthinklike, vited}. Selection alone is not enough, however: questions about order, duration, or change are answered from what happens \textit{between} the selected frames~\citep{xiao2021next, liu2024tempcompass}, so keeping only the evidence frames removes the context that makes them readable.

More recently, \textit{thinking-with-videos} methods move evidence localization into inference. The model searches for relevant segments, inspects them closely, and feeds the result back into its reasoning~\citep{wang2024videoagent, yang2025vca, yuan2025videodeepresearch, liu2026videotemp, yang2026longvt, zhang2026deep, pan2025timesearch}. Localization becomes explicit, but each question now costs several rounds of search or tool calls. \method pays this cost during training instead. The teacher view, uniform global coverage together with question-conditioned local windows, is built only at training time, and the student is measured against it. At inference the student runs the ordinary single-pass pipeline: no search, no grounding module, and no extra forward passes.

\subsection{Reinforcement Learning for Video-LLMs}

Reinforcement learning is now the standard route to reasoning ability in large models. Early alignment methods optimize a learned preference model with policy-gradient algorithms~\citep{ouyang2022training, schulman2017proximal}. Reinforcement learning with verifiable rewards instead checks the final answer with a rule, and estimates the advantage by comparing responses sampled for the same prompt, so no critic is needed~\citep{shao2024deepseekmath, guo2025deepseekr1}. Reinforcement learning samples from the current policy, so training stays on the distribution the model uses at test time.

Adapting the recipe to video has produced a large body of work. Video-R1 adds temporal rewards that discourage answering from a single frame~\citep{videor1}, and related methods target temporal consistency and self-evolution~\citep{huang2026evovid, videoutr}, spatiotemporal perception~\citep{li2025videochat, wang2025time}, grounded reasoning with explicit evidence~\citep{meng2026openo3videogroundedvideoreasoning}, difficulty-aware optimization~\citep{park2025deepvideo}, and adaptive temporal search~\citep{pan2025timesearch}. These works change what the reward measures or which trajectories are sampled, but the reward stays a single scalar attached to the whole trajectory.

This creates a credit-assignment problem that video makes worse. The same reward is copied to every token, yet the tokens of a video response rely on the visual stream to very different degrees: a token that names an observed action and a token that closes a sentence receive equal credit. In the text domain, this has been addressed with process reward models and step-level credit~\citep{luo2024improve, cui2025process, cheng2026stop, li2025treepo}, and with token-level signals read off the policy's own uncertainty~\citep{cui2025entropy, cheng2025reasoning, xu2026beyond, wang2026distilled}. Both define importance over the reasoning text and are blind to the video. We also draw inspiration from these text domain token-level RL methods. Closer to our setting, some video methods score tokens by how much the model attends to visual tokens~\citep{du2026appoattentionguidedperceptionpolicy, wang2026video}. These scores are read from a single forward pass and can be driven by language priors: they say where the model looked, not whether what it saw supports the token. \method obtains the weight by comparing the token likelihood assigned by a fixed teacher under the privileged view with that assigned by the student rollout policy under the uniform view, using the same question and response prefix.

\subsection{On-Policy Distillation and Self-Distillation}

Knowledge distillation transfers capability by matching a student to a teacher's predictive distribution~\citep{hinton2015distilling, kim2016sequence}. Run on a fixed corpus of teacher-generated sequences, it supervises the student on states that may differ from those visited during generation. On-policy distillation (OPD) removes this mismatch: the response is sampled from the student, and a token-level divergence to the teacher is minimized on exactly the states the student reaches~\citep{gu2024minillm, agarwal2024policy, lu2025onpolicydistillation}. This gives supervision at every position for roughly one extra teacher forward pass per rollout. Later work studies the choice of divergence, the weighting of tokens, the interaction with RL, and the failure modes of the objective~\citep{wu2025rethinking, xu2025kdrl, jin2026entropy, xu2026tiptokenimportanceonpolicy, fu2026revisiting, li2026rethinking, wu2026lightning}.

On-policy self-distillation (OPSD) derives the teacher from the same underlying model and supplies privileged information that the student does not receive~\citep{hubotter2026reinforcement, zhao2026self, shenfeld2026self}. This provides supervision without requiring a larger external teacher; the choice of whether to freeze or update the teacher is a separate design decision. In our implementation, teacher and student share their initial checkpoint, and the teacher remains fixed as the student is trained. So far the privileged information has been almost always textual: a reference rationale~\citep{yang2026self}, a longer context~\citep{ye2026policy, zhang2026opsdlonpolicyselfdistillationlongcontext}, or feedback from the environment~\citep{shenfeld2026self}. Related variants study weak-to-strong pairings, self-supervised teachers, reasoning compression, and adaptive supervision horizons~\citep{yu2026weak, tan2026self, sang2026policy, hou2026dash}.

In the multimodal setting the privileged information becomes visual. Vision-OPD gives the teacher evidence-centered crops while the student sees the full image~\citep{visionopd}. Later methods reweight vision-critical tokens~\citep{liu2026visualadvantageonpolicydistillationvisionlanguage}, transfer reasoning from a text teacher~\citep{vold}, steer gradients toward visual grounding~\citep{decomposedopd, vicur}, stabilize the objective for multimodal reasoners~\citep{gndpo}, and study contrastive forms of asymmetric visual supervision~\citep{li2026visual, liang2026visual}. In the temporal domain, Video-OPD applies OPD to video grounding~\citep{videoopd}, and rationale- or agent-based methods transfer temporal reasoning through generated text~\citep{aotd, vited}. All of them work in output space: what is matched is a distribution over the next token.

That interface loses information. A logit is the last step of a computation that has already merged the visual stream, and a growing literature argues that much of what matters for reasoning lives in the intermediate states rather than in the tokens they produce~\citep{hao2025traininglargelanguagemodels, latentvisualreasoning, xia2026perceptionreasoningdynamiclatent, zhang2026latentvisualcachevideo}. Hidden-state distillation has accordingly been studied for multimodal models and for pre-training~\citep{llavadi, llavakd, hiddenkd}, and OPRD extends OPD into representation space for text LLMs~\citep{oprd}. Video sharpens the point: cues from distant frames often need to be merged into one spatiotemporal representation well before the first answer token, so matching output distributions leaves that merging step unsupervised.

\method differs from this literature in three ways. First, the privileged information is temporal rather than textual: another view of the same video, combining uniform global coverage with question-conditioned local windows, which does not require a stronger model. Second, the teacher view does not replace the reward with an imitation objective; the verifier still sets the direction of learning, and the two views are compared only to redistribute credit inside a response. Third, the transfer acts at two levels, on token weights and on response-end spatiotemporal representations, so that both what the student reinforces and what it internally computes are shaped by the richer view.

\section{Details of Experiment Settings} \label{app:details}

This section provides the detailed experimental setup for Sections~\ref{sec:motivation} and \ref{sec:experiments}: the data and decoding protocol (Appendix~\ref{app:motivated-protocol}), the construction of the four input views and the question taxonomy used by the teacher-view comparison (Appendices~\ref{app:teacher-view-construction} and~\ref{appendix:global-temporal-questions}), the teacher-view results on a second model family (Appendix~\ref{app:teacher-view-qwen3vl}), and the two measurements behind Figure~\ref{fig:motivation-combined}(c,d) (Appendices~\ref{app:ecr-protocol} and~\ref{app:ssa-protocol}). We use the notation from Section~\ref{sec:pre}: $v$ denotes the student view, $\tilde{v}$ the teacher view, and the student model and the teacher model are both initialized from the same checkpoint.

\subsection{Evaluation Data and Decoding}
\label{app:motivated-protocol}

The view comparison and the two motivation diagnostics use $1{,}000$ samples held out from Video-Evidence-96K and disjoint from the training split. Questions are in multiple-choice form, and accuracy is the exact match of the predicted option letter; for the credit measurement we cap generation at $2048$ new tokens.

For a controlled comparison, every view in Appendix~\ref{app:teacher-view-construction} is assembled from exactly the same number of frames.

\subsection{Construction of the Four Input Views}
\label{app:teacher-view-construction}

Every record in Video-Evidence-96K stores a question-conditioned relevance score over a densely sampled pool of candidate frames, from which we take the ranking used below. We construct the four views at total frame budgets of $B\in\{16,32,64\}$, reusing the same candidate pool and relevance ranking across budgets. Due to computational resource constraints, we do not train \method with frame budgets larger than $64$. A set of windows is \textit{disjoint} when no two of them share a frame, which is what keeps the total frame count exact. Window centers are selected in decreasing relevance order, skipping any candidate whose window overlaps an earlier one. For all sampling methods, the selected frames are arranged in chronological order before being fed to the model. Table~\ref{app:tab-views} gives the composition at each budget.

\begin{table}[h]
\centering
\caption{Composition of the four input views defined in Section~\ref{sec:motivation-teacher-view} at total budgets of 16, 32, and 64 frames. Ranked frames are selected by question-conditioned relevance and used directly in Key frames or as centers of disjoint windows in Key windows and Global with local. All four views use the same total number of frames within each budget.}
\label{app:tab-views}
\begin{tabular}{lcccc}
\toprule
View & Uniform frames & Ranked frames & Window size & Total \\
\midrule
\multicolumn{5}{l}{\textit{16-frame budget}} \\
Vanilla          & 16 & --- & --- & 16 \\
Key frames       & --- & 16 & --- & 16 \\
Key windows      & --- & 4 & 4 & 16 \\
Global with local & 8 & 2 & 4 & 16 \\
\midrule
\multicolumn{5}{l}{\textit{32-frame budget}} \\
Vanilla          & 32 & --- & --- & 32 \\
Key frames       & --- & 32 & --- & 32 \\
Key windows      & --- & 8 & 4 & 32 \\
Global with local & 16 & 4 & 4 & 32 \\
\midrule
\multicolumn{5}{l}{\textit{64-frame budget}} \\
Vanilla          & 64 & --- & --- & 64 \\
Key frames       & --- & 64 & --- & 64 \\
Key windows      & --- & 16 & 4 & 64 \\
Global with local & 32 & 8 & 4 & 64 \\
\bottomrule
\end{tabular}
\end{table}

The trained student uses the \textit{vanilla} view at inference: $B$ frames sampled uniformly over the whole video, with no access to the relevance ranking. \textit{Key frames} spends the entire budget on the $B$ highest-ranked frames, and \textit{key windows} spends it on $B/4$ disjoint four-frame windows selected by the ranking; the two views use the same ranking but allocate the budget between evidence breadth and local temporal context differently. \textit{Global with local} splits the budget equally, combining $B/2$ uniform frames with $B/8$ disjoint four-frame windows, so that it retains a coarse view of the whole video while still resolving the question-relevant moments more finely than uniform sampling would. At 32 frames this gives 16 uniform frames and four local windows; at 16 frames it gives eight uniform frames and two local windows. The window size stays fixed across budgets, so the smaller views reduce the number of selected windows while preserving the same allocation between global and local coverage.

Key frames and key windows use the same relevance ranking but allocate the frame budget differently, whereas global with local combines ranked windows with uniform coverage. All four views use the same frame budget, allowing us to compare their allocation of global and local coverage.

\subsection{Global-Temporal and Local-Evidence Questions}
\label{appendix:global-temporal-questions}

Section~\ref{sec:motivation-teacher-view} splits the evaluation into \textit{global-temporal} and \textit{local-evidence} questions. A question is global-temporal when answering it requires integrating information across separated parts of the video: the annotated evidence identifies relevant moments, but their interpretation depends on broader context, such as how an activity develops, whether a behavior recurs, or how a scene changes from beginning to end. A question is local-evidence when the answer can be determined from the annotated span and its immediate temporal neighborhood. Local-evidence questions may still require several consecutive frames, for example to distinguish the steps of a short action. The distinction concerns the temporal scope of the information needed for the answer.

We use GPT-5.6 Sol~\citep{openai2026gpt56} to assign the labels from the question, the answer options, and the annotated visual evidence with its temporal locations. The annotation protocol has two steps:

\begin{enumerate}[itemsep=2pt, leftmargin=*]
\item \textbf{Identify the required visual facts.} Specify the observable facts needed to distinguish the correct answer from the alternatives, and record the temporal location of each fact. These may include an action, an object state, an interaction, or a relation between events. A plausible answer based on language alone does not count as visual support.
\item \textbf{Determine the temporal scope.} Assign \textit{local-evidence} when all required facts are available within one coherent local event and its immediate temporal neighborhood. Assign \textit{global-temporal} when answering requires combining facts from separated episodes or making a judgment about the video as a whole. The local event must be a coherent activity; an interval drawn broadly enough to include unrelated episodes does not make the question local.
\end{enumerate}

Question labels are assigned once, before model evaluation.
We then assess the quality of the key-frame annotations separately from the local/global labels. An annotation passes this check when the selected frames support the claimed question-relevant visual facts at their recorded temporal locations and no unresolved annotation error, unclear visual evidence, or question ambiguity prevents applying the labeling rule. On random sample of 300 records, GPT-5.6 Sol judges $94.3\%$ of the annotations acceptable, and manual inspection accepts 275 of 300 annotations ($91.67\%$).
On these 300 records, the GPT and human assessments agree on $96.67\%$ of the pass/fail judgments, and Cohen's $\kappa$~\citep{cohen1960coefficient} is approximately $0.74$.

The four categories below account for the global-temporal side of the split. For each we give the reason a view restricted to the annotated evidence is insufficient, together with two examples beyond those quoted in the main text.

\begin{itemize}[itemsep=2pt, leftmargin=*]
\item \textbf{Theme and purpose.} The answer summarizes the video as a whole rather than any of its parts, so evidence frames may identify one activity without establishing that it is the dominant one. \textit{``What is the main topic of this video?''}; \textit{``Which word best describes the atmosphere of the event?''}
  \item \textbf{Event progression and temporal structure.} The answer summarizes how a multi-stage
  activity unfolds across the video, so an evidence-only view may show individual stages without
  revealing their transitions or overall organization. \textit{``What are the main stages of the
  experiment, from setting up the apparatus to reporting the result?''}; \textit{``How does the cooking
  demonstration progress from preparation to serving?''}

  \item \textbf{Recurring patterns and routines.} The answer identifies a behavior that recurs across
  multiple episodes, so one annotated window may show an instance but cannot establish the pattern or
  its consistency throughout the video. \textit{``What does the instructor consistently do before
  introducing each new exercise?''}; \textit{``Which actions does the player repeat after each failed
  attempt throughout the video?''}

\item \textbf{State evolution.} The answer is a change between two points in time, whereas a view that keeps only the most relevant window keeps only one endpoint of the comparison. \textit{``How does the colour of the liquid change during the video?''}; \textit{``What is different about the room at the end compared with the beginning?''}
\end{itemize}

\subsection{Teacher-View Comparison on Qwen3-VL Model Family}
\label{app:teacher-view-qwen3vl}

\begin{figure}[htbp]
    \centering
    \includegraphics[width=\linewidth]{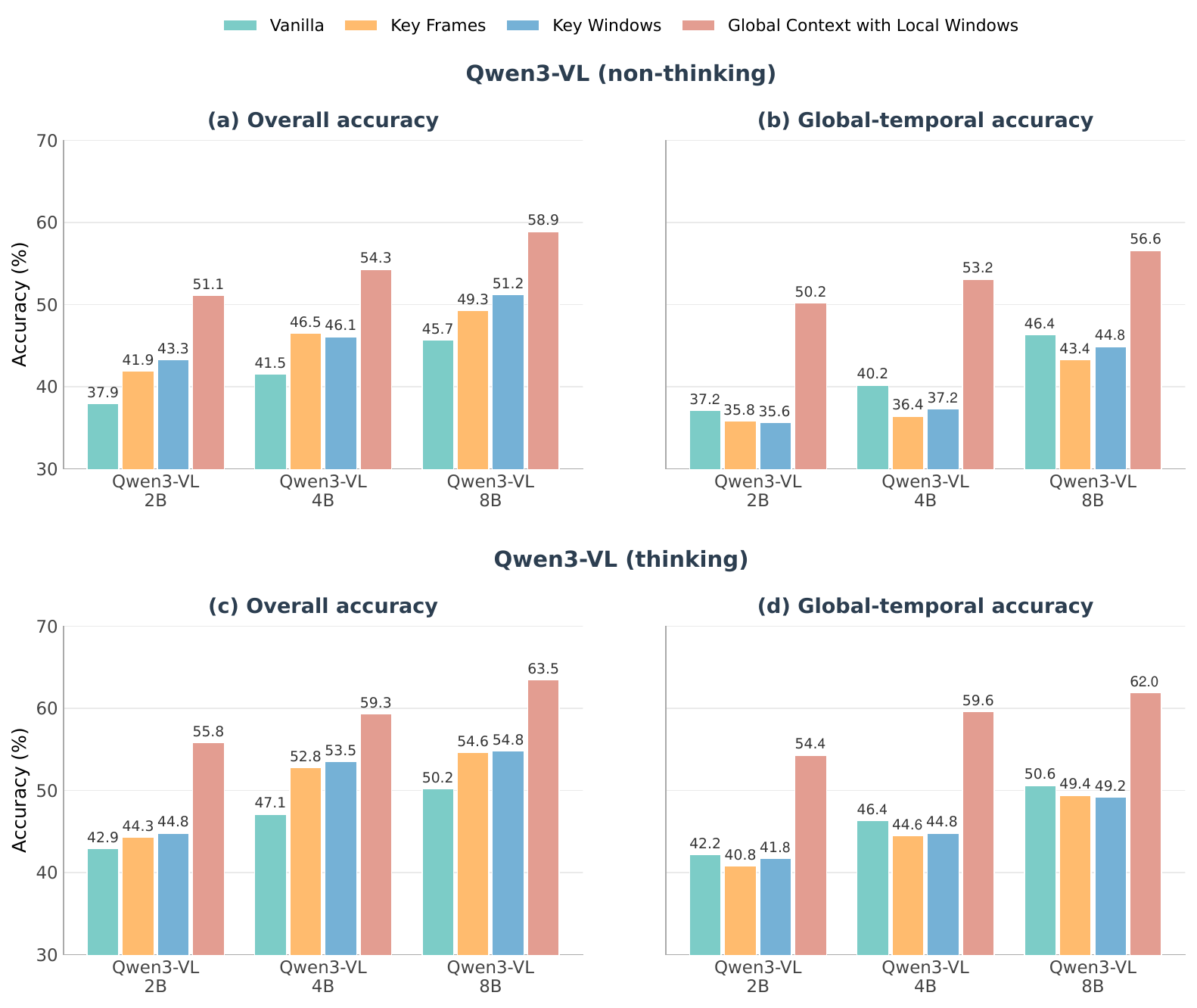}
    \caption{\textbf{Teacher-view comparison on the Qwen3-VL model family.} \textbf{(a, b)} Non-thinking variants. \textbf{(c, d)} Thinking variants. The left column reports accuracy on all questions; the right column reports accuracy on the global-temporal subset. Each panel compares the four input views defined in Table~\ref{app:tab-views} on the 2B, 4B, and 8B models.  Bars are labeled with accuracy in percent, and all four panels use the same vertical scale. In both modes and at all three model sizes, evidence-only views improve overall accuracy while reducing global-temporal accuracy relative to vanilla; global context with local windows improves both metrics.}
    \label{app:teacher-views-qwen3vl}
\end{figure}

Figure~\ref{fig:motivation-combined}(a,b) compares the four input views on Qwen3.5-0.8B, 4B, and 9B. We extend the teacher-view comparison to Qwen3-VL-2B, 4B, and 8B, considering both non-thinking and thinking variants. The views follow the definitions in Appendix~\ref{app:teacher-view-construction}, and the global-temporal subset follows the question taxonomy in Appendix~\ref{appendix:global-temporal-questions}. Figure~\ref{app:teacher-views-qwen3vl} reports the non-thinking results in the top row and the thinking results in the bottom row, with overall accuracy on the left and global-temporal accuracy on the right. Overall accuracy uses all $1{,}000$ held-out questions, and global-temporal accuracy uses the $500$-question subset. All differences discussed below are absolute percentage-point differences in accuracy; model sizes are listed in the order 2B, 4B, and 8B, respectively.

\paragraph{Non-thinking models: overall improvements can conceal a loss of global context.}
Key frames improve overall accuracy over vanilla by $4.0$, $5.0$, and $3.6$ points, while key windows improve it by $5.4$, $4.6$, and $5.5$ points. These gains show that focusing the input on question-relevant evidence helps when performance is aggregated over all questions. Neither evidence-only construction improves the global-temporal subset: key frames reduce accuracy by $1.4$, $3.8$, and $3.0$ points, and key windows reduce it by $1.6$, $3.0$, and $1.6$ points. Expanding selected frames into local windows therefore does not consistently recover the broader context needed by these questions. The ordering of key frames and key windows varies across models and question subsets.

\paragraph{Combining global coverage with local evidence improves both metrics.}
Global context with local windows reaches overall accuracies of $51.1\%$, $54.3\%$, and $58.9\%$, exceeding vanilla by $13.2$, $12.8$, and $13.2$ points. Its global-temporal accuracies are $50.2\%$, $53.2\%$, and $56.6\%$, corresponding to gains of $13.0$, $13.0$, and $10.2$ points. The advantage remains substantial when the reference is the stronger of the two evidence-only views for each model and metric: the margins are $7.8$, $7.8$, and $7.7$ points overall, and $14.4$, $16.0$, and $11.8$ points on global-temporal questions. Thus, the combined view retains the overall benefit of focused evidence while avoiding the global-temporal degradation observed with either evidence-only view.

\paragraph{Thinking models show the same trade-off across views.}
For the thinking variants, global context with local windows obtains overall accuracies of $55.8\%$, $59.3\%$, and $63.5\%$, compared with vanilla accuracies of $42.9\%$, $47.1\%$, and $50.2\%$. The gains are $12.9$, $12.2$, and $13.3$ points. Key frames reach $44.3\%$, $52.8\%$, and $54.6\%$, and key windows reach $44.8\%$, $53.5\%$, and $54.8\%$. Although key windows are the stronger evidence-only view on overall accuracy for all three thinking models, the combined view still improves over them by $11.0$, $5.8$, and $8.7$ points. The overall advantage of combining broad temporal coverage with focused local evidence therefore persists across both evaluated reasoning modes.

The global-temporal results reveal the same limitation of evidence-only inputs in thinking mode. Relative to vanilla accuracies of $42.2\%$, $46.4\%$, and $50.6\%$, key frames reduce accuracy by $1.4$, $1.8$, and $1.2$ points, and key windows reduce it by $0.4$, $1.6$, and $1.4$ points. Global context with local windows instead reaches $54.4\%$, $59.6\%$, and $62.0\%$, improving over vanilla by $12.2$, $13.2$, and $11.4$ points. Its margins over the stronger evidence-only view are $12.6$, $14.8$, and $12.6$ points. Thus, the overall gains of evidence-only views continue to coexist with a loss on global-temporal questions in the thinking variants, whereas the combined view improves both metrics at every evaluated scale.

\paragraph{Implications for teacher-view design.}
Both Qwen3-VL modes reproduce the qualitative trade-off observed for Qwen3.5: selecting relevant evidence can improve the aggregate score while reducing accuracy on questions about the video as a whole. Across all six Qwen3-VL model--mode configurations, global context with local windows is the only tested alternative to vanilla that improves both overall and global-temporal accuracy, and it achieves the highest accuracy on both metrics. The deployed \method student continues to use uniformly sampled frames without evidence annotations.

\subsection{Measuring the Concentration of the Evidence Gap}
\label{app:ecr-protocol}

For each question we sample one response from the student under its own view $v$, keep the responses the verifier accepts, and re-score exactly those token sequences under the teacher view $\tilde{v}$ in a single forward pass. We restrict the diagnostic to verifier-accepted responses and compare their token likelihoods under the two views. This yields a per-token log-probability difference
\begin{equation}
\label{eq:delta-logp}
\Delta \log p_t
= \log \pi_{\mathrm{teacher}}(y_t \mid q, \tilde{v}, y_{<t})
- \log \pi_{\mathrm{student}}(y_t \mid q, v, y_{<t}),
\end{equation}
for the same question, response token, and prefix. In the static diagnostic of Figure~\ref{fig:motivation-combined}(c), both policies use the common initial checkpoint $\theta_0$, so $\Delta \log p_t$ isolates the effect of changing the visual input at position $t$. This equality of weights applies to the initial diagnostic; during training, the student evolves while the teacher remains fixed.

Within each response we sort tokens by $|\Delta \log p_t|$ in decreasing order and accumulate their share of $\sum_t |\Delta \log p_t|$. We normalize each response separately before averaging the concentration curves. Figure~\ref{fig:motivation-combined} (c) in Section~\ref{sec:motivation} plots the mean curve over the evaluation set against the \textit{uniform baseline}, the diagonal that a response would trace if every one of its tokens contributed equally to the gap. The shaded band spans the 10th to 90th percentile of the individual per-video curves at each token fraction, so its width reports how much the concentration varies from video to video rather than any uncertainty in the mean. The reported statistic is the token fraction at which the mean curve reaches $50\%$.

Section~\ref{sec:cs} tracks a scalar variant of this measurement during training: the share of $\sum_t |\Delta\log p_t|$ carried by the top decile of tokens within each response. Here the student distribution comes from the evaluated training checkpoint under $v$, and the teacher distribution comes from the fixed initial checkpoint $\theta_0$ under $\tilde v$. 

\subsection{Measuring Output and Hidden-State Alignment}
\label{app:ssa-protocol}

Figure~\ref{fig:motivation-combined}(d) measures whether output-level alignment also brings the student and teacher hidden states closer. We run on-policy self-distillation from Equation~\ref{eq:opsd}: teacher and student are initialized from the same checkpoint $\theta_0$, the teacher remains fixed, and responses are sampled from the evolving student under $v$. Only the student is trained, using the token-level KL divergence to the fixed teacher under $\tilde{v}$. There is no reward term or representation-level loss in this diagnostic. We train for 300 steps and evaluate both quantities at fixed intervals on the held-out split described in Appendix~\ref{app:motivated-protocol}.

For the output measurement, we score each sampled response with the current student under $v$ and the fixed teacher under $\tilde{v}$, holding the question, generated tokens, and prefix at each response position identical. We report the KL divergence from the student distribution to the teacher distribution, averaged over valid response positions and then over the evaluation batch. A smaller value means that the student more closely matches the fixed teacher's next-token distributions on its own responses. 

To measure the corresponding hidden-state alignment, we use centered kernel alignment (CKA)~\citep{kornblith2019similarity}. CKA compares the similarity structure of response-end representations across matched responses: it asks whether pairs of responses that are similar under one view are also similar under the other.

For decoder layer $\ell$, let $X_\ell,Y_\ell\in\mathbb{R}^{m\times d_\ell}$ be the response-end hidden states for $m$ matched responses, computed by the current student under $v$ and the fixed teacher under $\tilde v$, respectively. As in Section~\ref{sec:method-ssa}, we take the final valid response token after layer $\ell$, so both states refer to the same generated text position. After centering the columns of $X_\ell$ and $Y_\ell$, we form $K_\ell=X_\ell^c(X_\ell^c)^\top$ and $L_\ell=Y_\ell^c(Y_\ell^c)^\top$. Linear CKA is the normalized similarity between these two Gram matrices:
\begin{equation}
\mathrm{CKA}(X_\ell,Y_\ell)
=\frac{\langle K_\ell,L_\ell\rangle_F}
{\|K_\ell\|_F\,\|L_\ell\|_F},
\qquad
\langle K_\ell,L_\ell\rangle_F
=\sum_{i=1}^{m}\sum_{j=1}^{m}(K_\ell)_{ij}(L_\ell)_{ij}.
\label{eq:app-linear-cka}
\end{equation}
where $\langle\cdot,\cdot\rangle_F$ is the Frobenius inner product. CKA is bounded between $0$ and $1$ for the centered Gram matrices used here. A value of $1$ means that the two views induce the same pairwise similarity structure up to scale; a value near $0$ means that the structures are weakly aligned. The measure is invariant to rotations and isotropic rescaling of the hidden coordinates, so it captures the organization of the representations rather than their arbitrary basis or norm.

We plot $1-\mathrm{CKA}$, averaged over the evaluated decoder layers:
\begin{equation}
D_{\mathrm{hidden}}
=\frac{1}{|\mathcal{S}|}\sum_{\ell\in\mathcal{S}}
\left(1-\mathrm{CKA}(X_\ell,Y_\ell)\right).
\label{eq:app-hidden-divergence}
\end{equation}
Each layer's CKA is computed across the batch before the layer average. The SSA loss in Equation~\ref{eq:ssa} instead uses cosine similarity between paired response-end states.

The output KL and $1-\mathrm{CKA}$ use the same matched responses but measure different quantities and are not on the same scale. In Figure~\ref{fig:motivation-combined}(d), output KL decreases to $0.046$, while $1-\mathrm{CKA}$ remains at $0.386$. Thus, output-only matching reduces the next-token divergence while leaving a distinct hidden-state similarity structure; SSA aligns the response-end states at matched depths.

\section{Experimental Data and Supplementary Experiments}
\label{app:data-and-supplementary}

This section details the construction and composition of Video-Evidence-96K, including the global-temporal and local-evidence questions defined in Appendix~\ref{appendix:global-temporal-questions}. We then describe the baselines, benchmarks, and training settings, followed by the full results across model families and frame budgets and the corresponding component ablations.

\subsection{Construction of Video-Evidence-96K}
\label{app:video-evidence-96k}

 We construct Video-Evidence-96K in four stages: source aggregation, record validation, question-conditioned evidence annotation, and view assembly. It retains 96,705 question--video records, of which $1{,}000$ are held out for diagnostic evaluation and 95,705 are used for training. Table~\ref{app:tab-data-pipeline} tracks these counts through validation and the held-out split, Table~\ref{app:tab-data-stats} reports the training sources, and Table~\ref{app:tab-question-composition} reports the temporal scope of the questions in each split.

\paragraph{Source aggregation.}
We start from 104,257 video--question--answer records collected from four public sources: LLaVA-Video-178K~\citep{zhang2024llava}, NExT-QA~\citep{xiao2021next}, Perception~\citep{patraucean2023perception}, and STAR~\citep{wu2024starbenchmarksituatedreasoning}. The sources cover complementary aspects of video understanding: LLaVA-Video-178K contributes broad-domain footage with diverse question types, NExT-QA emphasizes causal and temporal questions, Perception stresses fine-grained perceptual skills, and STAR provides situated-reasoning questions over short action-centric clips. Every record is normalized into a shared schema consisting of the video path, the question, lettered multiple-choice options, and a single correct option letter, so that the rule-based verifier of Section~\ref{sec:pre} applies uniformly across sources. The unit of data is a question--video pair: questions referring to the same video remain separate records because they can require different evidence and different temporal scopes.

\paragraph{Record validation.}
We then apply four checks in order. (i) The question and answer fields must be non-empty after whitespace normalization. (ii) The options must be well formed: unique option labels, non-duplicate option texts, and exactly one option matching the reference answer. (iii) The video file must exist and decode end to end. (iv) The video must supply sufficient temporal candidates for evidence sampling: after candidate-pool extraction (below), a record is kept only when its pool can support the disjoint local windows and the global track of the largest teacher configuration. These checks remove 7,552 records in total, leaving 96,705 valid records. The subsequent held-out split reserves $1{,}000$ of these records for evaluation; these are valid examples withheld from training.

\begin{table}[h]
\centering
\caption{Construction pipeline of Video-Evidence-96K. Validation rows report excluded records and the number remaining after each check. The final row reserves $1{,}000$ valid records for diagnostic evaluation, leaving 95,705 training records.}
\label{app:tab-data-pipeline}
\begin{tabular}{llrr}
\toprule
Stage & Criterion & Excluded / reserved & Remaining \\
\midrule
Source aggregation & four public sources & --- & 104,257 \\
Field validation & non-empty question and answer & 812 & 103,445 \\
Option validation & unique labels, single correct option & 1,134 & 102,311 \\
Video validation & file exists and decodes end to end & 1,313 & 100,998 \\
Temporal validation & sufficient evidence candidates & 4,293 & 96,705 \\
Held-out diagnostics & evaluation split of Section~\ref{sec:motivation} & 1,000 & 95,705 \\
\bottomrule
\end{tabular}
\end{table}

\paragraph{Question-conditioned evidence annotation.}
For each retained record we extract a candidate pool of up to $512$ frames sampled uniformly over the full video; shorter videos contribute every decodable frame. An automatic question-conditioned frame-relevance annotator~\citep{videoitg} scores each candidate frame against the question, and we store the full relevance ranking with the record rather than a hard top-$k$ cut. Storing the ranking allows the four diagnostic views of Appendix~\ref{app:teacher-view-construction} and the training views at every student budget $N\in\{16,32,64\}$ to use the same underlying evidence annotation. Since the scores are conditioned on the question, two records from the same video can select different moments. The ranking identifies where relevant evidence occurs; the local windows and global track determine how that evidence is presented to the teacher.

\paragraph{Evidence quality.}
We use the annotation audit described in Appendix~\ref{appendix:global-temporal-questions}. On the same 300 randomly sampled records, GPT-5.6 Sol accepts $94.3\%$ of the evidence annotations and human inspection accepts 275 of 300 ($91.67\%$). An annotation is accepted when the selected frames support the question-relevant visual facts at the recorded temporal locations and no unresolved ambiguity prevents applying the labeling rule.

\paragraph{View assembly.}
At training time, the teacher view combines a global track of $N/2$ uniformly sampled frames with $N/8$ disjoint local windows of four frames centered on the top-ranked candidates, following the window construction of Appendix~\ref{app:teacher-view-construction}. Windows are expanded in rank order, a window that would overlap an earlier local window is skipped, and the selected frames are arranged in chronological order before encoding. The local windows contribute $N/2$ frames, giving the teacher the same total frame budget $N$ as the student: 16, 32, or 64 frames. The global track provides coverage of the video as a whole, while the local windows resolve the moments most relevant to the question. The student receives only $N$ uniformly sampled frames and does not access the relevance ranking at training or inference.

\paragraph{Dataset splits and source composition.}
From the 96,705 validated records we hold out $1{,}000$ for the diagnostic experiments of Section~\ref{sec:motivation}, leaving 95,705 training records, with approximately 4.4 questions per video. Table~\ref{app:tab-data-stats} counts training records after the held-out examples have been removed. Its source percentages use the 95,705-record training split as the denominator; they describe the distribution of question--video pairs rather than the distribution of unique videos.

\begin{table}[h]
\centering
\caption{Composition of the Video-Evidence-96K training split.}
\label{app:tab-data-stats}
\begin{tabular}{lrr}
\toprule
Source & Records & Share \\
\midrule
LLaVA-Video-178K & 70,468 & 73.63\% \\
NExT-QA & 7,542 & 7.88\% \\
Perception & 6,345 & 6.63\% \\
STAR & 11,350 & 11.86\% \\
\midrule
Total & 95,705 & 100.00\% \\
\bottomrule
\end{tabular}
\end{table}

\paragraph{Global-temporal and local-evidence composition.}
We use the question taxonomy in Appendix~\ref{appendix:global-temporal-questions} to describe the temporal scope of the dataset. GPT-5.6 Sol assigns the labels from the question, answer options, and annotated visual evidence with its temporal locations, first identifying the facts needed to distinguish the answer and then determining how widely those facts are distributed in time. A local-evidence question can be answered within one coherent event and its immediate temporal neighborhood, even when several consecutive frames are needed. A global-temporal question requires combining separated episodes or making a judgment about the video as a whole. Video duration and source-dataset identity do not determine the label: different questions about the same video can fall into different subsets.

The global-temporal subset includes questions about theme and purpose, event progression, recurring patterns, and state evolution. The local-evidence subset covers questions whose relevant action, object state, or interaction can be resolved within one selected event and its surroundings.

Table~\ref{app:tab-question-composition} separates the composition of the training set from that of the held-out diagnostic set. Each count refers to question--video records, with the share computed within the corresponding split. Thus, the global-temporal share in training is the number of global-temporal training records divided by 95,705, whereas its share in the diagnostic set uses $1{,}000$ as the denominator. The complementary local-evidence share is computed over the same records. Reporting both splits specifies how much global-temporal supervision is available during training and how much weight that subset carries in the overall diagnostic accuracy.

\begin{table}[h]
\centering
\caption{Composition of Video-Evidence-96K by question type, following Appendix~\ref{appendix:global-temporal-questions}. Counts are question--video records, and each percentage uses the total of its own split as the denominator.}
\label{app:tab-question-composition}
\begin{tabular}{lrrrrr}
\toprule
 & & \multicolumn{2}{c}{Global-temporal} & \multicolumn{2}{c}{Local-evidence} \\
\cmidrule(lr){3-4}\cmidrule(lr){5-6}
Split & Total & Records & Percentage & Records & Percentage \\
\midrule
Training & 95,705 & {20,137} & {21.04\%} & {75,568} & {78.96\%} \\
Held-out diagnostics & 1,000 & {500} & {50\%} & {500} & {50\%} \\
\bottomrule
\end{tabular}
\end{table}

The same held-out question labels define the subsets throughout the teacher-view comparisons and the post-training analysis in Appendix~\ref{app:global-temporal-after}. Overall accuracy is weighted by the number of questions in each subset, while global-temporal and local-evidence accuracies are calculated within their respective subsets. This is why an evidence-only view can improve overall accuracy while reducing global-temporal accuracy: gains on local-evidence questions can outweigh losses on the global-temporal subset. Reporting the subset sizes alongside both accuracies makes the contribution of each question type explicit.

\subsection{Baselines and Benchmarks}
\label{app:baselines-benchmarks}

\subsubsection{Baselines}

Within each model and mode, all post-trained methods start from the same released checkpoint and use Video-Evidence-96K as the common source of question--video pairs. GRPO, OPSD, and \method use the same training split, optimizer settings, rollout budget, and number of updates. SFT instead uses traces generated for these question--video pairs, as described below.

\begin{itemize}[itemsep=2pt, leftmargin=*]
\item \textbf{Vanilla} is the released checkpoint with no post-training. It fixes the starting point every other row is measured against.
\item \textbf{SFT} is supervised fine-tuning on chain-of-thought traces generated by Qwen3.5-397B-A17B for the training question--video pairs. It is the off-policy counterpart of the methods below: the supervision is dense and comes from a far larger model, but it is collected once rather than sampled from the evolving student policy.
\item \textbf{GRPO} is the verifier-driven RL objective of Equation~\ref{eq:rl}, sampled from the student view alone.
\item \textbf{OPSD} is on-policy self-distillation as described in Equation~\ref{eq:opsd}. It keeps the student's own rollouts and the same fixed teacher checkpoint and teacher view as \method, but replaces the reward with the token-level divergence between the student and teacher next-token distributions.
\end{itemize}

\subsubsection{Benchmarks}

We evaluate on five multiple-choice video benchmarks, all read with uniform frame sampling at the configuration's budget $N$ and with no evidence annotation. These benchmarks assess transfer beyond the held-out Video-Evidence-96K diagnostics. A prediction is correct when the predicted option letter matches the reference, and \textbf{Avg.} is the unweighted macro-average over the five benchmark scores. The global-temporal/local-evidence labels describe the Video-Evidence-96K diagnostic subsets; they are distinct from benchmark-specific divisions such as Video-MME duration and Video-MME~2 question groups.

\begin{itemize}[itemsep=2pt, leftmargin=*]
\item \textbf{Video-MME}~\citep{fu2025video} provides a broad evaluation of video understanding across diverse content domains and temporal spans. Its questions cover perception and reasoning, while its short, medium, and long duration splits examine how well models handle increasingly extended video context. This combination makes it useful for assessing whether post-training gains extend beyond a particular scene type or temporal scale. Under our fixed input-frame budgets, the duration breakdown also shows how the gains vary as the sampled frames must cover longer videos. We report overall accuracy and the three duration splits, with the Qwen3.5 breakdown in Tables~\ref{tab:qwen35-duration-0.8B}--\ref{tab:qwen35-duration-27B}.
\item \textbf{MVBench}~\citep{li2024mvbench} targets temporal understanding through multiple-choice tasks that require information from more than one frame, spanning temporal perception and reasoning about actions, objects, and their changes. It complements the broader domain and duration coverage of Video-MME by testing how models use sequences of observations.
\item \textbf{LongVideoBench}~\citep{wu2024longvideobench} was introduced to evaluate referring reasoning in long, interleaved video--language contexts. A question refers to a particular part of the video, and the model must locate the relevant content within a much longer context before answering. The tasks range from perceiving details at the referred moment to reasoning about relations between events, making both evidence localization and contextual interpretation important. Its value for our evaluation is that a model must retain question-relevant detail despite substantial surrounding content; a coarse account of the video's overall topic is insufficient. We evaluate on the validation split using the uniform frame-sampling protocol described above.
\item \textbf{Video-MME~2}~\citep{fu2026videommev2} extends video evaluation to visual information aggregation, temporal dynamics, and complex reasoning. Its grouped-question design examines whether a model's understanding remains coherent across related questions, adding a dimension that isolated question answering does not capture. The \textit{logic} groups probe successive steps in a reasoning chain, while the \textit{relevance} groups probe different aspects or levels of detail of the same capability. These demands complement the duration-based analysis of Video-MME: the model must combine observations and reason about their relationships, in addition to recognizing individual events. We report the two subsets separately, using the scoring convention specified below.
\item \textbf{MLVU}~\citep{zhou2025mlvu} was introduced to evaluate long-video understanding across diverse video genres and complementary task types. Its task design distinguishes holistic understanding, retrieval of a single relevant detail, and integration of multiple details. The multiple-choice tasks include topic reasoning, retrieval of a brief event from a long video, and reasoning about action order or repetition. These tasks expose different requirements: global understanding depends on broad temporal coverage, detail retrieval depends on preserving a localized observation, and action ordering or counting requires combining observations over time. We evaluate the multiple-choice portion, which tests both global context use and access to local evidence within a common answer format.
\end{itemize}

For Video-MME~2, we report per-question accuracy in the \textit{logic} and \textit{relevance} subsets. The combined score weights the two subset accuracies by their question counts, and Tables~\ref{tab:qwen35-full-16}--\ref{tab:qwen35-full-64} report the Qwen3.5 subset results separately. For consistency and reproducibility, we use the official evaluation prompts for each benchmark, respectively.

\subsection{Training Hyper-Parameters}
\label{app:training-hparams}

Table~\ref{app:tab-hparams} reports the training and evaluation settings for GRPO, OPSD, and \method. Following Section~\ref{sec:set_up}, each run uses $1{,}000$ training steps, with the optimizer settings, prompt and rollout counts, and rollout decoding held fixed within a model--mode--budget configuration. The two \method-specific settings are the evidence-contrast clipping threshold $c$ of Equation~\ref{eq:ecr-profile} and the alignment weight $\lambda_{\mathrm{SSA}}$ of Equation~\ref{eq:owlet-objective}. Verifier rewards and group-normalized advantages apply to GRPO and \method; OPSD optimizes the token-level distillation objective without a reward term.

\begin{table}[h]
\centering
\caption{Training and evaluation hyper-parameters.}
\label{app:tab-hparams}
\setlength{\tabcolsep}{6pt}
\small
\begin{tabular}{ll}
\toprule
Setting & Value \\
\midrule
\multicolumn{2}{l}{\textit{Optimization (shared by GRPO, OPSD, and \method)}} \\
Learning rate & $1\times 10^{-6}$, constant, no warm-up \\
Prompts per step / rollouts per prompt $G$ & $16$ / $8$ \\
Mini-batch size & $16$  \\
Training steps & $1,000$ \\
Clipping range $\epsilon$ & $0.2$ \\
KL penalty to a reference policy & none \\
Advantage estimator (GRPO and \method) & group-normalized (GRPO) \\
Reward (GRPO and \method) & rule-based match on the option letter \\
Devices & $8$ NVIDIA H200 GPUs (140GB each) \\
\midrule
\multicolumn{2}{l}{\textit{Rollout decoding}} \\
Temperature / top-$p$ / top-$k$ & $1.0$ / $0.99$ / disabled \\
Max response length & $2048$ tokens \\
\midrule
\multicolumn{2}{l}{\textit{Views}} \\
Student frames $N$ & $16$, $32$, or $64$ (uniform) \\
Teacher global track & $N/2$ uniform frames \\
Teacher local windows & $N/8$ disjoint windows of $4$ frames \\
\midrule
\multicolumn{2}{l}{\textit{Teacher parameters (OPSD and \method)}} \\
Default teacher & fixed at the common initial checkpoint $\theta_0$ \\
\midrule
\multicolumn{2}{l}{\textit{\method-specific}} \\
ECR clipping threshold $c$ & $5$ \\
ECR applied to & positively-advantaged responses \\
SSA weight $\lambda_{\mathrm{SSA}}$ & $0.05$ \\
SSA responses & verifier-accepted only \\
\midrule
\multicolumn{2}{l}{\textit{Evaluation decoding}} \\
Temperature  & $0.0$ \\
Max new tokens & $2048$ \\
Frames & $N$ uniform, no evidence annotation \\
\bottomrule
\end{tabular}
\end{table}

\subsection{Per-Benchmark Results and Detailed Analysis}
\label{app:main-results-analysis}

Tables~\ref{tab:qwen35-full-16}--\ref{tab:qwen35-full-64} give the per-benchmark results for Qwen3.5-0.8B, 4B, 9B, and 27B in non-thinking mode at student frame budgets of 16, 32, and 64. They extend Table~\ref{tab:main} to the smaller frame budgets and report the two Video-MME~2 question groups separately. Tables~\ref{tab:qwen35-duration-0.8B}--\ref{tab:qwen35-duration-27B} further break the Video-MME results down into short, medium, and long videos, with one table per model. Together, these results show how the comparison changes across benchmark tasks, temporal spans, input budgets, and model scales.

\paragraph{Table configuration.}
The per-benchmark tables group rows by backbone and compare the base model, SFT, GRPO, OPSD, and \method at a fixed frame budget. The duration tables instead fix the backbone and group rows by frame budget. All scores are accuracies in percent. Within Video-MME~2, \textit{Logic} and \textit{Rel.} denote the logic and relevance question subsets, and \textit{All} combines their accuracies using the respective question counts. In the Video-MME duration tables, \textit{All} averages the short, medium, and long splits. The \textbf{Avg.} column weights the five benchmarks equally, with Video-MME~2 contributing its combined score once. Boldface identifies the best score in each metric column within a model and frame budget. Scores are displayed to one decimal place, and improvements below are expressed in percentage points. 

\paragraph{Comparison with post-training baselines.}
Across the twelve model--budget configurations, \method achieves the highest accuracy on all five benchmarks and the highest macro-average. The strongest baseline by macro-average varies between GRPO and OPSD, while \method improves on it by 2.2--4.3 points. At 64 frames, the margins are 3.4, 4.1, 4.3, and 4.3 points for the 0.8B, 4B, 9B, and 27B models, respectively. Thus, the advantage persists across model scales and is not limited to the smallest backbones. The comparison with GRPO supports the value of supplementing verifiable rewards with supervision from the evidence-privileged view. The comparison with OPSD, which uses the same teacher view, further supports transferring that supervision through evidence-aware credit and hidden-state alignment. The separate contributions of ECR and SSA are examined in Section~\ref{sec:ab} and Appendix~\ref{app:ablation-results}.

\paragraph{Performance across video durations.}
The Video-MME duration tables show improvements over the baselines on short, medium, and long videos at every evaluated model size and frame budget. At 27B with 64 frames, \method reaches 87.8, 74.8, and 65.8 on the three splits, exceeding the strongest baseline in each split by 2.7, 4.0, and 5.2 points, respectively. The larger margin on long videos in this configuration is consistent with the role of the global track: the teacher retains temporal context around the question-relevant evidence even when the video spans many events. Improvements on the short split also show that the benefit extends to videos with a smaller temporal span, where local transitions can still determine the answer. Across the duration tables, the size of the gain varies with the backbone and frame budget, while the improvement over the corresponding baselines remains consistent.

\paragraph{Temporal understanding and long-video reasoning.}
The gains also extend across MVBench, LongVideoBench, and MLVU, which place different demands on the use of temporal evidence. At 4B with 64 frames, \method scores 72.2 on MVBench, 63.4 on LongVideoBench, and 58.8 on MLVU, improving over the strongest baseline for each benchmark by 3.3, 3.6, and 7.7 points. MLVU also shows a substantial gain at 9B with 64 frames: \method reaches 61.6, compared with 53.7 for the strongest baseline. These results cover reasoning about temporal changes, locating and interpreting referred content in a long video, and combining global understanding with access to local details. The gains across these complementary tasks are consistent with the teacher view's combination of global coverage and question-conditioned local windows.

\paragraph{Logic and relevance in Video-MME~2.}
\method improves on the strongest baseline in both question subsets for all twelve model--budget configurations. At 27B with 64 frames, logic accuracy increases to 37.6 from a strongest-baseline score of 35.2, while relevance accuracy increases to 32.9 from 30.4. The corresponding improvements are 2.4 and 2.5 points, so the increase in the combined score is supported by both subsets. The margins need not be equally large: at 9B with 64 frames, the gains are 0.5 points on logic and 2.5 points on relevance.

\paragraph{Robustness across frame budgets.}
Table~\ref{tab:budget} summarizes macro-average accuracy across input budgets. For each model size, the macro-average of \method increases as the student budget grows from 16 to 32 and then to 64 frames, and it leads the baselines at every budget. The 27B model reaches 57.9, 60.4, and 64.5, exceeding the strongest baseline at the corresponding budget by 2.2, 2.3, and 4.3 points. More input frames and evidence-guided post-training therefore provide complementary gains in these configurations. Across the four model sizes, \method also exceeds the strongest baseline using twice as many frames in seven of eight adjacent-budget comparisons: 16 against 32 frames and 32 against 64 frames for each model. The exception is 27B at 16 frames, which scores 57.9 against 58.1 for the strongest 32-frame baseline. At 32 frames, the same model exceeds the strongest 64-frame baseline, 60.4 versus 60.2. These comparisons under uniform sampling suggest that \method learns to make more effective use of the student's available observations.

\paragraph{Comparison across model scales.}
The benefits also affect comparisons between backbones. Qwen3.5-4B with \method scores 53.3, 54.6, and 58.2 at 16, 32, and 64 frames, respectively. These scores exceed the strongest Qwen3.5-9B baseline at the same budgets, which reaches 50.5, 52.9, and 55.5, giving margins of 2.8, 1.7, and 2.7 points. At the same time, \method continues to improve the 9B and 27B models over their own baselines. The results thus support evidence-guided post-training both as a way to strengthen smaller models and as a complementary improvement for larger backbones.

\begin{table}[t]
\centering
\caption{Accuracy (\%) for Qwen3.5 in non-thinking mode at 16 input frames, with the two Video-MME~2 question groups reported separately. \textit{Logic} and \textit{Rel.} are the group types and \textit{All} combines them by question count. We \textbf{bold} the best result for each backbone and benchmark.}
\label{tab:qwen35-full-16}
\resizebox{\textwidth}{!}{
\setlength{\tabcolsep}{4pt}
\renewcommand{\arraystretch}{1.1}
\begin{tabular}{llcccccccc}
\toprule
\textbf{Backbone Model} & \textbf{Method} & \textbf{Video-MME} & \textbf{MVBench} & \textbf{LongVideoBench} & \multicolumn{3}{c}{\textbf{Video-MME 2}} & \textbf{MLVU} & \textbf{Avg.} \\
\cmidrule(lr){6-8}
 & & & & & Logic & Rel. & All & & \\
\midrule
\multirow{5}{*}{Qwen3.5-0.8B} & Vanilla & 45.1 & 49.4 & 44.9 & 17.8 & 14.4 & 15.6 & 34.2 & 37.8 \\
 & SFT & 45.5 & 50.8 & 45.9 & 18.5 & 14.8 & 16.1 & 34.7 & 38.6 \\
 & GRPO & 46.8 & 50.7 & 47.0 & 19.6 & 16.3 & 17.5 & 35.8 & 39.6 \\
 & OPSD & 46.2 & 50.4 & 47.5 & 18.2 & 14.9 & 16.1 & 35.1 & 39.1 \\
 & \green \method & \green \textbf{50.3} & \green \textbf{56.4} & \green \textbf{50.6} & \green \textbf{20.5} & \green \textbf{17.6} & \green \textbf{18.6} & \green \textbf{38.9} & \green \textbf{42.9} \\
\midrule
\multirow{5}{*}{Qwen3.5-4B} & Vanilla & 57.2 & 64.8 & 53.7 & 24.3 & 19.9 & 21.5 & 42.7 & 48.0 \\
 & SFT & 58.3 & 64.9 & 54.3 & 24.3 & 20.8 & 22.0 & 43.0 & 48.5 \\
 & GRPO & 58.8 & 67.6 & 56.3 & 25.7 & 20.8 & 22.5 & 43.7 & 49.8 \\
 & OPSD & 59.0 & 65.5 & 54.2 & 25.4 & 21.1 & 22.6 & 44.1 & 49.1 \\
 & \green \method & \green \textbf{62.8} & \green \textbf{69.2} & \green \textbf{59.5} & \green \textbf{27.3} & \green \textbf{22.5} & \green \textbf{24.2} & \green \textbf{50.8} & \green \textbf{53.3} \\
\midrule
\multirow{5}{*}{Qwen3.5-9B} & Vanilla & 58.2 & 63.4 & 56.7 & 26.7 & 20.0 & 22.3 & 44.3 & 49.0 \\
 & SFT & 58.9 & 64.6 & 57.6 & 27.4 & 20.8 & 23.1 & 44.6 & 49.8 \\
 & GRPO & 60.2 & 64.2 & 58.5 & 29.0 & 21.0 & 23.8 & 45.6 & 50.5 \\
 & OPSD & 60.0 & 64.7 & 57.7 & 28.9 & 21.2 & 23.9 & 45.4 & 50.3 \\
 & \green \method & \green \textbf{62.2} & \green \textbf{68.2} & \green \textbf{62.4} & \green \textbf{32.9} & \green \textbf{25.1} & \green \textbf{27.8} & \green \textbf{50.6} & \green \textbf{54.2} \\
\midrule
\multirow{5}{*}{Qwen3.5-27B} & Vanilla & 64.8 & 66.4 & 60.7 & 30.5 & 25.6 & 27.3 & 52.5 & 54.3 \\
 & SFT & 64.7 & 66.5 & 61.6 & 31.3 & 26.2 & 28.0 & 53.3 & 54.8 \\
 & GRPO & 66.0 & 68.9 & 62.2 & 30.9 & 26.0 & 27.7 & 53.8 & 55.7 \\
 & OPSD & 65.8 & 68.8 & 61.6 & 32.6 & 26.7 & 28.8 & 53.3 & 55.7 \\
 & \green \method & \green \textbf{68.2} & \green \textbf{73.6} & \green \textbf{63.2} & \green \textbf{32.9} & \green \textbf{27.7} & \green \textbf{29.6} & \green \textbf{55.0} & \green \textbf{57.9} \\
\bottomrule
\end{tabular}
}
\end{table}

\begin{table}[t]
\centering
\caption{Accuracy (\%) for Qwen3.5 in non-thinking mode at 32 input frames, with the two Video-MME~2 question groups reported separately. \textit{Logic} and \textit{Rel.} are the group types and \textit{All} combines them by question count. We \textbf{bold} the best result for each backbone and benchmark.}
\label{tab:qwen35-full-32}
\resizebox{\textwidth}{!}{
\setlength{\tabcolsep}{4pt}
\renewcommand{\arraystretch}{1.1}
\begin{tabular}{llcccccccc}
\toprule
\textbf{Backbone Model} & \textbf{Method} & \textbf{Video-MME} & \textbf{MVBench} & \textbf{LongVideoBench} & \multicolumn{3}{c}{\textbf{Video-MME 2}} & \textbf{MLVU} & \textbf{Avg.} \\
\cmidrule(lr){6-8}
 & & & & & Logic & Rel. & All & & \\
\midrule
\multirow{5}{*}{Qwen3.5-0.8B} & Vanilla & 48.2 & 51.8 & 46.4 & 17.7 & 14.7 & 15.8 & 38.1 & 40.1 \\
 & SFT & 48.8 & 52.4 & 46.9 & 19.5 & 15.5 & 16.9 & 39.4 & 40.9 \\
 & GRPO & 50.1 & 54.8 & 48.2 & 19.5 & 16.4 & 17.5 & 38.9 & 41.9 \\
 & OPSD & 49.6 & 53.4 & 48.0 & 19.5 & 15.5 & 16.9 & 39.7 & 41.5 \\
 & \green \method & \green \textbf{53.9} & \green \textbf{57.8} & \green \textbf{53.0} & \green \textbf{22.2} & \green \textbf{18.0} & \green \textbf{19.4} & \green \textbf{42.0} & \green \textbf{45.2} \\
\midrule
\multirow{5}{*}{Qwen3.5-4B} & Vanilla & 61.4 & 65.4 & 55.1 & 24.9 & 20.1 & 21.8 & 45.3 & 49.8 \\
 & SFT & 62.6 & 66.7 & 56.5 & 25.9 & 20.2 & 22.2 & 45.5 & 50.7 \\
 & GRPO & 63.2 & 67.6 & 58.0 & 26.0 & 21.8 & 23.2 & 46.9 & 51.8 \\
 & OPSD & 62.9 & 66.6 & 56.1 & 25.7 & 20.5 & 22.3 & 47.0 & 51.0 \\
 & \green \method & \green \textbf{67.0} & \green \textbf{68.7} & \green \textbf{61.4} & \green \textbf{28.7} & \green \textbf{22.7} & \green \textbf{24.8} & \green \textbf{51.2} & \green \textbf{54.6} \\
\midrule
\multirow{5}{*}{Qwen3.5-9B} & Vanilla & 62.0 & 64.4 & 58.7 & 30.6 & 22.0 & 25.0 & 46.5 & 51.3 \\
 & SFT & 62.6 & 65.6 & 60.0 & 32.3 & 23.2 & 26.4 & 46.6 & 52.2 \\
 & GRPO & 63.2 & 65.5 & 60.9 & 32.0 & 24.0 & 26.8 & 48.0 & 52.9 \\
 & OPSD & 63.2 & 65.2 & 60.3 & 32.4 & 23.2 & 26.4 & 47.9 & 52.6 \\
 & \green \method & \green \textbf{67.7} & \green \textbf{70.0} & \green \textbf{65.8} & \green \textbf{32.8} & \green \textbf{24.7} & \green \textbf{27.5} & \green \textbf{54.2} & \green \textbf{57.1} \\
\midrule
\multirow{5}{*}{Qwen3.5-27B} & Vanilla & 68.9 & 68.5 & 63.9 & 32.8 & 26.5 & 28.7 & 53.9 & 56.8 \\
 & SFT & 69.0 & 67.2 & 61.9 & 33.6 & 27.8 & 29.8 & 52.4 & 56.1 \\
 & GRPO & 70.4 & 69.4 & 64.5 & 35.5 & 27.4 & 30.2 & 54.9 & 57.9 \\
 & OPSD & 70.7 & 69.7 & 64.9 & 35.5 & 28.1 & 30.7 & 54.6 & 58.1 \\
 & \green \method & \green \textbf{73.3} & \green \textbf{71.9} & \green \textbf{66.2} & \green \textbf{36.3} & \green \textbf{29.8} & \green \textbf{32.1} & \green \textbf{58.6} & \green \textbf{60.4} \\
\bottomrule
\end{tabular}
}
\end{table}

\begin{table}[t]
\centering
\caption{Accuracy (\%) for Qwen3.5 in non-thinking mode at 64 input frames, with the two Video-MME~2 question groups reported separately. \textit{Logic} and \textit{Rel.} are the group types and \textit{All} combines them by question count. We \textbf{bold} the best result for each backbone and benchmark.}
\label{tab:qwen35-full-64}
\resizebox{\textwidth}{!}{
\setlength{\tabcolsep}{4pt}
\renewcommand{\arraystretch}{1.1}
\begin{tabular}{llcccccccc}
\toprule
\textbf{Backbone Model} & \textbf{Method} & \textbf{Video-MME} & \textbf{MVBench} & \textbf{LongVideoBench} & \multicolumn{3}{c}{\textbf{Video-MME 2}} & \textbf{MLVU} & \textbf{Avg.} \\
\cmidrule(lr){6-8}
 & & & & & Logic & Rel. & All & & \\
\midrule
\multirow{5}{*}{Qwen3.5-0.8B} & Vanilla & 49.4 & 51.8 & 48.9 & 17.4 & 14.9 & 15.8 & 41.9 & 41.6 \\
 & SFT & 49.5 & 52.2 & 49.5 & 17.2 & 14.9 & 15.7 & 42.1 & 41.8 \\
 & GRPO & 50.8 & 53.9 & 50.4 & 19.2 & 18.1 & 18.5 & 43.3 & 43.4 \\
 & OPSD & 50.9 & 53.9 & 51.4 & 18.7 & 16.4 & 17.2 & 44.1 & 43.5 \\
 & \green \method & \green \textbf{54.9} & \green \textbf{58.6} & \green \textbf{55.9} & \green \textbf{22.6} & \green \textbf{18.4} & \green \textbf{19.9} & \green \textbf{45.2} & \green \textbf{46.9} \\
\midrule
\multirow{5}{*}{Qwen3.5-4B} & Vanilla & 64.7 & 67.4 & 58.0 & 27.1 & 23.4 & 24.7 & 50.3 & 53.0 \\
 & SFT & 65.1 & 68.0 & 58.4 & 27.5 & 24.6 & 25.6 & 50.7 & 53.6 \\
 & GRPO & 65.2 & 68.3 & 59.8 & 28.2 & 25.1 & 26.2 & 50.8 & 54.1 \\
 & OPSD & 65.5 & 68.9 & 58.3 & 28.3 & 24.2 & 25.6 & 51.1 & 53.9 \\
 & \green \method & \green \textbf{68.3} & \green \textbf{72.2} & \green \textbf{63.4} & \green \textbf{31.8} & \green \textbf{26.8} & \green \textbf{28.5} & \green \textbf{58.8} & \green \textbf{58.2} \\
\midrule
\multirow{5}{*}{Qwen3.5-9B} & Vanilla & 65.0 & 67.1 & 61.9 & 31.2 & 22.5 & 25.6 & 52.7 & 54.4 \\
 & SFT & 65.9 & 67.2 & 62.5 & 32.4 & 23.3 & 26.5 & 53.7 & 55.2 \\
 & GRPO & 66.3 & 67.7 & 63.4 & 31.9 & 24.1 & 26.8 & 53.5 & 55.5 \\
 & OPSD & 66.1 & 68.7 & 62.9 & 32.7 & 23.5 & 26.7 & 53.3 & 55.5 \\
 & \green \method & \green \textbf{70.7} & \green \textbf{72.8} & \green \textbf{65.2} & \green \textbf{33.2} & \green \textbf{26.6} & \green \textbf{28.9} & \green \textbf{61.6} & \green \textbf{59.8} \\
\midrule
\multirow{5}{*}{Qwen3.5-27B} & Vanilla & 71.4 & 70.2 & 64.2 & 33.6 & 28.8 & 30.4 & 61.5 & 59.5 \\
 & SFT & 71.8 & 70.5 & 64.2 & 33.4 & 29.1 & 30.6 & 61.8 & 59.8 \\
 & GRPO & 72.0 & 71.1 & 64.5 & 35.2 & 30.4 & 32.1 & 61.1 & 60.2 \\
 & OPSD & 71.4 & 70.9 & 65.1 & 34.1 & 28.9 & 30.7 & 62.9 & 60.2 \\
 & \green \method & \green \textbf{76.1} & \green \textbf{76.2} & \green \textbf{68.8} & \green \textbf{37.6} & \green \textbf{32.9} & \green \textbf{34.6} & \green \textbf{66.7} & \green \textbf{64.5} \\
\bottomrule
\end{tabular}
}
\end{table}

\begin{table}[h]
\centering
\caption{Video-MME accuracy (\%) by video duration (Qwen3.5-0.8B). \textbf{All} averages the three splits. We \textbf{bold} the best result for each frame budget and duration.}
\label{tab:qwen35-duration-0.8B}
\footnotesize
\setlength{\tabcolsep}{6pt}
\renewcommand{\arraystretch}{1.1}
\begin{tabular}{llcccc}
\toprule
\textbf{Frames} & \textbf{Method} & \textbf{Short} & \textbf{Medium} & \textbf{Long} & \textbf{All} \\
\midrule
\multirow{5}{*}{16} & Vanilla & 57.1 & 41.5 & 36.7 & 45.1 \\
 & SFT & 57.4 & 41.3 & 37.9 & 45.5 \\
 & GRPO & 59.5 & 42.7 & 38.2 & 46.8 \\
 & OPSD & 57.9 & 42.5 & 38.3 & 46.2 \\
 & \green \method & \green \textbf{61.8} & \green \textbf{47.9} & \green \textbf{41.3} & \green \textbf{50.3} \\
\midrule
\multirow{5}{*}{32} & Vanilla & 58.7 & 46.7 & 39.2 & 48.2 \\
 & SFT & 59.6 & 46.7 & 40.2 & 48.8 \\
 & GRPO & 61.0 & 49.0 & 40.2 & 50.1 \\
 & OPSD & 59.2 & 48.1 & 41.4 & 49.6 \\
 & \green \method & \green \textbf{67.0} & \green \textbf{51.5} & \green \textbf{43.2} & \green \textbf{53.9} \\
\midrule
\multirow{5}{*}{64} & Vanilla & 62.4 & 47.3 & 38.5 & 49.4 \\
 & SFT & 62.1 & 47.1 & 39.4 & 49.5 \\
 & GRPO & 63.9 & 48.8 & 39.6 & 50.8 \\
 & OPSD & 64.4 & 47.9 & 40.5 & 50.9 \\
 & \green \method & \green \textbf{69.1} & \green \textbf{52.7} & \green \textbf{43.0} & \green \textbf{54.9} \\
\bottomrule
\end{tabular}
\end{table}

\begin{table}[h]
\centering
\caption{Video-MME accuracy (\%) by video duration (Qwen3.5-4B). \textbf{All} averages the three splits. We \textbf{bold} the best result for each frame budget and duration.}
\label{tab:qwen35-duration-4B}
\footnotesize
\setlength{\tabcolsep}{6pt}
\renewcommand{\arraystretch}{1.1}
\begin{tabular}{llcccc}
\toprule
\textbf{Frames} & \textbf{Method} & \textbf{Short} & \textbf{Medium} & \textbf{Long} & \textbf{All} \\
\midrule
\multirow{5}{*}{16} & Vanilla & 69.1 & 54.6 & 48.1 & 57.2 \\
 & SFT & 70.3 & 55.9 & 48.6 & 58.3 \\
 & GRPO & 69.8 & 56.2 & 50.5 & 58.8 \\
 & OPSD & 70.3 & 56.1 & 50.6 & 59.0 \\
 & \green \method & \green \textbf{73.7} & \green \textbf{59.9} & \green \textbf{55.0} & \green \textbf{62.8} \\
\midrule
\multirow{5}{*}{32} & Vanilla & 73.4 & 58.8 & 52.1 & 61.4 \\
 & SFT & 74.2 & 60.4 & 53.2 & 62.6 \\
 & GRPO & 74.5 & 61.6 & 53.5 & 63.2 \\
 & OPSD & 74.7 & 60.6 & 53.6 & 62.9 \\
 & \green \method & \green \textbf{79.0} & \green \textbf{64.7} & \green \textbf{57.4} & \green \textbf{67.0} \\
\midrule
\multirow{5}{*}{64} & Vanilla & 76.7 & 62.9 & 54.5 & 64.7 \\
 & SFT & 77.3 & 63.2 & 54.7 & 65.1 \\
 & GRPO & 77.1 & 64.1 & 54.3 & 65.2 \\
 & OPSD & 78.1 & 63.2 & 55.3 & 65.5 \\
 & \green \method & \green \textbf{80.1} & \green \textbf{67.6} & \green \textbf{57.3} & \green \textbf{68.3} \\
\bottomrule
\end{tabular}
\end{table}

\begin{table}[h]
\centering
\caption{Video-MME accuracy (\%) by video duration (Qwen3.5-9B). \textbf{All} averages the three splits. We \textbf{bold} the best result for each frame budget and duration.}
\label{tab:qwen35-duration-9B}
\footnotesize
\setlength{\tabcolsep}{6pt}
\renewcommand{\arraystretch}{1.1}
\begin{tabular}{llcccc}
\toprule
\textbf{Frames} & \textbf{Method} & \textbf{Short} & \textbf{Medium} & \textbf{Long} & \textbf{All} \\
\midrule
\multirow{5}{*}{16} & Vanilla & 69.1 & 54.8 & 50.7 & 58.2 \\
 & SFT & 70.2 & 54.0 & 52.4 & 58.9 \\
 & GRPO & 70.4 & 58.0 & 52.1 & 60.2 \\
 & OPSD & 70.7 & 56.7 & 52.6 & 60.0 \\
 & \green \method & \green \textbf{71.2} & \green \textbf{60.8} & \green \textbf{54.6} & \green \textbf{62.2} \\
\midrule
\multirow{5}{*}{32} & Vanilla & 75.1 & 58.8 & 52.1 & 62.0 \\
 & SFT & 75.3 & 59.5 & 52.9 & 62.6 \\
 & GRPO & 77.1 & 59.1 & 53.4 & 63.2 \\
 & OPSD & 75.7 & 60.3 & 53.7 & 63.2 \\
 & \green \method & \green \textbf{79.6} & \green \textbf{67.3} & \green \textbf{56.1} & \green \textbf{67.7} \\
\midrule
\multirow{5}{*}{64} & Vanilla & 76.6 & 63.8 & 54.5 & 65.0 \\
 & SFT & 77.6 & 64.7 & 55.3 & 65.9 \\
 & GRPO & 78.2 & 65.4 & 55.3 & 66.3 \\
 & OPSD & 77.4 & 65.0 & 55.9 & 66.1 \\
 & \green \method & \green \textbf{82.0} & \green \textbf{70.7} & \green \textbf{59.4} & \green \textbf{70.7} \\
\bottomrule
\end{tabular}
\end{table}

\begin{table}[h]
\centering
\caption{Video-MME accuracy (\%) by video duration (Qwen3.5-27B). \textbf{All} averages the three splits. We \textbf{bold} the best result for each frame budget and duration.}
\label{tab:qwen35-duration-27B}
\footnotesize
\setlength{\tabcolsep}{6pt}
\renewcommand{\arraystretch}{1.1}
\begin{tabular}{llcccc}
\toprule
\textbf{Frames} & \textbf{Method} & \textbf{Short} & \textbf{Medium} & \textbf{Long} & \textbf{All} \\
\midrule
\multirow{5}{*}{16} & Vanilla & 74.5 & 63.5 & 56.5 & 64.8 \\
 & SFT & 74.3 & 62.8 & 57.1 & 64.7 \\
 & GRPO & 75.6 & 64.5 & 57.9 & 66.0 \\
 & OPSD & 75.4 & 64.1 & 58.0 & 65.8 \\
 & \green \method & \green \textbf{77.9} & \green \textbf{67.7} & \green \textbf{59.0} & \green \textbf{68.2} \\
\midrule
\multirow{5}{*}{32} & Vanilla & 79.7 & 67.5 & 59.6 & 68.9 \\
 & SFT & 80.1 & 66.5 & 60.3 & 69.0 \\
 & GRPO & 81.3 & 68.2 & 61.7 & 70.4 \\
 & OPSD & 82.0 & 67.9 & 62.3 & 70.7 \\
 & \green \method & \green \textbf{84.0} & \green \textbf{71.1} & \green \textbf{64.8} & \green \textbf{73.3} \\
\midrule
\multirow{5}{*}{64} & Vanilla & 84.1 & 70.2 & 59.8 & 71.4 \\
 & SFT & 84.0 & 70.8 & 60.6 & 71.8 \\
 & GRPO & 85.1 & 70.4 & 60.5 & 72.0 \\
 & OPSD & 84.2 & 70.0 & 60.0 & 71.4 \\
 & \green \method & \green \textbf{87.8} & \green \textbf{74.8} & \green \textbf{65.8} & \green \textbf{76.1} \\
\bottomrule
\end{tabular}
\end{table}

\subsection{Results on the Qwen3-VL Family}
\label{app:qwen3vl-results}

Section~\ref{sec:main_res} reports the Qwen3.5 family in non-thinking mode. We extend the evaluation to Qwen3-VL~\citep{qwen3vl} at 2B, 4B, and 8B, considering both non-thinking and thinking modes at student frame budgets of 16, 32, and 64. The resulting eighteen model--mode--budget configurations test whether evidence-guided post-training remains effective across a second backbone family, different model sizes, and generation with or without an explicit reasoning trace. The baselines and evaluation protocol follow Appendix~\ref{app:baselines-benchmarks}. At inference, all methods use uniformly sampled student frames; the evidence-privileged teacher view is used only during training.

\paragraph{Table configuration.}
Tables~\ref{tab:qwen3vl-nonthinking-full-16}--\ref{tab:qwen3vl-nonthinking-full-64} report the non-thinking results, and Tables~\ref{tab:qwen3vl-thinking-full-16}--\ref{tab:qwen3vl-thinking-full-64} report the thinking results. Each table fixes the mode and frame budget and compares the five methods within each backbone. The Video-MME duration breakdowns appear in Tables~\ref{tab:qwen3vl-nonthinking-duration-2B}--\ref{tab:qwen3vl-nonthinking-duration-8B} and Tables~\ref{tab:qwen3vl-thinking-duration-2B}--\ref{tab:qwen3vl-thinking-duration-8B}, respectively. The metric definitions and aggregation conventions are the same as in Appendix~\ref{app:main-results-analysis}. All improvements below are percentage-point differences from the strongest of SFT, GRPO, and OPSD for the specified metric, model size, mode, and frame budget.
\begin{table}[t]
\centering
\caption{Accuracy (\%) for Qwen3-VL in non-thinking mode at 16 input frames, with the two Video-MME~2 question groups reported separately. \textit{Logic} and \textit{Rel.} are the group types and \textit{All} combines them by question count. We \textbf{bold} the best result for each backbone and benchmark.}
\label{tab:qwen3vl-nonthinking-full-16}
\resizebox{\textwidth}{!}{
\setlength{\tabcolsep}{4pt}
\renewcommand{\arraystretch}{1.1}
\begin{tabular}{llcccccccc}
\toprule
\textbf{Backbone Model} & \textbf{Method} & \textbf{Video-MME} & \textbf{MVBench} & \textbf{LongVideoBench} & \multicolumn{3}{c}{\textbf{Video-MME 2}} & \textbf{MLVU} & \textbf{Avg.} \\
\cmidrule(lr){6-8}
 & & & & & Logic & Rel. & All & & \\
\midrule
\multirow{5}{*}{Qwen3-VL-2B} & Vanilla & 45.2 & 53.1 & 42.9 & 14.4 & 14.3 & 14.3 & 34.8 & 38.1 \\
 & SFT & 45.8 & 53.1 & 43.4 & 15.4 & 14.9 & 15.1 & 35.8 & 38.6 \\
 & GRPO & 47.0 & 55.4 & 44.2 & 15.8 & 15.9 & 15.9 & 37.2 & 39.9 \\
 & OPSD & 46.5 & 53.2 & 45.3 & 15.9 & 15.5 & 15.6 & 36.7 & 39.5 \\
 & \green \method & \green \textbf{52.4} & \green \textbf{61.2} & \green \textbf{50.7} & \green \textbf{19.9} & \green \textbf{18.1} & \green \textbf{18.7} & \green \textbf{40.6} & \green \textbf{44.7} \\
\midrule
\multirow{5}{*}{Qwen3-VL-4B} & Vanilla & 48.7 & 60.1 & 48.5 & 14.6 & 15.3 & 15.0 & 37.3 & 41.9 \\
 & SFT & 50.1 & 60.9 & 49.9 & 14.2 & 16.2 & 15.5 & 37.1 & 42.7 \\
 & GRPO & 50.4 & 62.3 & 49.5 & 16.7 & 18.2 & 17.7 & 39.1 & 43.8 \\
 & OPSD & 50.2 & 60.8 & 49.9 & 15.3 & 16.7 & 16.2 & 38.9 & 43.2 \\
 & \green \method & \green \textbf{58.9} & \green \textbf{69.8} & \green \textbf{55.3} & \green \textbf{23.8} & \green \textbf{19.0} & \green \textbf{20.7} & \green \textbf{47.0} & \green \textbf{50.3} \\
\midrule
\multirow{5}{*}{Qwen3-VL-8B} & Vanilla & 57.9 & 63.0 & 53.2 & 26.2 & 21.5 & 23.1 & 40.3 & 47.5 \\
 & SFT & 59.2 & 63.5 & 54.6 & 27.3 & 22.1 & 23.9 & 40.0 & 48.3 \\
 & GRPO & 59.2 & 64.6 & 53.8 & 27.3 & 21.6 & 23.6 & 42.1 & 48.7 \\
 & OPSD & 59.1 & 64.2 & 54.2 & 27.1 & 22.6 & 24.2 & 41.6 & 48.7 \\
 & \green \method & \green \textbf{60.7} & \green \textbf{68.2} & \green \textbf{58.5} & \green \textbf{29.8} & \green \textbf{23.6} & \green \textbf{25.8} & \green \textbf{45.6} & \green \textbf{51.8} \\
\bottomrule
\end{tabular}
}
\end{table}

\begin{table}[t]
\centering
\caption{Accuracy (\%) for Qwen3-VL in non-thinking mode at 32 input frames, with the two Video-MME~2 question groups reported separately. \textit{Logic} and \textit{Rel.} are the group types and \textit{All} combines them by question count. We \textbf{bold} the best result for each backbone and benchmark.}
\label{tab:qwen3vl-nonthinking-full-32}
\resizebox{\textwidth}{!}{
\setlength{\tabcolsep}{4pt}
\renewcommand{\arraystretch}{1.1}
\begin{tabular}{llcccccccc}
\toprule
\textbf{Backbone Model} & \textbf{Method} & \textbf{Video-MME} & \textbf{MVBench} & \textbf{LongVideoBench} & \multicolumn{3}{c}{\textbf{Video-MME 2}} & \textbf{MLVU} & \textbf{Avg.} \\
\cmidrule(lr){6-8}
 & & & & & Logic & Rel. & All & & \\
\midrule
\multirow{5}{*}{Qwen3-VL-2B} & Vanilla & 49.4 & 53.8 & 44.3 & 16.1 & 15.3 & 15.6 & 37.1 & 40.0 \\
 & SFT & 50.3 & 54.8 & 46.9 & 17.1 & 16.4 & 16.6 & 38.8 & 41.5 \\
 & GRPO & 51.8 & 55.0 & 45.4 & 18.2 & 16.6 & 17.2 & 40.2 & 41.9 \\
 & OPSD & 51.0 & 55.3 & 46.4 & 18.1 & 16.4 & 17.0 & 38.1 & 41.6 \\
 & \green \method & \green \textbf{56.2} & \green \textbf{60.9} & \green \textbf{54.9} & \green \textbf{22.7} & \green \textbf{18.1} & \green \textbf{19.7} & \green \textbf{45.3} & \green \textbf{47.4} \\
\midrule
\multirow{5}{*}{Qwen3-VL-4B} & Vanilla & 54.3 & 61.4 & 51.1 & 16.4 & 15.1 & 15.6 & 45.1 & 45.5 \\
 & SFT & 55.7 & 62.7 & 54.0 & 17.9 & 17.9 & 17.9 & 45.9 & 47.2 \\
 & GRPO & 55.0 & 64.2 & 51.8 & 18.8 & 15.1 & 16.4 & 47.3 & 46.9 \\
 & OPSD & 55.8 & 62.8 & 52.5 & 18.5 & 17.0 & 17.5 & 46.2 & 47.0 \\
 & \green \method & \green \textbf{62.4} & \green \textbf{71.5} & \green \textbf{58.3} & \green \textbf{25.9} & \green \textbf{21.5} & \green \textbf{23.1} & \green \textbf{50.4} & \green \textbf{53.1} \\
\midrule
\multirow{5}{*}{Qwen3-VL-8B} & Vanilla & 62.1 & 64.2 & 55.1 & 28.6 & 22.0 & 24.3 & 45.6 & 50.3 \\
 & SFT & 63.2 & 64.4 & 57.2 & 28.6 & 23.1 & 25.0 & 46.6 & 51.3 \\
 & GRPO & 62.7 & 65.9 & 54.4 & 29.2 & 23.4 & 25.4 & 46.5 & 51.0 \\
 & OPSD & 63.1 & 64.8 & 55.4 & 29.4 & 23.2 & 25.4 & 47.1 & 51.1 \\
 & \green \method & \green \textbf{65.9} & \green \textbf{70.1} & \green \textbf{60.5} & \green \textbf{31.9} & \green \textbf{23.9} & \green \textbf{26.7} & \green \textbf{51.2} & \green \textbf{54.9} \\
\bottomrule
\end{tabular}
}
\end{table}

\begin{table}[t]
\centering
\caption{Accuracy (\%) for Qwen3-VL in non-thinking mode at 64 input frames, with the two Video-MME~2 question groups reported separately. \textit{Logic} and \textit{Rel.} are the group types and \textit{All} combines them by question count. We \textbf{bold} the best result for each backbone and benchmark.}
\label{tab:qwen3vl-nonthinking-full-64}
\resizebox{\textwidth}{!}{
\setlength{\tabcolsep}{4pt}
\renewcommand{\arraystretch}{1.1}
\begin{tabular}{llcccccccc}
\toprule
\textbf{Backbone Model} & \textbf{Method} & \textbf{Video-MME} & \textbf{MVBench} & \textbf{LongVideoBench} & \multicolumn{3}{c}{\textbf{Video-MME 2}} & \textbf{MLVU} & \textbf{Avg.} \\
\cmidrule(lr){6-8}
 & & & & & Logic & Rel. & All & & \\
\midrule
\multirow{5}{*}{Qwen3-VL-2B} & Vanilla & 49.2 & 53.9 & 45.1 & 16.8 & 17.3 & 17.1 & 40.3 & 41.1 \\
 & SFT & 50.1 & 54.6 & 46.8 & 18.1 & 18.0 & 18.0 & 41.8 & 42.3 \\
 & GRPO & 51.4 & 55.0 & 47.6 & 17.9 & 19.7 & 19.1 & 41.7 & 42.9 \\
 & OPSD & 50.7 & 54.0 & 47.1 & 18.1 & 18.2 & 18.2 & 42.1 & 42.4 \\
 & \green \method & \green \textbf{58.3} & \green \textbf{62.2} & \green \textbf{56.0} & \green \textbf{22.8} & \green \textbf{20.2} & \green \textbf{21.1} & \green \textbf{48.2} & \green \textbf{49.2} \\
\midrule
\multirow{5}{*}{Qwen3-VL-4B} & Vanilla & 56.4 & 63.0 & 51.3 & 17.6 & 16.4 & 16.8 & 50.1 & 47.5 \\
 & SFT & 57.7 & 64.4 & 52.6 & 17.9 & 16.5 & 17.0 & 51.2 & 48.6 \\
 & GRPO & 58.5 & 64.4 & 53.2 & 18.3 & 17.5 & 17.8 & 52.6 & 49.3 \\
 & OPSD & 57.4 & 63.6 & 53.3 & 18.3 & 16.9 & 17.4 & 51.5 & 48.6 \\
 & \green \method & \green \textbf{64.7} & \green \textbf{71.0} & \green \textbf{60.8} & \green \textbf{28.1} & \green \textbf{22.8} & \green \textbf{24.7} & \green \textbf{56.9} & \green \textbf{55.6} \\
\midrule
\multirow{5}{*}{Qwen3-VL-8B} & Vanilla & 66.0 & 64.2 & 56.5 & 30.2 & 24.1 & 26.3 & 51.8 & 53.0 \\
 & SFT & 66.5 & 65.3 & 58.0 & 30.6 & 25.1 & 27.0 & 52.9 & 53.9 \\
 & GRPO & 67.5 & 65.0 & 57.1 & 31.2 & 25.5 & 27.5 & 52.2 & 53.9 \\
 & OPSD & 67.0 & 65.1 & 57.2 & 31.5 & 24.3 & 26.8 & 53.2 & 53.9 \\
 & \green \method & \green \textbf{69.4} & \green \textbf{70.6} & \green \textbf{62.8} & \green \textbf{33.5} & \green \textbf{26.8} & \green \textbf{29.2} & \green \textbf{55.2} & \green \textbf{57.4} \\
\bottomrule
\end{tabular}
}
\end{table}

\begin{table}[h]
\centering
\caption{Video-MME accuracy (\%) by video duration for Qwen3-VL in non-thinking mode (Qwen3-VL-2B). \textbf{All} averages the three splits. We \textbf{bold} the best result for each frame budget and duration.}
\label{tab:qwen3vl-nonthinking-duration-2B}
\footnotesize
\setlength{\tabcolsep}{6pt}
\renewcommand{\arraystretch}{1.1}
\begin{tabular}{llcccc}
\toprule
\textbf{Frames} & \textbf{Method} & \textbf{Short} & \textbf{Medium} & \textbf{Long} & \textbf{All} \\
\midrule
\multirow{5}{*}{16} & Vanilla & 58.4 & 43.1 & 34.2 & 45.2 \\
 & SFT & 59.0 & 43.4 & 34.9 & 45.8 \\
 & GRPO & 59.8 & 44.9 & 36.3 & 47.0 \\
 & OPSD & 59.9 & 44.6 & 35.1 & 46.5 \\
 & \green \method & \green \textbf{64.4} & \green \textbf{51.4} & \green \textbf{41.4} & \green \textbf{52.4} \\
\midrule
\multirow{5}{*}{32} & Vanilla & 65.1 & 48.0 & 35.0 & 49.4 \\
 & SFT & 66.0 & 49.2 & 35.7 & 50.3 \\
 & GRPO & 67.4 & 50.2 & 37.7 & 51.8 \\
 & OPSD & 66.4 & 50.0 & 36.7 & 51.0 \\
 & \green \method & \green \textbf{69.0} & \green \textbf{56.0} & \green \textbf{43.7} & \green \textbf{56.2} \\
\midrule
\multirow{5}{*}{64} & Vanilla & 66.7 & 47.6 & 33.4 & 49.2 \\
 & SFT & 67.3 & 48.5 & 34.5 & 50.1 \\
 & GRPO & 69.2 & 50.5 & 34.4 & 51.4 \\
 & OPSD & 67.9 & 49.1 & 35.1 & 50.7 \\
 & \green \method & \green \textbf{72.7} & \green \textbf{57.2} & \green \textbf{45.0} & \green \textbf{58.3} \\
\bottomrule
\end{tabular}
\end{table}

\begin{table}[h]
\centering
\caption{Video-MME accuracy (\%) by video duration for Qwen3-VL in non-thinking mode (Qwen3-VL-4B). \textbf{All} averages the three splits. We \textbf{bold} the best result for each frame budget and duration.}
\label{tab:qwen3vl-nonthinking-duration-4B}
\footnotesize
\setlength{\tabcolsep}{6pt}
\renewcommand{\arraystretch}{1.1}
\begin{tabular}{llcccc}
\toprule
\textbf{Frames} & \textbf{Method} & \textbf{Short} & \textbf{Medium} & \textbf{Long} & \textbf{All} \\
\midrule
\multirow{5}{*}{16} & Vanilla & 62.2 & 45.9 & 37.9 & 48.7 \\
 & SFT & 64.2 & 47.7 & 38.4 & 50.1 \\
 & GRPO & 64.7 & 46.8 & 39.8 & 50.4 \\
 & OPSD & 64.4 & 47.0 & 39.2 & 50.2 \\
 & \green \method & \green \textbf{69.1} & \green \textbf{58.4} & \green \textbf{49.2} & \green \textbf{58.9} \\
\midrule
\multirow{5}{*}{32} & Vanilla & 68.9 & 50.4 & 43.8 & 54.3 \\
 & SFT & 71.1 & 50.9 & 45.1 & 55.7 \\
 & GRPO & 69.3 & 51.5 & 44.3 & 55.0 \\
 & OPSD & 70.5 & 52.2 & 44.8 & 55.8 \\
 & \green \method & \green \textbf{74.7} & \green \textbf{59.9} & \green \textbf{52.8} & \green \textbf{62.4} \\
\midrule
\multirow{5}{*}{64} & Vanilla & 70.1 & 53.4 & 45.9 & 56.4 \\
 & SFT & 71.2 & 54.9 & 46.9 & 57.7 \\
 & GRPO & 72.0 & 55.7 & 47.8 & 58.5 \\
 & OPSD & 71.6 & 54.3 & 46.3 & 57.4 \\
 & \green \method & \green \textbf{77.1} & \green \textbf{63.6} & \green \textbf{53.4} & \green \textbf{64.7} \\
\bottomrule
\end{tabular}
\end{table}

\begin{table}[h]
\centering
\caption{Video-MME accuracy (\%) by video duration for Qwen3-VL in non-thinking mode (Qwen3-VL-8B). \textbf{All} averages the three splits. We \textbf{bold} the best result for each frame budget and duration.}
\label{tab:qwen3vl-nonthinking-duration-8B}
\footnotesize
\setlength{\tabcolsep}{6pt}
\renewcommand{\arraystretch}{1.1}
\begin{tabular}{llcccc}
\toprule
\textbf{Frames} & \textbf{Method} & \textbf{Short} & \textbf{Medium} & \textbf{Long} & \textbf{All} \\
\midrule
\multirow{5}{*}{16} & Vanilla & 68.4 & 55.1 & 50.0 & 57.9 \\
 & SFT & 69.6 & 57.2 & 50.9 & 59.2 \\
 & GRPO & 70.9 & 54.9 & 51.9 & 59.2 \\
 & OPSD & 69.5 & 56.4 & 51.4 & 59.1 \\
 & \green \method & \green \textbf{72.2} & \green \textbf{57.8} & \green \textbf{52.2} & \green \textbf{60.7} \\
\midrule
\multirow{5}{*}{32} & Vanilla & 74.0 & 58.0 & 54.1 & 62.1 \\
 & SFT & 74.6 & 60.0 & 55.0 & 63.2 \\
 & GRPO & 75.4 & 58.2 & 54.5 & 62.7 \\
 & OPSD & 75.3 & 58.4 & 55.5 & 63.1 \\
 & \green \method & \green \textbf{77.3} & \green \textbf{63.2} & \green \textbf{57.1} & \green \textbf{65.9} \\
\midrule
\multirow{5}{*}{64} & Vanilla & 78.7 & 64.0 & 55.3 & 66.0 \\
 & SFT & 79.2 & 64.5 & 55.8 & 66.5 \\
 & GRPO & 80.3 & 65.5 & 56.6 & 67.5 \\
 & OPSD & 80.0 & 65.0 & 55.9 & 67.0 \\
 & \green \method & \green \textbf{82.3} & \green \textbf{68.0} & \green \textbf{58.0} & \green \textbf{69.4} \\
\bottomrule
\end{tabular}
\end{table}

\begin{table}[t]
\centering
\caption{Accuracy (\%) for Qwen3-VL in thinking mode at 16 input frames, with the two Video-MME~2 question groups reported separately. \textit{Logic} and \textit{Rel.} are the group types and \textit{All} combines them by question count. We \textbf{bold} the best result for each backbone and benchmark.}
\label{tab:qwen3vl-thinking-full-16}
\resizebox{\textwidth}{!}{
\setlength{\tabcolsep}{4pt}
\renewcommand{\arraystretch}{1.1}
\begin{tabular}{llcccccccc}
\toprule
\textbf{Backbone Model} & \textbf{Method} & \textbf{Video-MME} & \textbf{MVBench} & \textbf{LongVideoBench} & \multicolumn{3}{c}{\textbf{Video-MME 2}} & \textbf{MLVU} & \textbf{Avg.} \\
\cmidrule(lr){6-8}
 & & & & & Logic & Rel. & All & & \\
\midrule
\multirow{5}{*}{Qwen3-VL-2B} & Base model & 49.3 & 60.5 & 45.1 & 17.8 & 15.3 & 16.2 & 35.1 & 41.3 \\
 & SFT & 51.0 & 61.5 & 46.0 & 19.4 & 16.8 & 17.7 & 37.1 & 42.7 \\
 & GRPO & 50.9 & 61.8 & 46.0 & 18.3 & 15.2 & 16.3 & 36.3 & 42.3 \\
 & OPSD & 50.0 & 61.3 & 46.6 & 19.9 & 16.6 & 17.8 & 35.2 & 42.2 \\
 & \green \method & \green \textbf{54.2} & \green \textbf{65.9} & \green \textbf{52.8} & \green \textbf{22.1} & \green \textbf{17.4} & \green \textbf{19.1} & \green \textbf{43.0} & \green \textbf{47.0} \\
\midrule
\multirow{5}{*}{Qwen3-VL-4B} & Base model & 54.5 & 64.2 & 55.4 & 23.1 & 18.6 & 20.2 & 41.3 & 47.1 \\
 & SFT & 56.0 & 64.8 & 55.5 & 23.8 & 20.4 & 21.6 & 42.8 & 48.1 \\
 & GRPO & 55.5 & 65.5 & 56.3 & 22.5 & 19.5 & 20.6 & 42.0 & 48.0 \\
 & OPSD & 55.5 & 65.9 & 56.4 & 24.5 & 19.1 & 21.0 & 42.6 & 48.3 \\
 & \green \method & \green \textbf{59.1} & \green \textbf{69.7} & \green \textbf{58.2} & \green \textbf{25.1} & \green \textbf{21.0} & \green \textbf{22.4} & \green \textbf{47.4} & \green \textbf{51.4} \\
\midrule
\multirow{5}{*}{Qwen3-VL-8B} & Base model & 55.3 & 65.3 & 54.0 & 23.1 & 19.1 & 20.5 & 40.0 & 47.0 \\
 & SFT & 55.8 & 66.1 & 55.1 & 23.5 & 19.5 & 20.9 & 41.7 & 47.9 \\
 & GRPO & 57.0 & 66.6 & 54.4 & 24.0 & 20.4 & 21.7 & 41.8 & 48.3 \\
 & OPSD & 56.4 & 66.6 & 55.5 & 24.0 & 19.5 & 21.1 & 41.3 & 48.2 \\
 & \green \method & \green \textbf{60.5} & \green \textbf{70.5} & \green \textbf{57.0} & \green \textbf{27.2} & \green \textbf{22.1} & \green \textbf{23.8} & \green \textbf{47.0} & \green \textbf{51.8} \\
\bottomrule
\end{tabular}
}
\end{table}

\begin{table}[t]
\centering
\caption{Accuracy (\%) for Qwen3-VL in thinking mode at 32 input frames, with the two Video-MME~2 question groups reported separately. \textit{Logic} and \textit{Rel.} are the group types and \textit{All} combines them by question count. We \textbf{bold} the best result for each backbone and benchmark.}
\label{tab:qwen3vl-thinking-full-32}
\resizebox{\textwidth}{!}{
\setlength{\tabcolsep}{4pt}
\renewcommand{\arraystretch}{1.1}
\begin{tabular}{llcccccccc}
\toprule
\textbf{Backbone Model} & \textbf{Method} & \textbf{Video-MME} & \textbf{MVBench} & \textbf{LongVideoBench} & \multicolumn{3}{c}{\textbf{Video-MME 2}} & \textbf{MLVU} & \textbf{Avg.} \\
\cmidrule(lr){6-8}
 & & & & & Logic & Rel. & All & & \\
\midrule
\multirow{5}{*}{Qwen3-VL-2B} & Base model & 52.2 & 61.0 & 51.0 & 20.1 & 16.1 & 17.5 & 38.0 & 44.0 \\
 & SFT & 53.3 & 62.1 & 52.3 & 20.3 & 16.7 & 18.0 & 38.1 & 44.8 \\
 & GRPO & 53.5 & 61.2 & 51.9 & 21.9 & 17.4 & 19.0 & 39.1 & 44.9 \\
 & OPSD & 53.1 & 62.1 & 52.9 & 21.3 & 17.4 & 18.8 & 39.1 & 45.2 \\
 & \green \method & \green \textbf{57.4} & \green \textbf{67.2} & \green \textbf{55.0} & \green \textbf{25.5} & \green \textbf{19.4} & \green \textbf{21.5} & \green \textbf{46.0} & \green \textbf{49.4} \\
\midrule
\multirow{5}{*}{Qwen3-VL-4B} & Base model & 58.9 & 67.0 & 57.3 & 25.2 & 20.3 & 22.0 & 49.4 & 50.9 \\
 & SFT & 59.4 & 68.0 & 57.9 & 25.8 & 20.4 & 22.3 & 49.4 & 51.4 \\
 & GRPO & 58.5 & 66.7 & 57.2 & 25.5 & 20.2 & 22.1 & 49.1 & 50.7 \\
 & OPSD & 59.5 & 67.3 & 57.5 & 26.2 & 20.8 & 22.7 & 49.0 & 51.2 \\
 & \green \method & \green \textbf{63.0} & \green \textbf{71.5} & \green \textbf{60.2} & \green \textbf{28.9} & \green \textbf{23.4} & \green \textbf{25.3} & \green \textbf{52.4} & \green \textbf{54.5} \\
\midrule
\multirow{5}{*}{Qwen3-VL-8B} & Base model & 61.2 & 66.1 & 58.2 & 26.1 & 20.1 & 22.2 & 45.0 & 50.5 \\
 & SFT & 62.1 & 66.9 & 59.1 & 26.2 & 20.7 & 22.6 & 45.5 & 51.3 \\
 & GRPO & 61.8 & 66.0 & 58.8 & 27.3 & 21.3 & 23.4 & 44.4 & 50.9 \\
 & OPSD & 61.6 & 66.9 & 58.5 & 26.4 & 20.9 & 22.8 & 45.3 & 51.0 \\
 & \green \method & \green \textbf{64.9} & \green \textbf{71.8} & \green \textbf{60.6} & \green \textbf{29.1} & \green \textbf{24.1} & \green \textbf{25.8} & \green \textbf{53.0} & \green \textbf{55.2} \\
\bottomrule
\end{tabular}
}
\end{table}

\begin{table}[t]
\centering
\caption{Accuracy (\%) for Qwen3-VL in thinking mode at 64 input frames, with the two Video-MME~2 question groups reported separately. \textit{Logic} and \textit{Rel.} are the group types and \textit{All} combines them by question count. We \textbf{bold} the best result for each backbone and benchmark.}
\label{tab:qwen3vl-thinking-full-64}
\resizebox{\textwidth}{!}{
\setlength{\tabcolsep}{4pt}
\renewcommand{\arraystretch}{1.1}
\begin{tabular}{llcccccccc}
\toprule
\textbf{Backbone Model} & \textbf{Method} & \textbf{Video-MME} & \textbf{MVBench} & \textbf{LongVideoBench} & \multicolumn{3}{c}{\textbf{Video-MME 2}} & \textbf{MLVU} & \textbf{Avg.} \\
\cmidrule(lr){6-8}
 & & & & & Logic & Rel. & All & & \\
\midrule
\multirow{5}{*}{Qwen3-VL-2B} & Base model & 55.5 & 63.1 & 51.4 & 20.1 & 18.1 & 18.8 & 46.0 & 47.0 \\
 & SFT & 56.4 & 65.2 & 52.1 & 20.6 & 19.5 & 19.9 & 46.2 & 48.0 \\
 & GRPO & 56.7 & 64.0 & 52.4 & 21.2 & 19.2 & 19.9 & 47.3 & 48.1 \\
 & OPSD & 56.5 & 64.7 & 52.0 & 20.9 & 18.8 & 19.5 & 47.1 & 48.0 \\
 & \green \method & \green \textbf{60.3} & \green \textbf{66.8} & \green \textbf{55.6} & \green \textbf{25.6} & \green \textbf{21.3} & \green \textbf{22.8} & \green \textbf{49.6} & \green \textbf{51.0} \\
\midrule
\multirow{5}{*}{Qwen3-VL-4B} & Base model & 61.8 & 69.1 & 58.1 & 26.1 & 21.1 & 22.8 & 52.1 & 52.8 \\
 & SFT & 63.2 & 70.3 & 58.8 & 27.5 & 21.3 & 23.5 & 52.7 & 53.7 \\
 & GRPO & 62.5 & 70.7 & 59.7 & 26.8 & 21.2 & 23.2 & 53.8 & 54.0 \\
 & OPSD & 63.0 & 69.6 & 58.9 & 26.6 & 21.6 & 23.4 & 52.1 & 53.4 \\
 & \green \method & \green \textbf{66.2} & \green \textbf{73.8} & \green \textbf{61.1} & \green \textbf{30.0} & \green \textbf{23.7} & \green \textbf{25.9} & \green \textbf{56.0} & \green \textbf{56.6} \\
\midrule
\multirow{5}{*}{Qwen3-VL-8B} & Base model & 63.0 & 68.7 & 59.9 & 26.9 & 22.3 & 23.9 & 48.8 & 52.9 \\
 & SFT & 64.3 & 70.0 & 60.1 & 28.0 & 24.0 & 25.4 & 49.9 & 53.9 \\
 & GRPO & 64.2 & 70.2 & 60.5 & 28.3 & 23.8 & 25.4 & 49.6 & 54.0 \\
 & OPSD & 64.2 & 69.4 & 61.4 & 27.5 & 22.3 & 24.1 & 49.5 & 53.7 \\
 & \green \method & \green \textbf{66.4} & \green \textbf{73.5} & \green \textbf{63.0} & \green \textbf{29.5} & \green \textbf{26.6} & \green \textbf{27.6} & \green \textbf{53.0} & \green \textbf{56.7} \\
\bottomrule
\end{tabular}
}
\end{table}

\begin{table}[h]
\centering
\caption{Video-MME accuracy (\%) by video duration for Qwen3-VL in thinking mode (Qwen3-VL-2B). \textbf{All} averages the three splits. We \textbf{bold} the best result for each frame budget and duration.}
\label{tab:qwen3vl-thinking-duration-2B}
\footnotesize
\setlength{\tabcolsep}{6pt}
\renewcommand{\arraystretch}{1.1}
\begin{tabular}{llcccc}
\toprule
\textbf{Frames} & \textbf{Method} & \textbf{Short} & \textbf{Medium} & \textbf{Long} & \textbf{All} \\
\midrule
\multirow{5}{*}{16} & Base model & 61.5 & 46.3 & 40.1 & 49.3 \\
 & SFT & 63.5 & 48.5 & 41.1 & 51.0 \\
 & GRPO & 63.0 & 47.5 & 42.2 & 50.9 \\
 & OPSD & 62.6 & 46.9 & 40.5 & 50.0 \\
 & \green \method & \green \textbf{66.0} & \green \textbf{52.8} & \green \textbf{43.8} & \green \textbf{54.2} \\
\midrule
\multirow{5}{*}{32} & Base model & 65.2 & 50.4 & 41.0 & 52.2 \\
 & SFT & 66.2 & 52.0 & 41.8 & 53.3 \\
 & GRPO & 66.1 & 51.6 & 42.7 & 53.5 \\
 & OPSD & 66.2 & 51.9 & 41.2 & 53.1 \\
 & \green \method & \green \textbf{70.3} & \green \textbf{54.4} & \green \textbf{47.3} & \green \textbf{57.4} \\
\midrule
\multirow{5}{*}{64} & Base model & 71.5 & 52.3 & 42.5 & 55.5 \\
 & SFT & 72.6 & 52.5 & 44.2 & 56.4 \\
 & GRPO & 73.4 & 53.8 & 42.9 & 56.7 \\
 & OPSD & 72.0 & 52.6 & 44.8 & 56.5 \\
 & \green \method & \green \textbf{74.6} & \green \textbf{59.3} & \green \textbf{47.0} & \green \textbf{60.3} \\
\bottomrule
\end{tabular}
\end{table}

\begin{table}[h]
\centering
\caption{Video-MME accuracy (\%) by video duration for Qwen3-VL in thinking mode (Qwen3-VL-4B). \textbf{All} averages the three splits. We \textbf{bold} the best result for each frame budget and duration.}
\label{tab:qwen3vl-thinking-duration-4B}
\footnotesize
\setlength{\tabcolsep}{6pt}
\renewcommand{\arraystretch}{1.1}
\begin{tabular}{llcccc}
\toprule
\textbf{Frames} & \textbf{Method} & \textbf{Short} & \textbf{Medium} & \textbf{Long} & \textbf{All} \\
\midrule
\multirow{5}{*}{16} & Base model & 65.4 & 52.1 & 46.0 & 54.5 \\
 & SFT & 67.1 & 53.4 & 47.5 & 56.0 \\
 & GRPO & 66.9 & 52.9 & 46.7 & 55.5 \\
 & OPSD & 66.2 & 53.0 & 47.4 & 55.5 \\
 & \green \method & \green \textbf{70.3} & \green \textbf{56.5} & \green \textbf{50.6} & \green \textbf{59.1} \\
\midrule
\multirow{5}{*}{32} & Base model & 72.4 & 56.1 & 48.2 & 58.9 \\
 & SFT & 73.2 & 56.6 & 48.3 & 59.4 \\
 & GRPO & 72.1 & 54.9 & 48.4 & 58.5 \\
 & OPSD & 73.3 & 56.7 & 48.5 & 59.5 \\
 & \green \method & \green \textbf{75.1} & \green \textbf{60.6} & \green \textbf{53.3} & \green \textbf{63.0} \\
\midrule
\multirow{5}{*}{64} & Base model & 74.2 & 61.1 & 50.3 & 61.8 \\
 & SFT & 75.8 & 62.2 & 51.6 & 63.2 \\
 & GRPO & 74.5 & 61.4 & 51.5 & 62.5 \\
 & OPSD & 75.4 & 62.3 & 51.4 & 63.0 \\
 & \green \method & \green \textbf{78.3} & \green \textbf{65.8} & \green \textbf{54.6} & \green \textbf{66.2} \\
\bottomrule
\end{tabular}
\end{table}

\begin{table}[h]
\centering
\caption{Video-MME accuracy (\%) by video duration for Qwen3-VL in thinking mode (Qwen3-VL-8B). \textbf{All} averages the three splits. We \textbf{bold} the best result for each frame budget and duration.}
\label{tab:qwen3vl-thinking-duration-8B}
\footnotesize
\setlength{\tabcolsep}{6pt}
\renewcommand{\arraystretch}{1.1}
\begin{tabular}{llcccc}
\toprule
\textbf{Frames} & \textbf{Method} & \textbf{Short} & \textbf{Medium} & \textbf{Long} & \textbf{All} \\
\midrule
\multirow{5}{*}{16} & Base model & 66.1 & 52.6 & 47.2 & 55.3 \\
 & SFT & 67.0 & 52.8 & 47.7 & 55.8 \\
 & GRPO & 68.4 & 54.4 & 48.2 & 57.0 \\
 & OPSD & 67.8 & 53.5 & 47.8 & 56.4 \\
 & \green \method & \green \textbf{71.4} & \green \textbf{58.8} & \green \textbf{51.2} & \green \textbf{60.5} \\
\midrule
\multirow{5}{*}{32} & Base model & 73.2 & 57.2 & 53.2 & 61.2 \\
 & SFT & 73.8 & 58.4 & 54.2 & 62.1 \\
 & GRPO & 73.7 & 57.9 & 53.7 & 61.8 \\
 & OPSD & 73.7 & 57.7 & 53.5 & 61.6 \\
 & \green \method & \green \textbf{76.6} & \green \textbf{62.5} & \green \textbf{55.7} & \green \textbf{64.9} \\
\midrule
\multirow{5}{*}{64} & Base model & 75.5 & 62.5 & 51.1 & 63.0 \\
 & SFT & 76.9 & 63.2 & 52.7 & 64.3 \\
 & GRPO & 76.8 & 63.5 & 52.2 & 64.2 \\
 & OPSD & 76.9 & 63.6 & 52.2 & 64.2 \\
 & \green \method & \green \textbf{78.1} & \green \textbf{66.2} & \green \textbf{54.8} & \green \textbf{66.4} \\
\bottomrule
\end{tabular}
\end{table}

\paragraph{Results in non-thinking mode.}
\method achieves the highest accuracy on all five benchmarks and the highest macro-average in each of the nine non-thinking configurations. Its macro-average advantage ranges from 3.1 to 6.5 points. For the 2B model, the macro-average rises from 44.7 at 16 frames to 47.4 at 32 frames and 49.2 at 64 frames, exceeding the corresponding strongest baselines by 4.8, 5.5, and 6.3 points. The 4B model also shows large gains: at 16 frames, it reaches 50.3 compared with 43.8 for GRPO, giving the largest macro-average margin of 6.5 points. The advantage remains positive at 8B, where the margins are 3.1, 3.6, and 3.5 points across the three budgets. These results show that the improvement extends across all evaluated Qwen3-VL sizes, with larger margins for the 2B and 4B models in this mode.

\paragraph{Results in thinking mode.}
The same ordering holds in all nine thinking configurations: \method leads the baselines on every benchmark and in macro-average accuracy. Using the reported macro-averages, its advantage ranges from 2.6 to 4.3 points. At 64 frames, the 2B, 4B, and 8B models reach 51.0, 56.6, and 56.7, improving on the strongest baselines by 2.9, 2.6, and 2.7 points, respectively. The strongest baseline varies across configurations: SFT leads at 2B with 16 frames, OPSD at 2B with 32 frames, and GRPO at 2B with 64 frames. The advantage of \method therefore persists across different baseline orderings and when the student produces an explicit reasoning trace. This supports the usefulness of visual-evidence supervision for thinking models as well as for models evaluated in non-thinking mode.

\paragraph{Temporal understanding and long-video reasoning.}
The per-benchmark results show that the macro-average improvement is supported by complementary video-understanding tasks. In non-thinking mode at 2B with 64 frames, \method scores 62.2 on MVBench, 56.0 on LongVideoBench, and 48.2 on MLVU, exceeding the strongest baseline on each benchmark by 7.2, 8.4, and 6.1 points. In thinking mode at 8B with 32 frames, it reaches 71.8 on MVBench and 53.0 on MLVU, compared with strongest-baseline scores of 66.9 and 45.5. These gains span temporal perception, reasoning about referred content within a long context, and tasks that combine broad video understanding with access to specific details. Their occurrence in both modes is consistent with the teacher view's use of global context together with question-conditioned local evidence.

\paragraph{Logic and relevance in Video-MME~2.}
Both question subsets improve over the strongest baseline in every model--mode--budget configuration. In non-thinking mode at 4B with 64 frames, \method reaches 28.1 on logic and 22.8 on relevance, compared with strongest-baseline scores of 18.3 and 17.5. The respective gains of 9.8 and 5.3 points show that the combined improvement is supported by both subsets, although their margins differ. In thinking mode at 8B with 64 frames, logic accuracy rises to 29.5 from 28.3 and relevance accuracy to 26.6 from 24.0, giving gains of 1.2 and 2.6 points. The subset breakdown thus reveals variation in which questions benefit most while retaining a positive improvement on both types under the common per-question accuracy protocol.

\paragraph{Performance across video durations.}
The Video-MME duration tables show that \method outperforms the baselines on short, medium, and long videos in all eighteen configurations. In non-thinking mode at 2B with 64 frames, its short, medium, and long accuracies are 72.7, 57.2, and 45.0, exceeding the strongest baselines in the corresponding splits by 3.5, 6.7, and 9.9 points. In thinking mode at 4B with 32 frames, the three scores are 75.1, 60.6, and 53.3, with margins of 1.8, 3.9, and 4.8 points. The larger gains on long videos in these configurations are consistent with retaining global temporal context around the selected evidence. The improvements on shorter videos additionally show that evidence-guided training remains useful when the relevant information lies within a smaller temporal span. 

\paragraph{Robustness across frame budgets.}
For each of the six model--mode combinations, the macro-average of \method increases as the student budget grows from 16 to 32 and then to 64 frames. For example, the 4B thinking model reaches 51.4, 54.5, and 56.6, and retains an advantage over the strongest baseline at every budget. Individual benchmark scores can respond differently to additional frames: the 2B thinking model's MVBench accuracy changes from 67.2 at 32 frames to 66.8 at 64 frames, while its macro-average increases from 49.4 to 51.0. The consistent result across budgets is therefore the improvement over baselines evaluated at the same frame budget. This pattern indicates that evidence-guided post-training remains useful both when the student view is sparse and when more frames are available.

\paragraph{Comparison between reasoning modes.}
The comparison between thinking and non-thinking modes varies with the backbone and frame budget. At 2B with 64 frames, \method reaches a macro-average of 51.0 in thinking mode and 49.2 in non-thinking mode. At 8B with the same frame budget, the ordering reverses: the non-thinking model scores 57.4 and the thinking model 56.7. In both cases, \method improves on the strongest baseline within each mode. Together with the Qwen3.5 results, these comparisons cover both evaluated backbone families and reasoning modes, with different margins across configurations.

\subsection{Additional Motivation Evidence Across Models and Frame Budgets}
\label{app:motivation-generalization}

The main text presents the motivation diagnostics for Qwen3.5-9B with a 64-frame student view. This subsection reports the same two supervision gaps on four additional configurations: Qwen3.5-0.8B at 16 and 64 frames, Qwen3.5-27B at 64 frames, and Qwen3-VL-8B in thinking mode at 64 frames. The left panel measures concentration of the teacher--student absolute log-probability gap across response tokens, and the right panel follows output KL and hidden divergence during a 300-step output-only distillation diagnostic.

\begin{figure*}[t]
    \centering
    \begin{subfigure}[t]{0.48\textwidth}
        \centering
        \includegraphics[width=\linewidth]{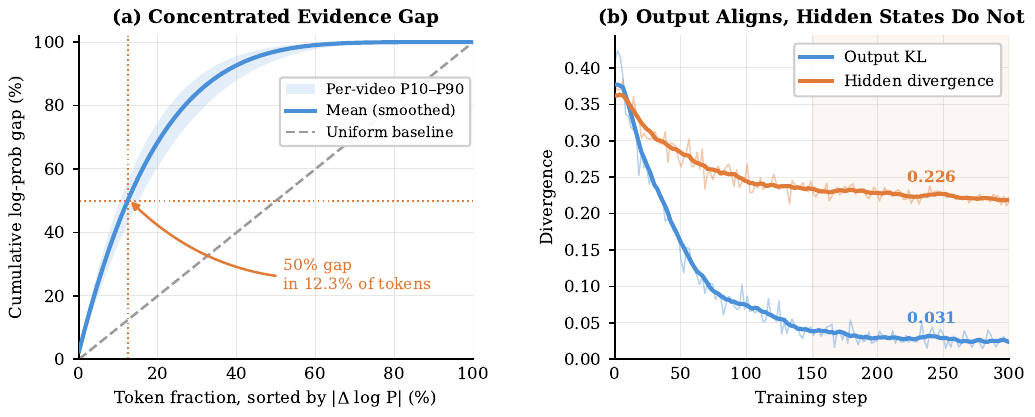}
        \caption{Qwen3.5-0.8B, non-thinking, $16$ frames.}
    \end{subfigure}\hfill
    \begin{subfigure}[t]{0.48\textwidth}
        \centering
        \includegraphics[width=\linewidth]{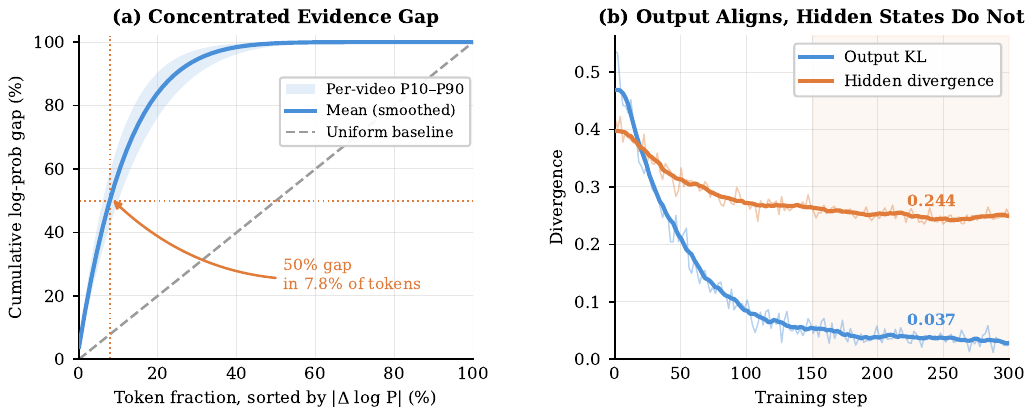}
        \caption{Qwen3.5-0.8B, non-thinking, $64$ frames.}
    \end{subfigure}

    \medskip
    \begin{subfigure}[t]{0.48\textwidth}
        \centering
        \includegraphics[width=\linewidth]{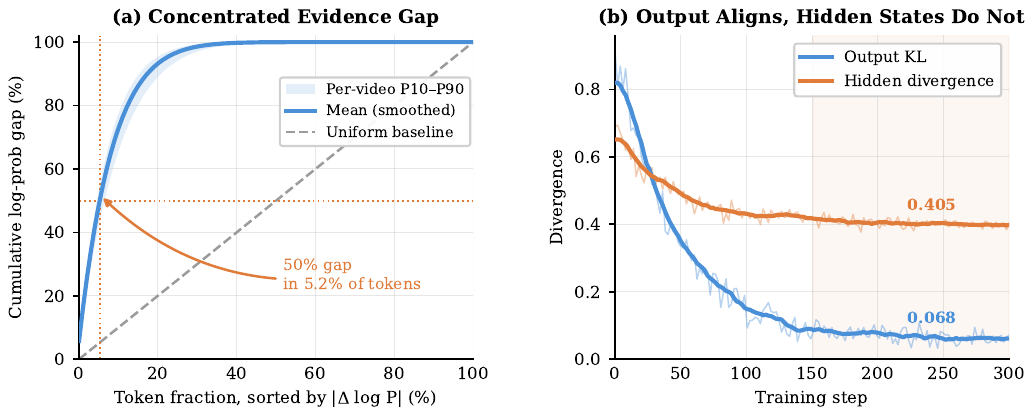}
        \caption{Qwen3.5-27B, non-thinking, $64$ frames.}
    \end{subfigure}\hfill
    \begin{subfigure}[t]{0.48\textwidth}
        \centering
        \includegraphics[width=\linewidth]{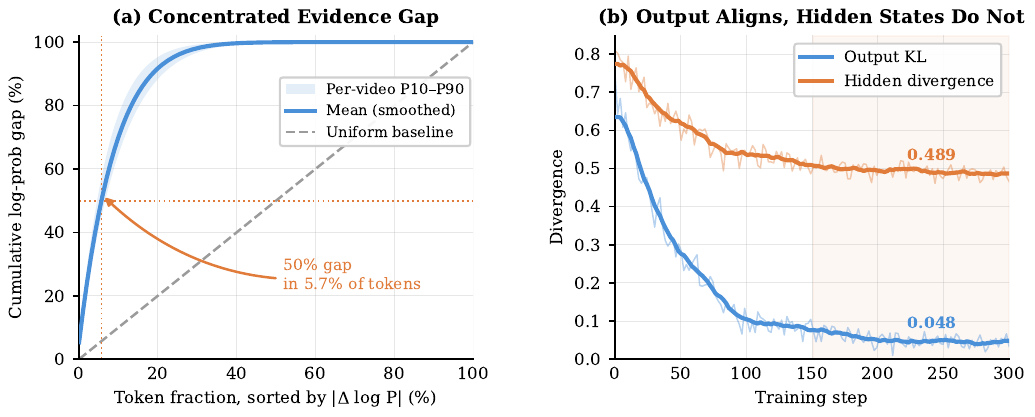}
        \caption{Qwen3-VL-8B, thinking, $64$ frames.}
    \end{subfigure}
    \caption{\textbf{Additional motivation diagnostics across four configurations.} Each subfigure contains the same two diagnostics: the left panel reports the cumulative absolute teacher--student log-probability gap, with the marked point giving the response-token fraction that carries $50\%$ of the gap; the right panel reports output KL and hidden divergence during 300 steps of output-only distillation. }
    \label{fig:motivation-generalization}
\end{figure*}

\paragraph{Concentrated evidence sensitivity across budgets and scales.}
For Qwen3.5-0.8B, increasing the student budget from $16$ to $64$ frames moves the $50\%$-gap point from $12.3\%$ to $7.8\%$ of response tokens. At $64$ frames, Qwen3.5-27B reaches the same point at $5.2\%$ of tokens. In each setting, a small fraction of response tokens carries half of the gap.

\paragraph{The pattern transfers to another backbone and reasoning mode.}
The Qwen3-VL-8B thinking configuration places $50\%$ of its gap on $5.7\%$ of response tokens at $64$ frames. Its curve has the same qualitative shape as the Qwen3.5 curves, with a small fraction of tokens accounting for a large fraction of the view-dependent likelihood shift.

\paragraph{Output alignment does not remove the hidden-state gap.}
The right panels show the complementary supervision gap. At the final plotted step, output KL lies between $0.031$ and $0.068$ across the four configurations, whereas hidden divergence remains between $0.226$ and $0.489$. For example, Qwen3.5-0.8B at $16$ frames reaches $0.031$ output KL but retains $0.226$ hidden divergence; Qwen3.5-27B at $64$ frames reaches $0.068$ and $0.405$, respectively; and Qwen3-VL-8B thinking reaches $0.048$ and $0.489$. The exact scales vary with the model and budget, but output-only matching consistently leaves a non-zero representation discrepancy in these diagnostics. This is the empirical gap that motivates adding SSA to the token-level evidence signal used by ECR.

\subsection{Ablation Across Models and Frame Budgets}
\label{app:ablation-results}

The ablations in this subsection extend the Qwen3.5-9B comparison in Section~\ref{sec:ab} to Qwen3.5-0.8B, 4B, and 27B in non-thinking mode, and to Qwen3-VL-2B, 4B, and 8B in both non-thinking and thinking modes. Each model--mode combination is considered at student frame budgets of $N\in\{16,32,64\}$, giving 27 configurations in the appendix. The comparisons address three parts of the design: the temporal context supplied by the teacher view, the allocation of positive credit across response tokens, and the alignment of the representations formed under the two views.

\paragraph{Comparison setup.}
Within each model--mode--budget configuration, the full method and its ablations share the training data, optimizer settings, student rollout budget, and evaluation protocol. Each variant changes one component while retaining the others. The student generates responses from the uniformly sampled view, and every evaluation uses $N$ uniformly sampled frames.

\paragraph{Ablation variants.}
The four traces in each panel correspond to the following configurations:
\begin{itemize}[itemsep=2pt, leftmargin=*]
\item \textbf{\method} uses the global-plus-local teacher view together with ECR and SSA. ECR redistributes positive response-level advantages according to the evidence contrast of Equation~\ref{eq:ecr-profile}, and SSA aligns response-end hidden states on verifier-accepted responses through Equation~\ref{eq:ssa}.
\item \textbf{w/ Key Windows} replaces the global-plus-local teacher view with question-conditioned local windows, while retaining ECR and SSA.
\item \textbf{w/o ECR} removes the evidence-dependent token weights while retaining the global-plus-local teacher view and SSA. In Equation~\ref{eq:ecr-profile}, the token advantages reduce to $\widetilde A_t^i=A^i$.
\item \textbf{w/o SSA} sets $\lambda_{\mathrm{SSA}}=0$ in Equation~\ref{eq:owlet-objective}, while preserving the global-plus-local teacher view and ECR.
\end{itemize}

Removing both ECR and SSA would recover the GRPO objective of Equation~\ref{eq:rl}; the reported ablations remove one component at a time.

\paragraph{Comparisons across configurations.}
Within each figure, the three panels hold the model and reasoning mode fixed while varying the student frame budget; the remaining figures vary model size and reasoning mode.

Across the grouped plots, removing ECR or SSA lowers accuracy by different amounts across tasks, and replacing the global-plus-local teacher view with key windows is the strongest reduction in many panels. Increasing the student frame budget changes absolute accuracy and individual margins, while the full method remains the leading configuration across the evaluated Qwen3.5 and Qwen3-VL panels.

\begin{figure*}[t]
    \centering
    \begin{subfigure}[t]{0.32\textwidth}
        \centering
        \includegraphics[width=\linewidth]{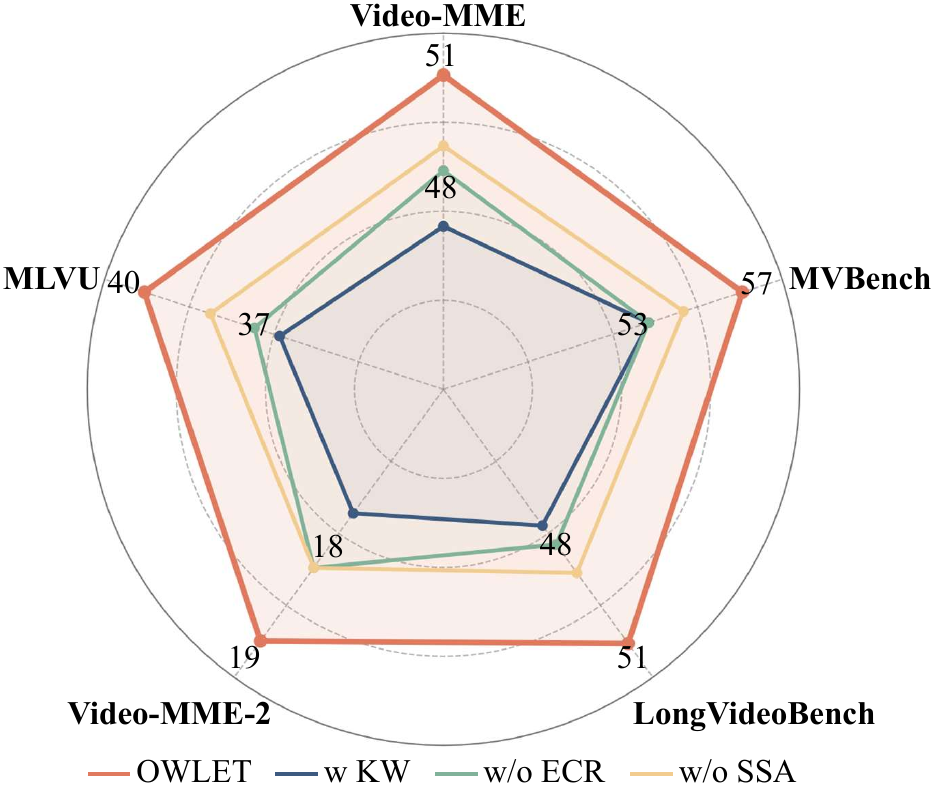}
        \caption{$16$ frames}
    \end{subfigure}\hfill
    \begin{subfigure}[t]{0.32\textwidth}
        \centering
        \includegraphics[width=\linewidth]{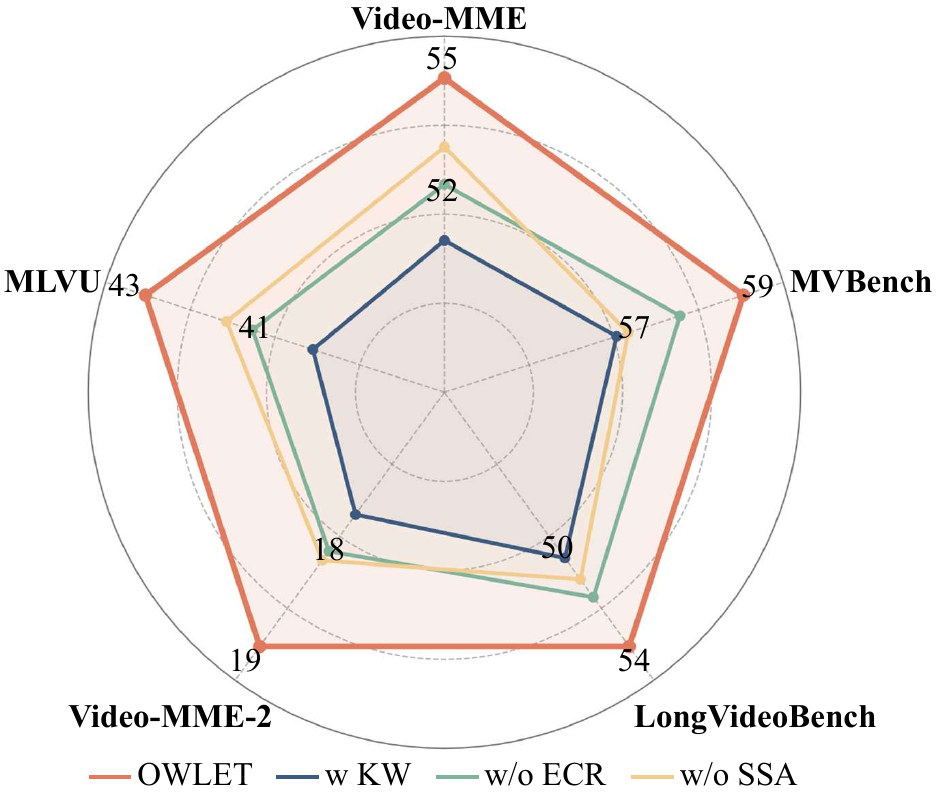}
        \caption{$32$ frames}
    \end{subfigure}\hfill
    \begin{subfigure}[t]{0.32\textwidth}
        \centering
        \includegraphics[width=\linewidth]{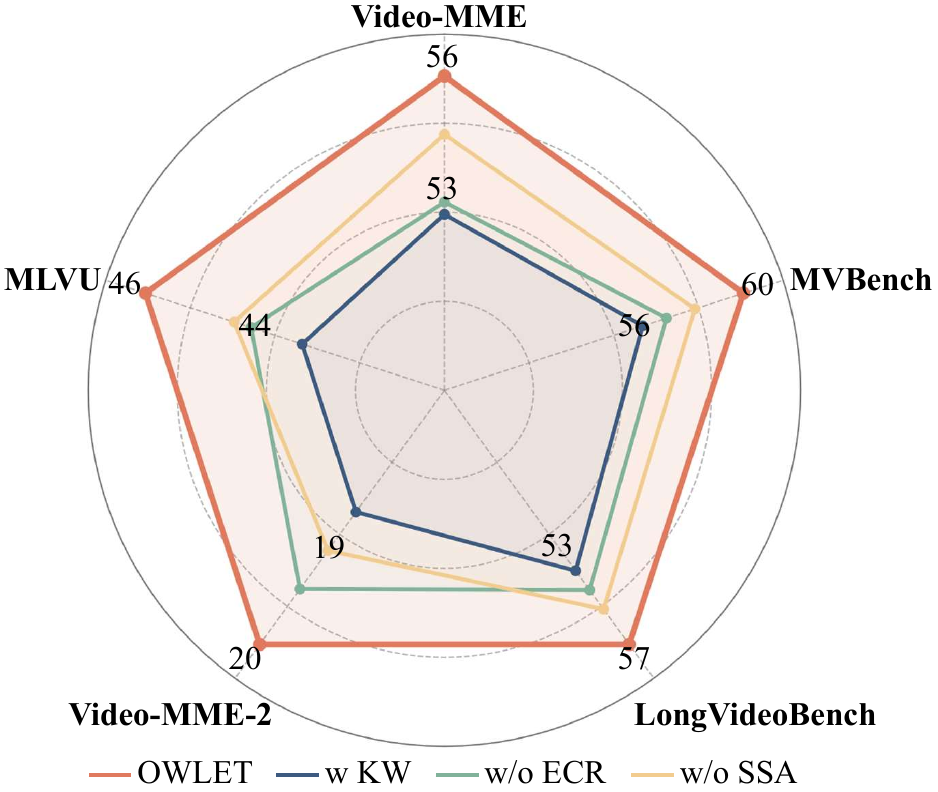}
        \caption{$64$ frames}
    \end{subfigure}
    \caption{\textbf{Ablation results for Qwen3.5-0.8B.}
    Panels (a)--(c) use $16$, $32$, and $64$ student frames, respectively.
    The five axes report accuracy (\%) on Video-MME, MVBench, LongVideoBench, Video-MME~2, and MLVU. The legend defines the key-window, w/o~ECR, and w/o~SSA variants described in the ablation text; evaluation uses uniformly sampled student frames.}
    \label{app:ablation-qwen35-08b}
\end{figure*}

\begin{figure*}[t]
    \centering
    \begin{subfigure}[t]{0.32\textwidth}
        \centering
        \includegraphics[width=\linewidth]{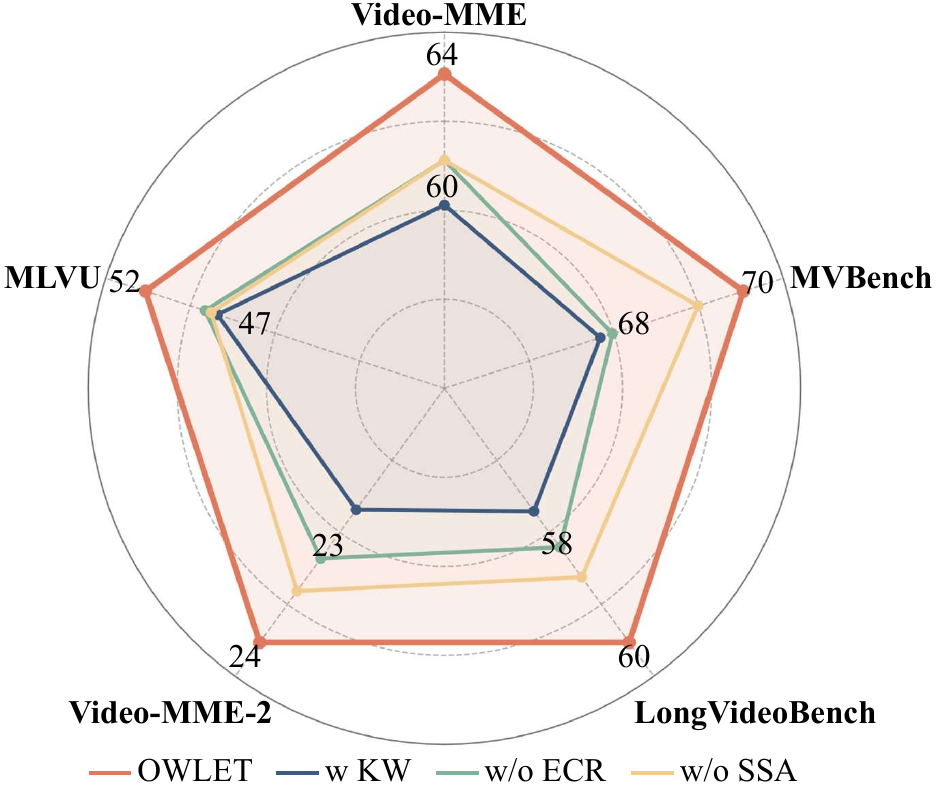}
        \caption{$16$ frames}
    \end{subfigure}\hfill
    \begin{subfigure}[t]{0.32\textwidth}
        \centering
        \includegraphics[width=\linewidth]{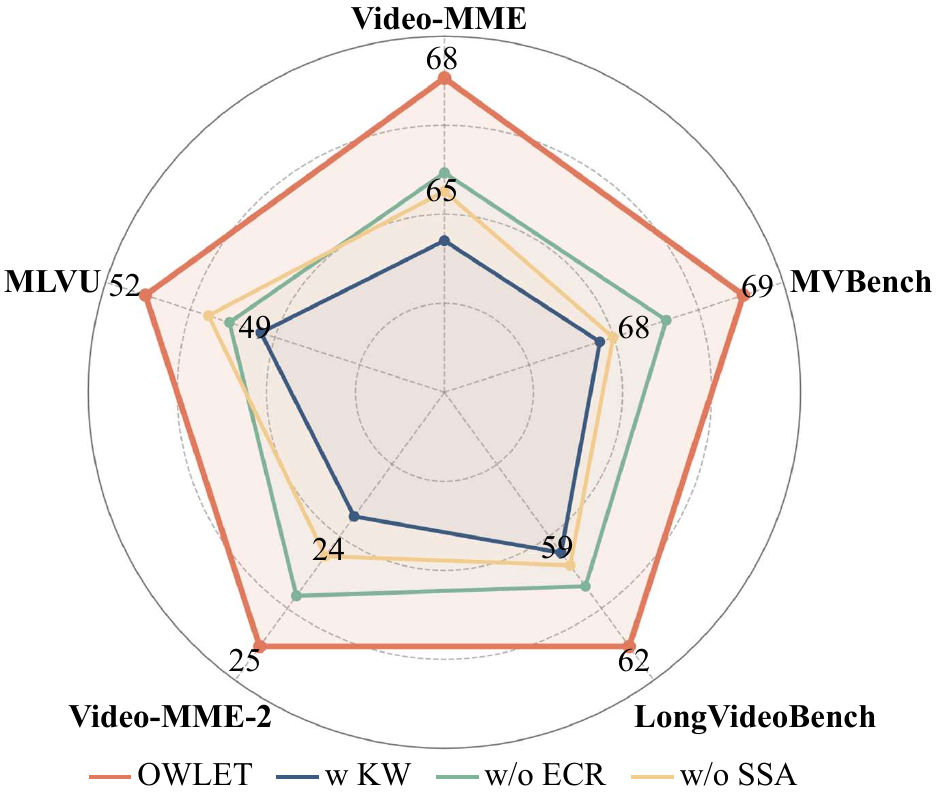}
        \caption{$32$ frames}
    \end{subfigure}\hfill
    \begin{subfigure}[t]{0.32\textwidth}
        \centering
        \includegraphics[width=\linewidth]{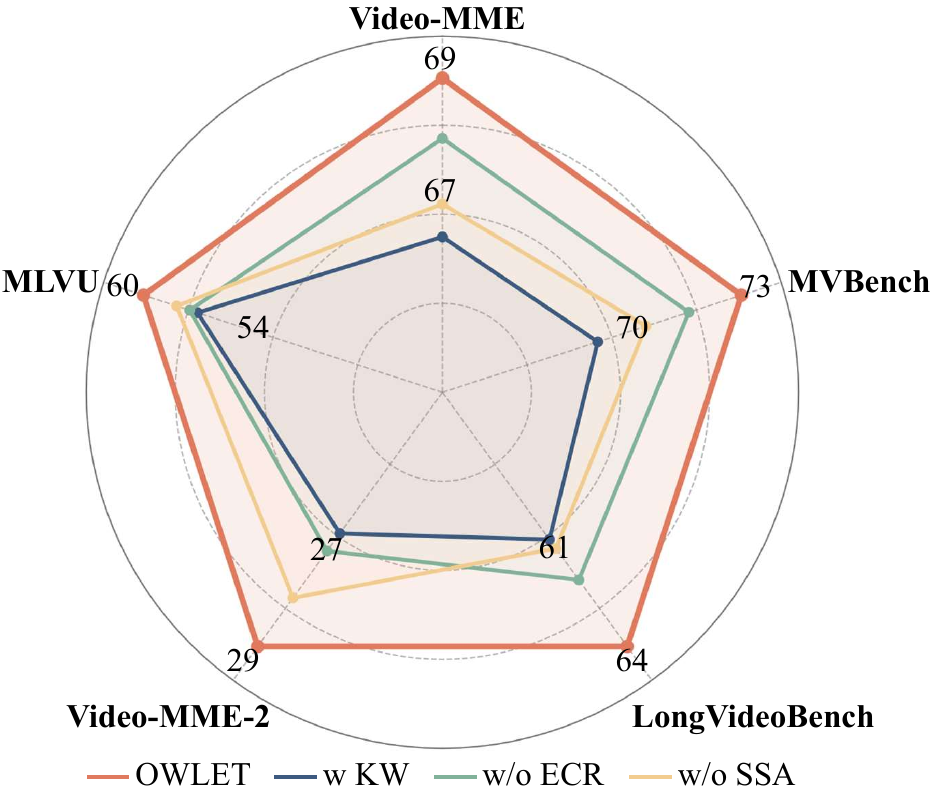}
        \caption{$64$ frames}
    \end{subfigure}
    \caption{\textbf{Ablation results for Qwen3.5-4B.}
    Panels (a)--(c) correspond to $16$, $32$, and $64$ uniformly sampled student frames.}
    \label{app:ablation-qwen35-4b}
\end{figure*}

\begin{figure*}[t]
    \centering
    \begin{subfigure}[t]{0.32\textwidth}
        \centering
        \includegraphics[width=\linewidth]{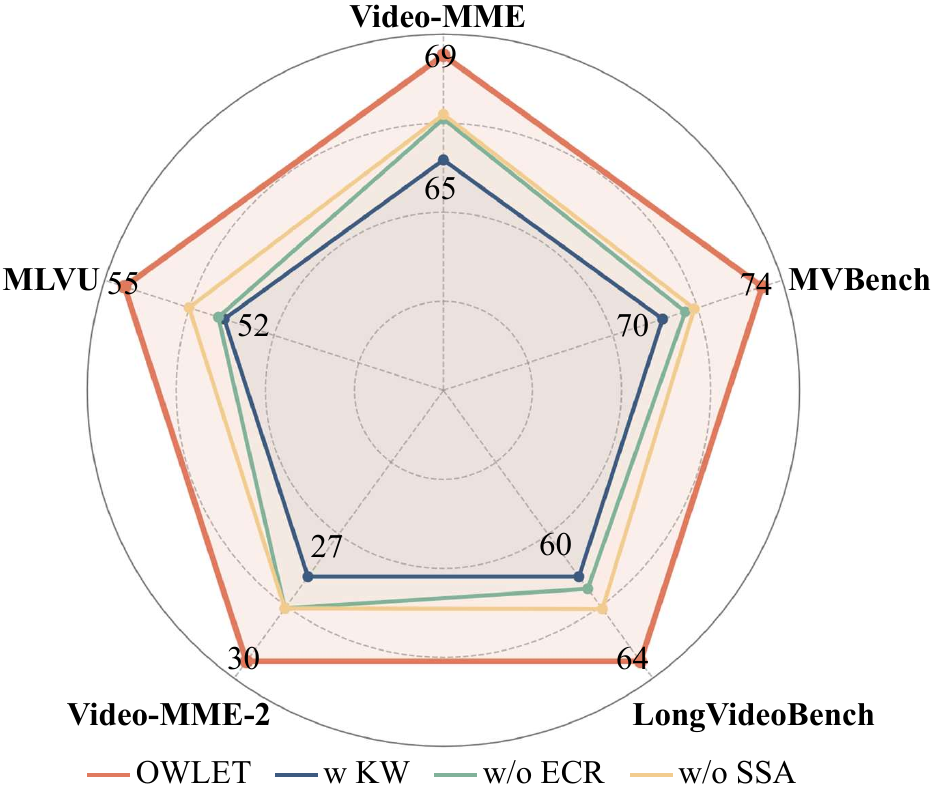}
        \caption{$16$ frames}
    \end{subfigure}\hfill
    \begin{subfigure}[t]{0.32\textwidth}
        \centering
        \includegraphics[width=\linewidth]{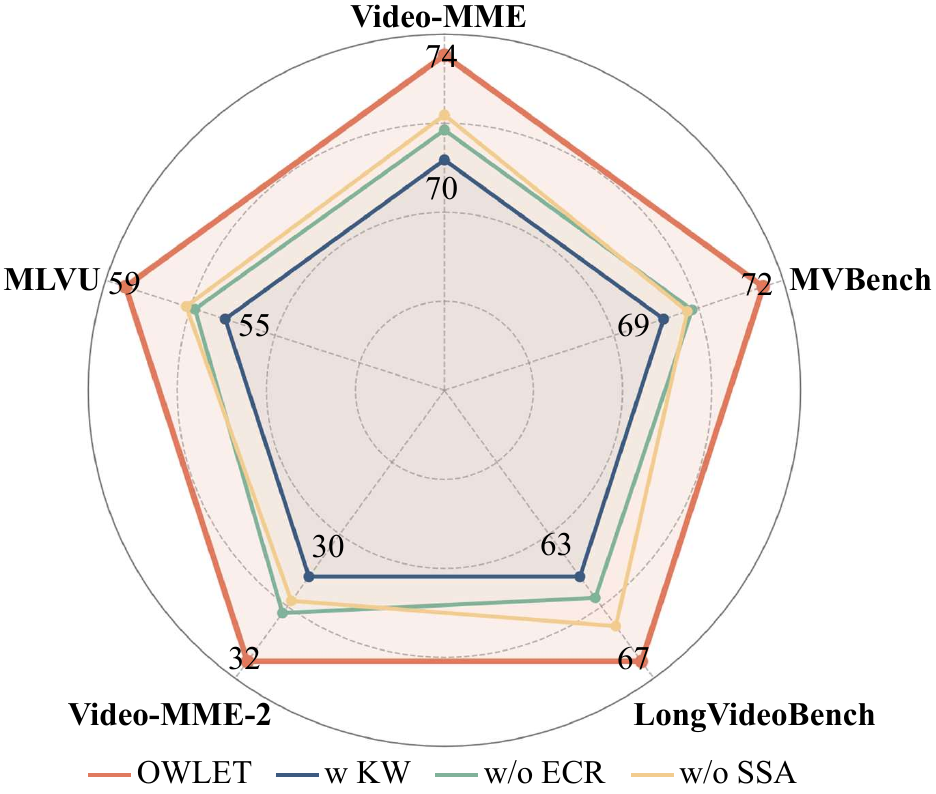}
        \caption{$32$ frames}
    \end{subfigure}\hfill
    \begin{subfigure}[t]{0.32\textwidth}
        \centering
        \includegraphics[width=\linewidth]{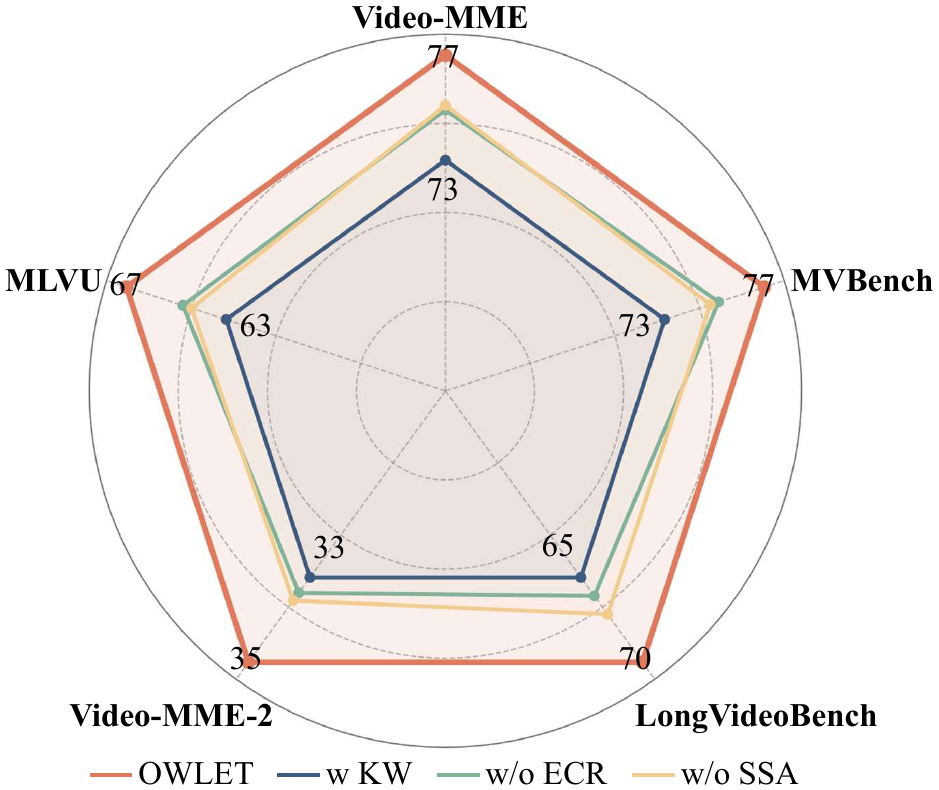}
        \caption{$64$ frames}
    \end{subfigure}
    \caption{\textbf{Ablation results for Qwen3.5-27B.}
    Panels (a)--(c) correspond to $16$, $32$, and $64$ uniformly sampled student frames.}
    \label{app:ablation-qwen35-27b}
\end{figure*}

\begin{figure*}[t]
    \centering
    \begin{subfigure}[t]{0.32\textwidth}
        \centering
        \includegraphics[width=\linewidth]{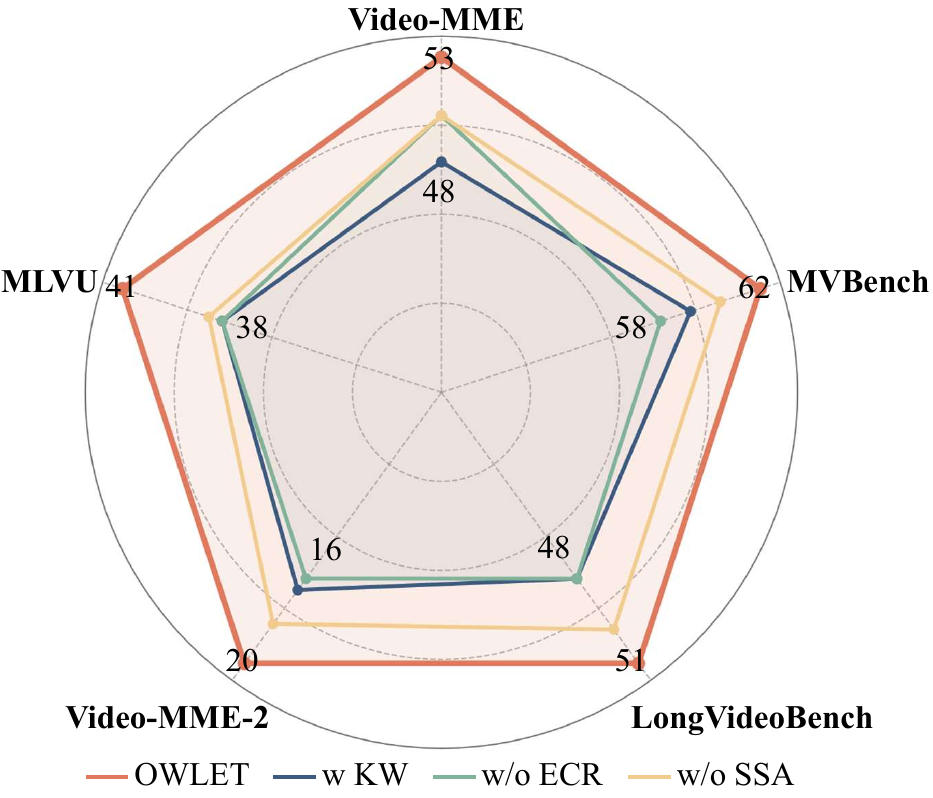}
        \caption{$16$ frames}
    \end{subfigure}\hfill
    \begin{subfigure}[t]{0.32\textwidth}
        \centering
        \includegraphics[width=\linewidth]{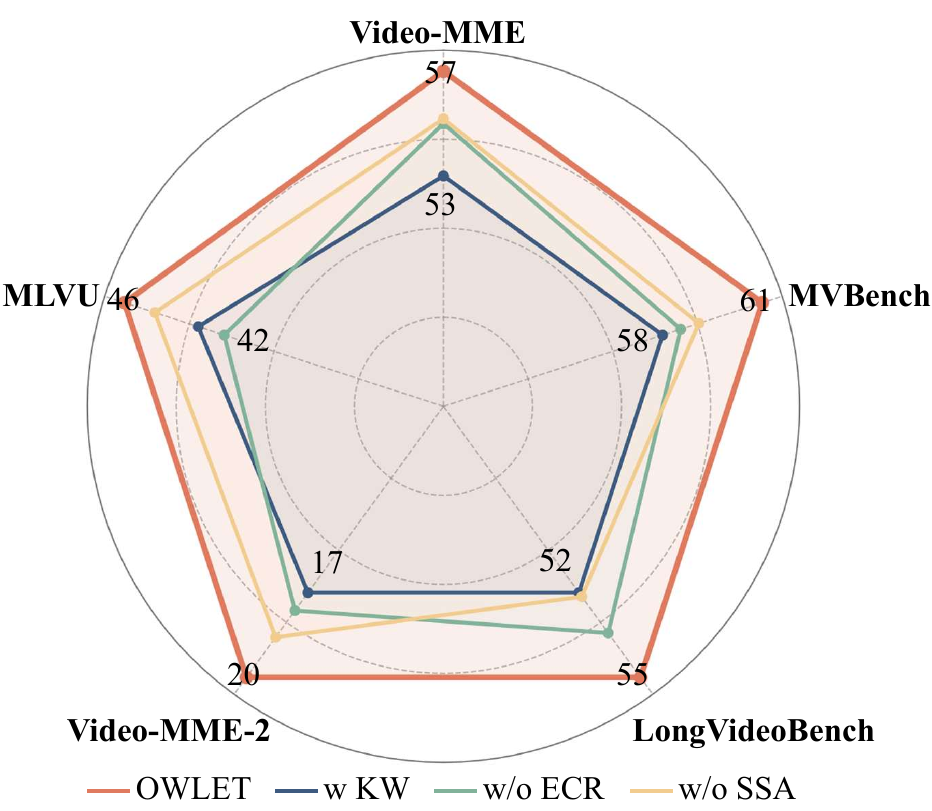}
        \caption{$32$ frames}
    \end{subfigure}\hfill
    \begin{subfigure}[t]{0.32\textwidth}
        \centering
        \includegraphics[width=\linewidth]{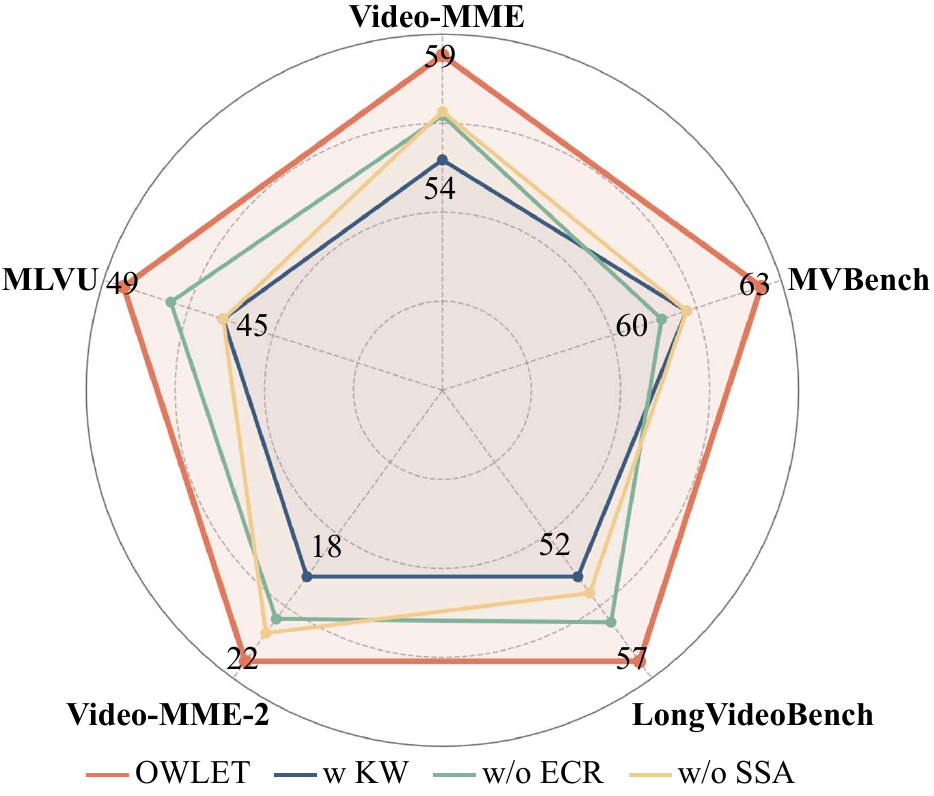}
        \caption{$64$ frames}
    \end{subfigure}
    \caption{\textbf{Ablation results for Qwen3-VL-2B in non-thinking mode.}
    Panels (a)--(c) correspond to $16$, $32$, and $64$ uniformly sampled student frames.}
    \label{app:ablation-qwen3vl-2b-non}
\end{figure*}

\begin{figure*}[t]
    \centering
    \begin{subfigure}[t]{0.32\textwidth}
        \centering
        \includegraphics[width=\linewidth]{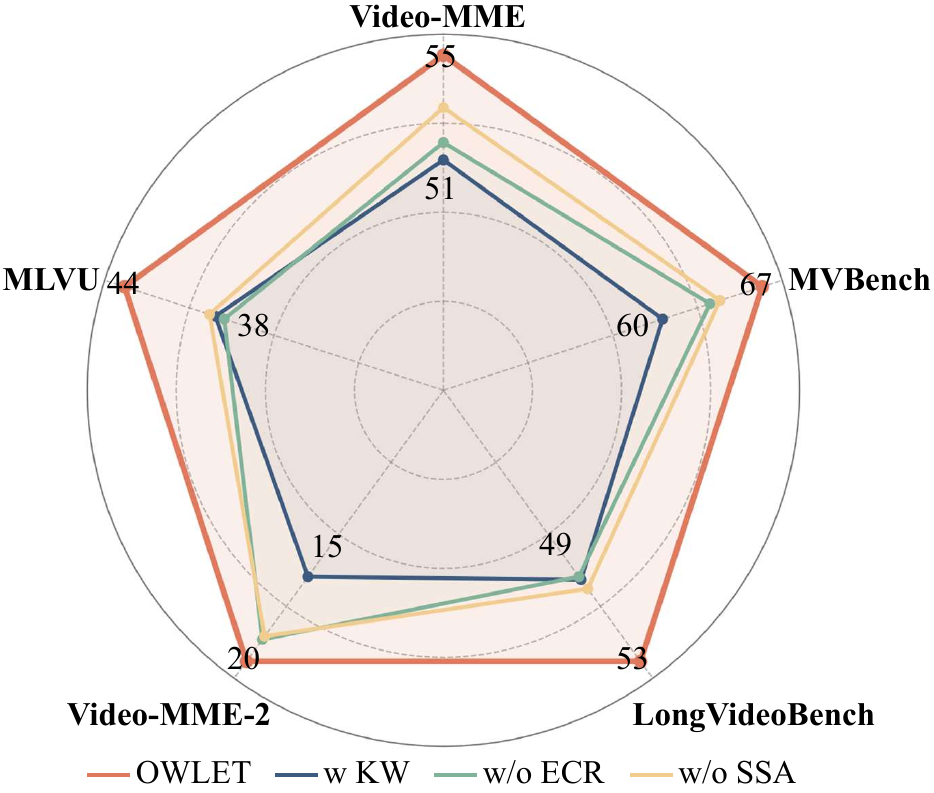}
        \caption{$16$ frames}
    \end{subfigure}\hfill
    \begin{subfigure}[t]{0.32\textwidth}
        \centering
        \includegraphics[width=\linewidth]{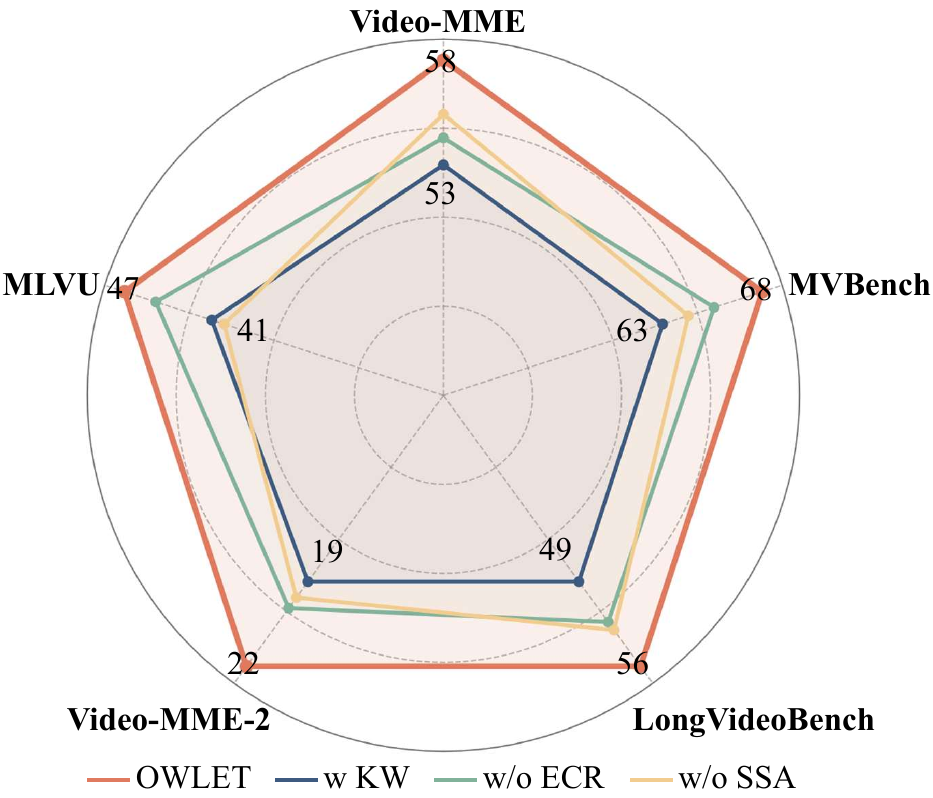}
        \caption{$32$ frames}
    \end{subfigure}\hfill
    \begin{subfigure}[t]{0.32\textwidth}
        \centering
        \includegraphics[width=\linewidth]{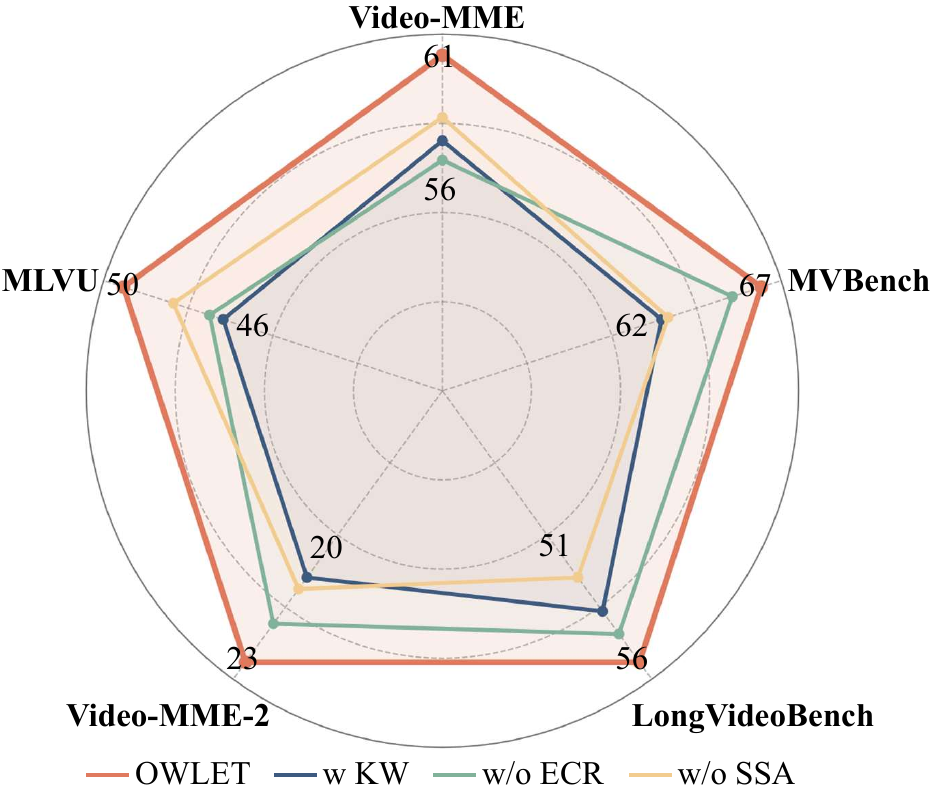}
        \caption{$64$ frames}
    \end{subfigure}
    \caption{\textbf{Ablation results for Qwen3-VL-2B in thinking mode.}
    Panels (a)--(c) correspond to $16$, $32$, and $64$ uniformly sampled student frames.}
    \label{app:ablation-qwen3vl-2b-thinking}
\end{figure*}

\begin{figure*}[t]
    \centering
    \begin{subfigure}[t]{0.32\textwidth}
        \centering
        \includegraphics[width=\linewidth]{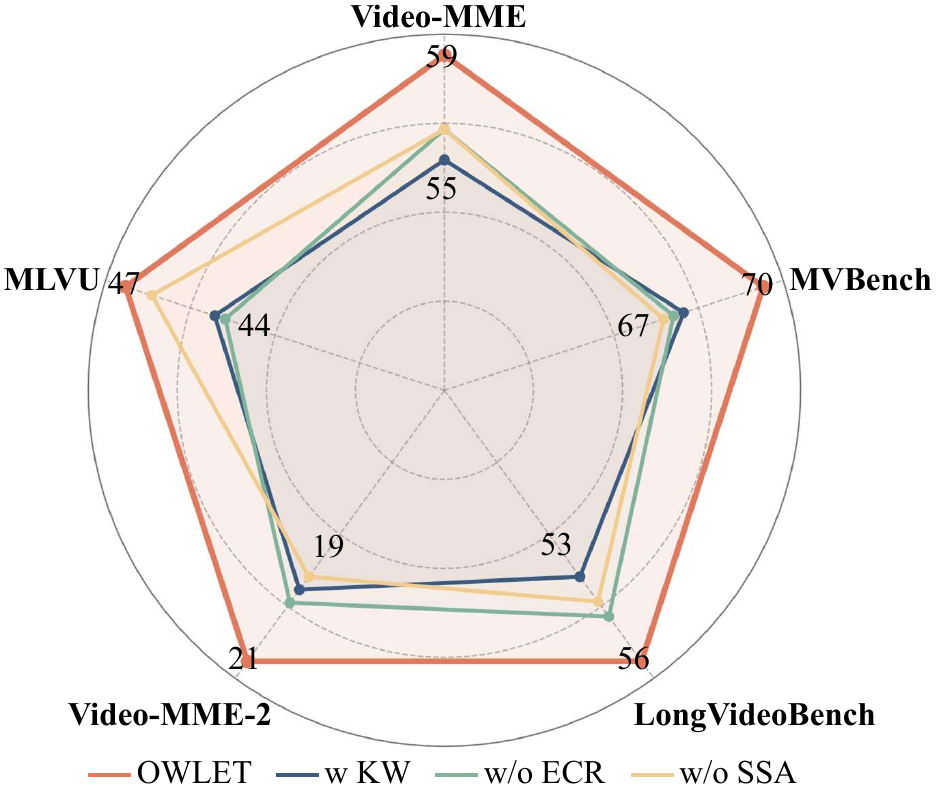}
        \caption{$16$ frames}
    \end{subfigure}\hfill
    \begin{subfigure}[t]{0.32\textwidth}
        \centering
        \includegraphics[width=\linewidth]{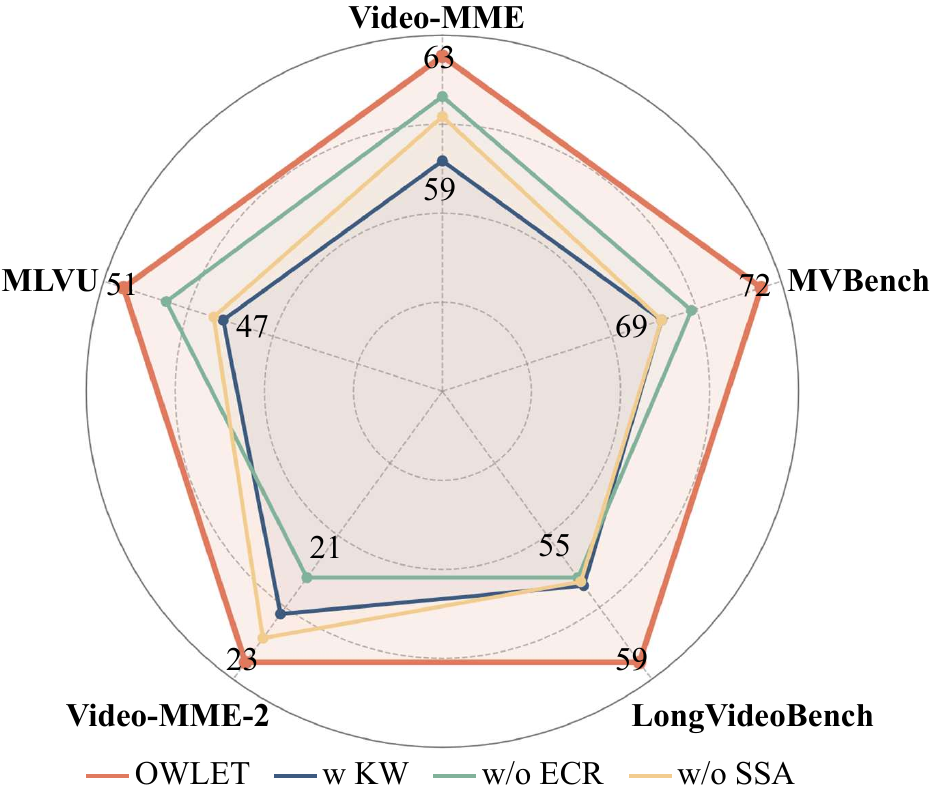}
        \caption{$32$ frames}
    \end{subfigure}\hfill
    \begin{subfigure}[t]{0.32\textwidth}
        \centering
        \includegraphics[width=\linewidth]{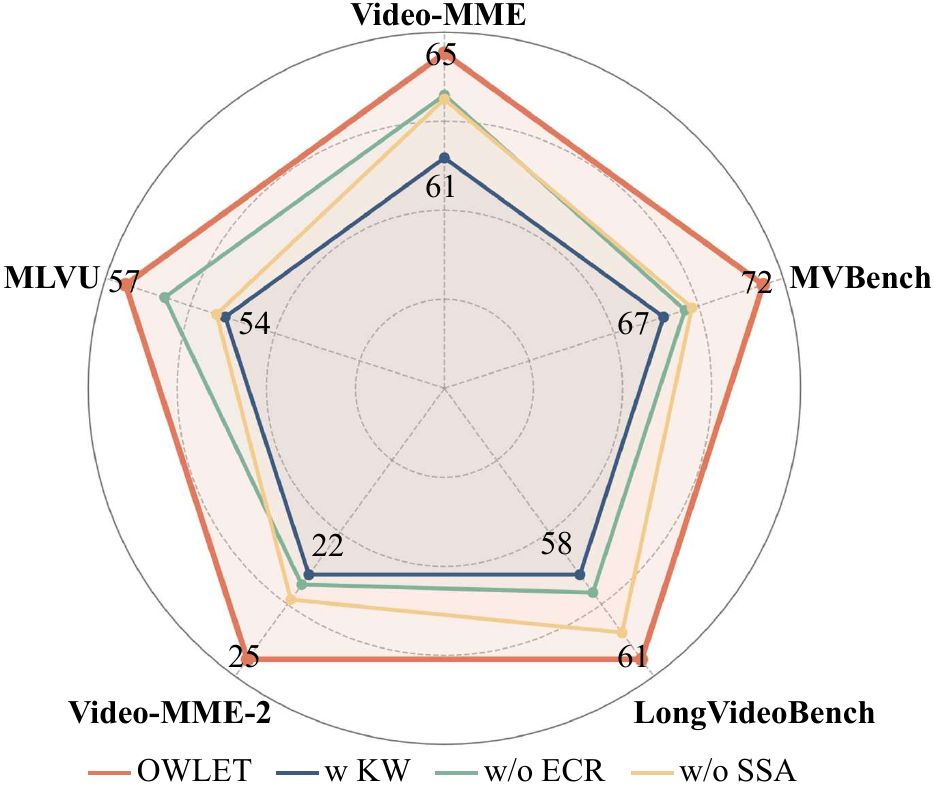}
        \caption{$64$ frames}
    \end{subfigure}
    \caption{\textbf{Ablation results for Qwen3-VL-4B in non-thinking mode.}
    Panels (a)--(c) correspond to $16$, $32$, and $64$ uniformly sampled student frames.}
    \label{app:ablation-qwen3vl-4b-non}
\end{figure*}

\begin{figure*}[t]
    \centering
    \begin{subfigure}[t]{0.32\textwidth}
        \centering
        \includegraphics[width=\linewidth]{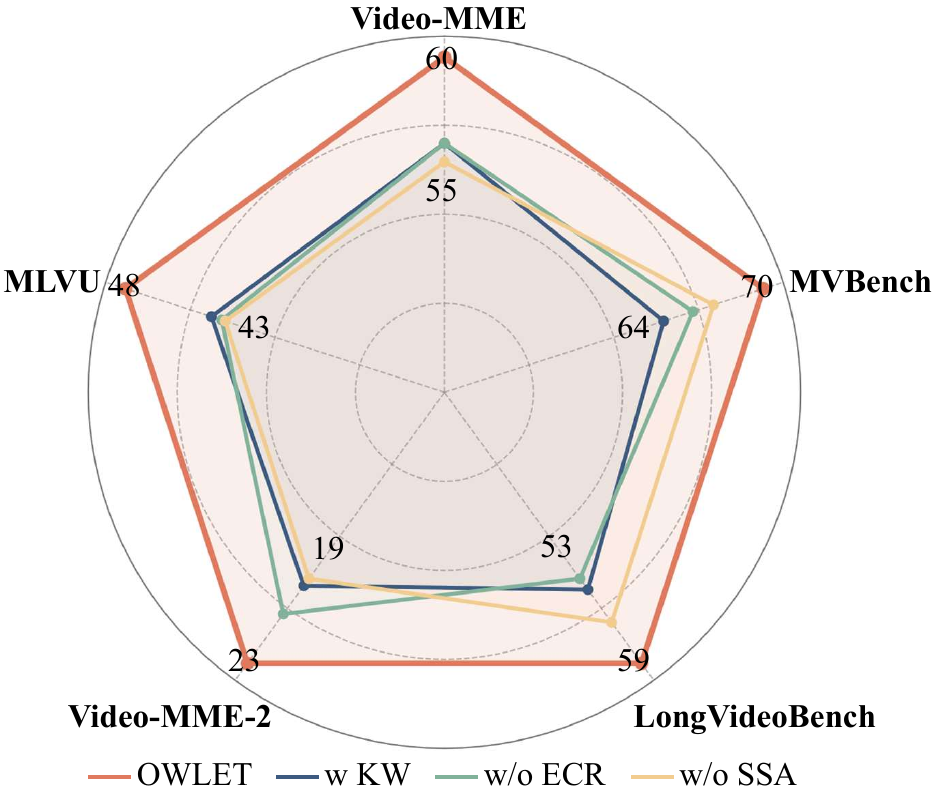}
        \caption{$16$ frames}
    \end{subfigure}\hfill
    \begin{subfigure}[t]{0.32\textwidth}
        \centering
        \includegraphics[width=\linewidth]{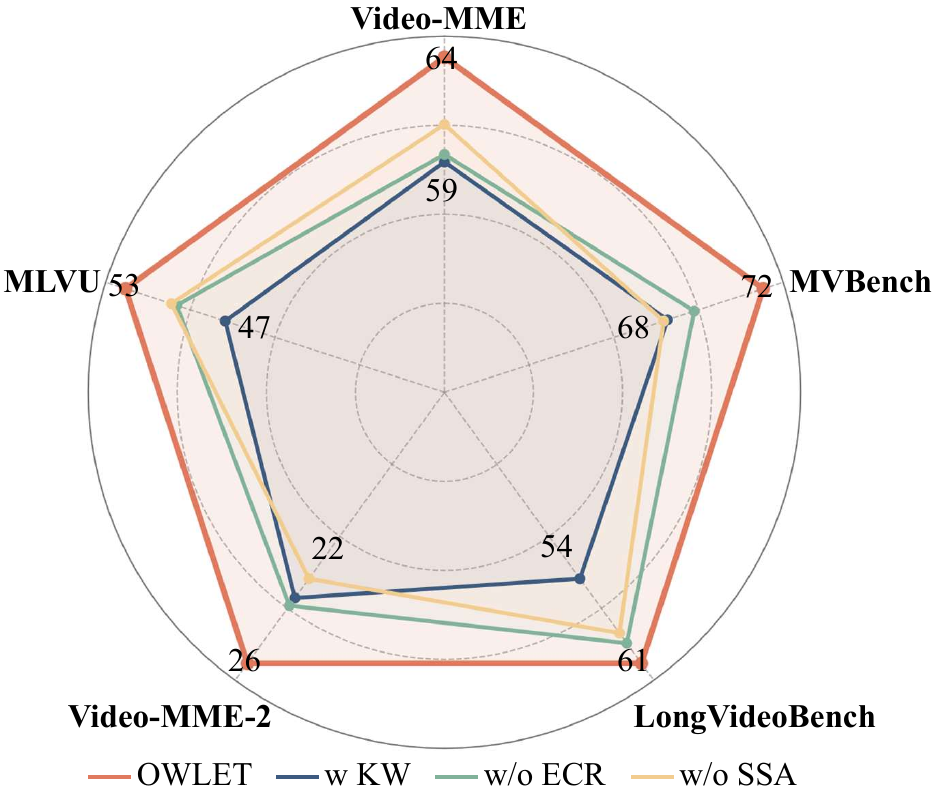}
        \caption{$32$ frames}
    \end{subfigure}\hfill
    \begin{subfigure}[t]{0.32\textwidth}
        \centering
        \includegraphics[width=\linewidth]{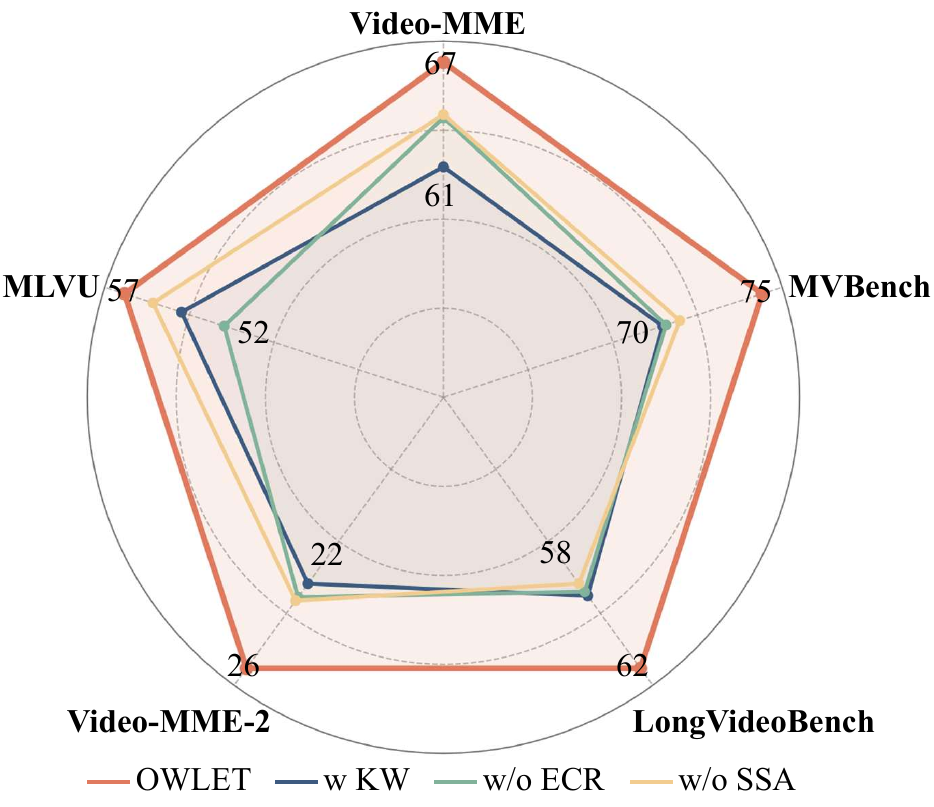}
        \caption{$64$ frames}
    \end{subfigure}
    \caption{\textbf{Ablation results for Qwen3-VL-4B in thinking mode.}
    Panels (a)--(c) correspond to $16$, $32$, and $64$ uniformly sampled student frames.}
    \label{app:ablation-qwen3vl-4b-thinking}
\end{figure*}

\begin{figure*}[t]
    \centering
    \begin{subfigure}[t]{0.32\textwidth}
        \centering
        \includegraphics[width=\linewidth]{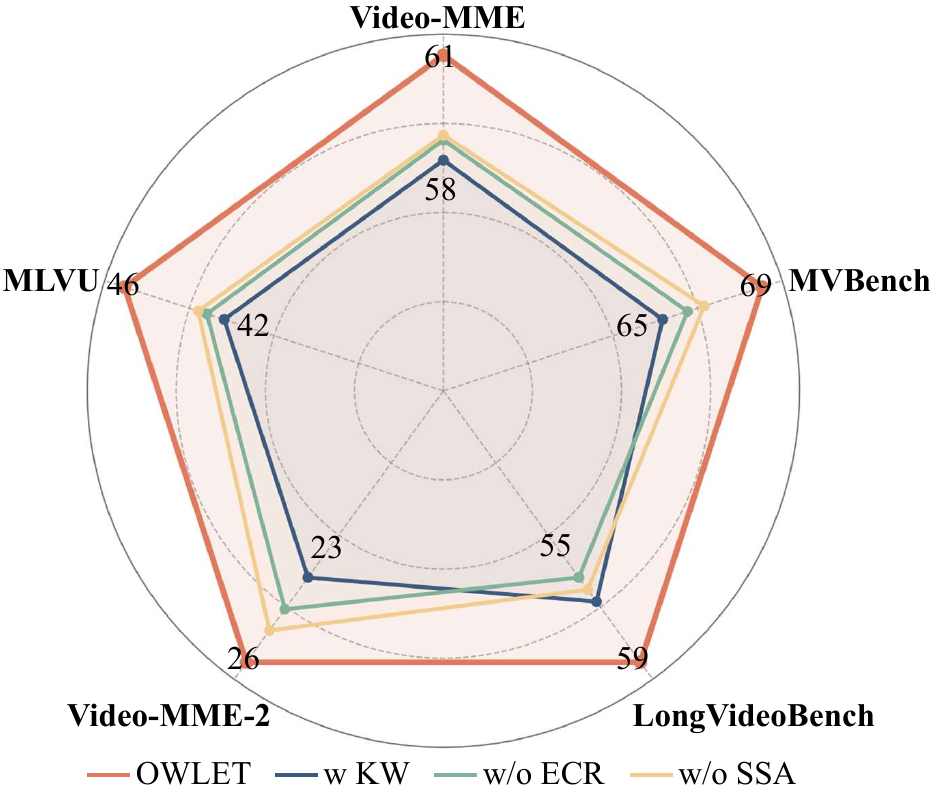}
        \caption{$16$ frames}
    \end{subfigure}\hfill
    \begin{subfigure}[t]{0.32\textwidth}
        \centering
        \includegraphics[width=\linewidth]{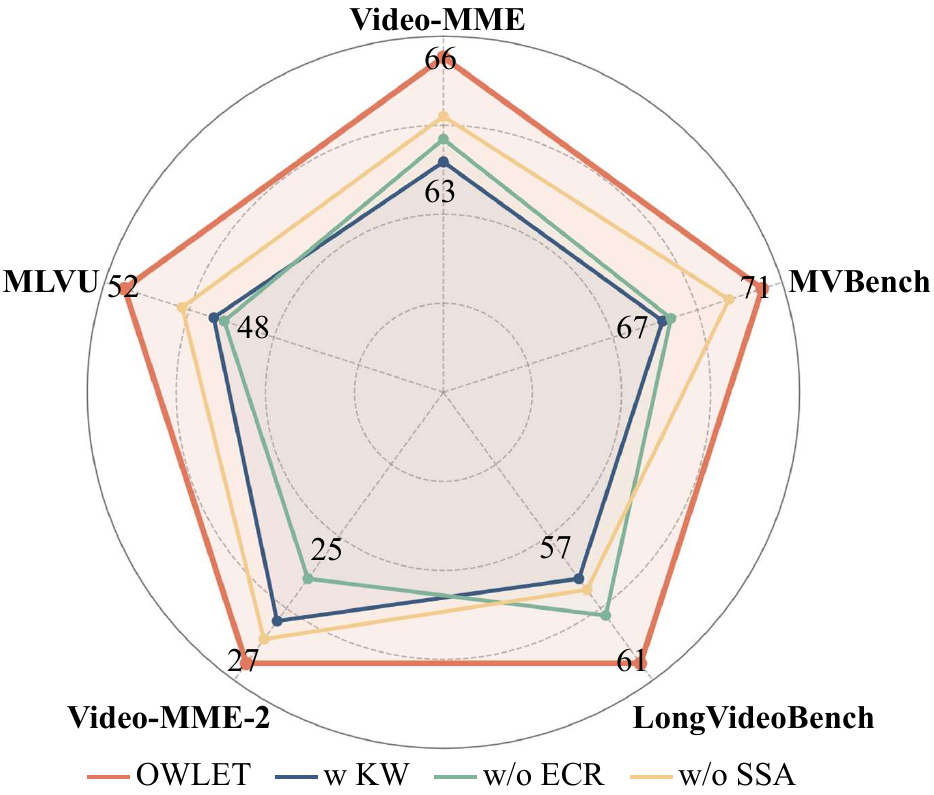}
        \caption{$32$ frames}
    \end{subfigure}\hfill
    \begin{subfigure}[t]{0.32\textwidth}
        \centering
        \includegraphics[width=\linewidth]{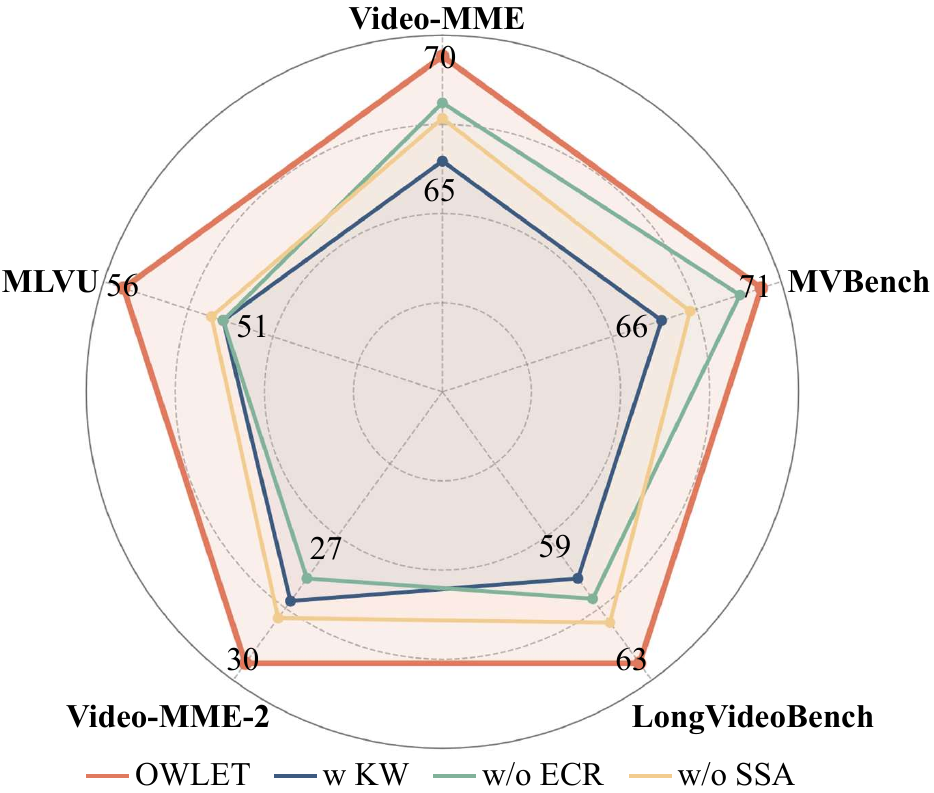}
        \caption{$64$ frames}
    \end{subfigure}
    \caption{\textbf{Ablation results for Qwen3-VL-8B in non-thinking mode.}
    Panels (a)--(c) correspond to $16$, $32$, and $64$ uniformly sampled student frames.}
    \label{app:ablation-qwen3vl-8b-non}
\end{figure*}

\begin{figure*}[t]
    \centering
    \begin{subfigure}[t]{0.32\textwidth}
        \centering
        \includegraphics[width=\linewidth]{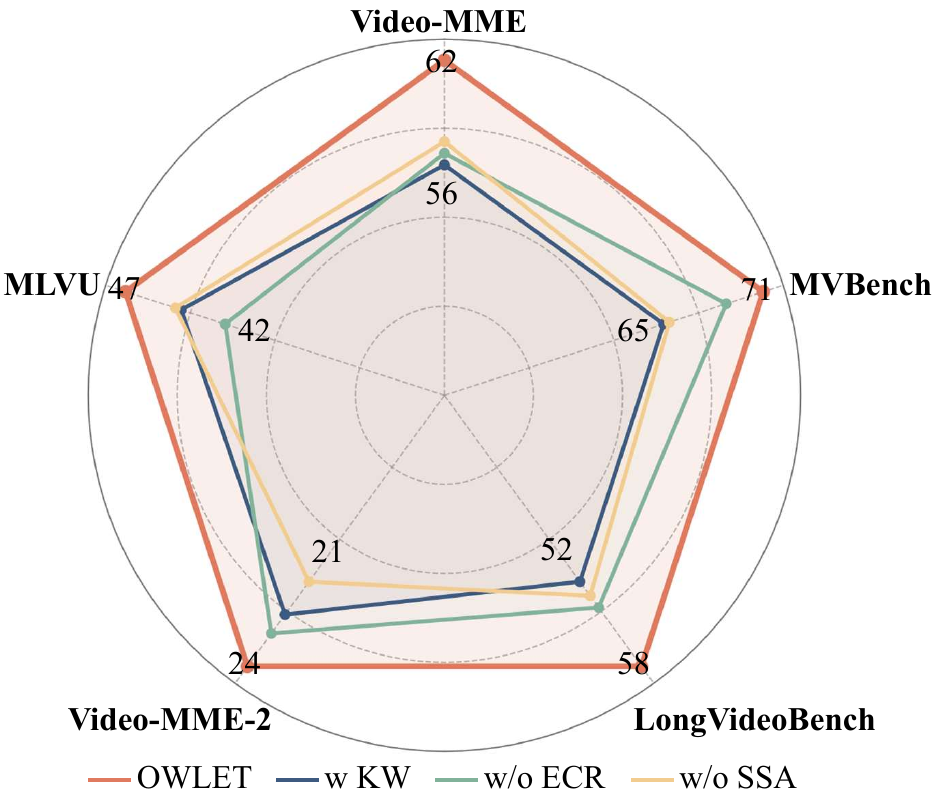}
        \caption{$16$ frames}
    \end{subfigure}\hfill
    \begin{subfigure}[t]{0.32\textwidth}
        \centering
        \includegraphics[width=\linewidth]{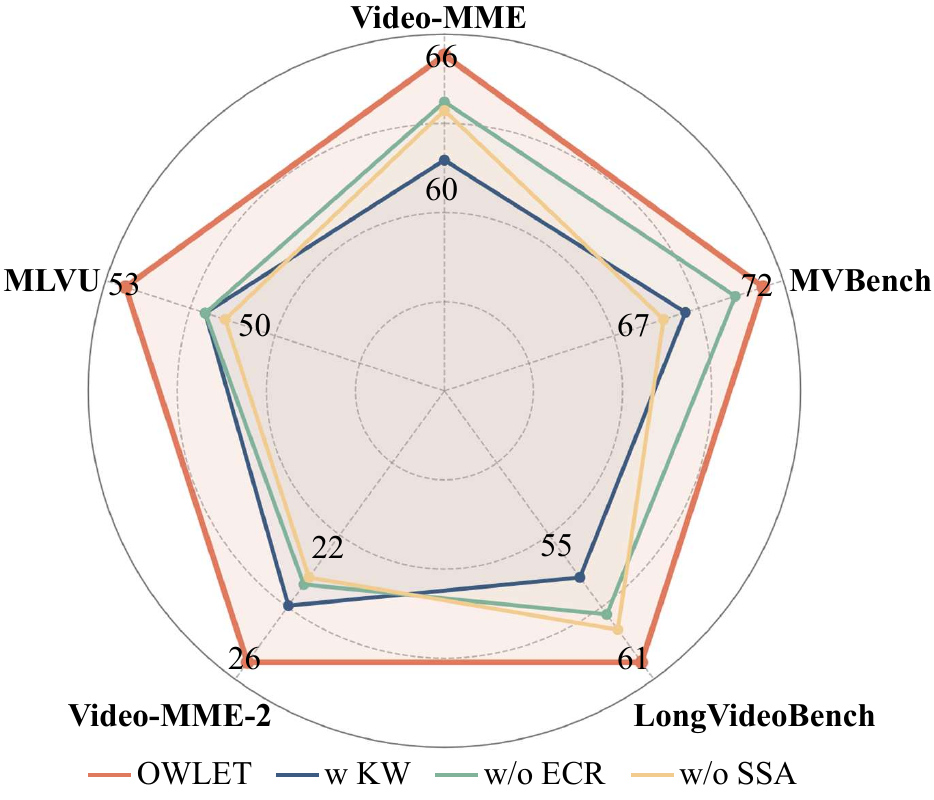}
        \caption{$32$ frames}
    \end{subfigure}\hfill
    \begin{subfigure}[t]{0.32\textwidth}
        \centering
        \includegraphics[width=\linewidth]{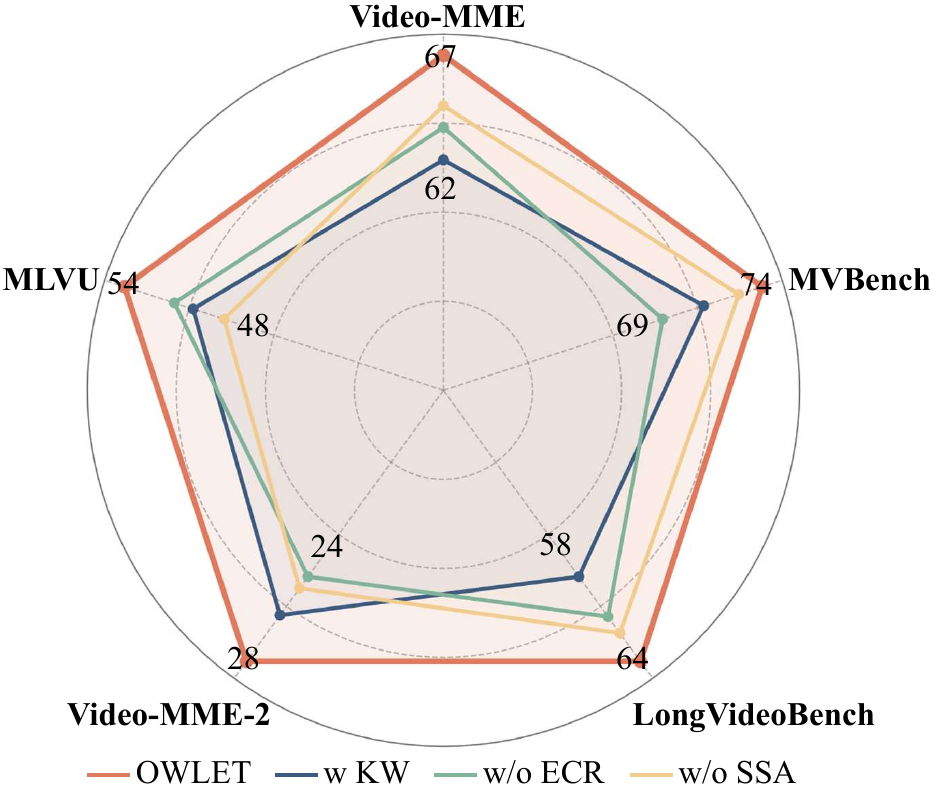}
        \caption{$64$ frames}
    \end{subfigure}
    \caption{\textbf{Ablation results for Qwen3-VL-8B in thinking mode.}
    Panels (a)--(c) correspond to $16$, $32$, and $64$ uniformly sampled student frames.}
    \label{app:ablation-qwen3vl-8b-thinking}
\end{figure*}
\section{Additional Analyses}
\label{app:additional-analyses}

This section reports nine analyses covering alternative distillation objectives, teacher and SSA settings, token-level credit, qualitative responses, training cost and response length, and post-training behavior beyond the standard recipe. Unless stated otherwise, every analysis uses Qwen3.5-9B with a 64-frame student view and the training configuration of Appendix~\ref{app:training-hparams}.

\subsection{Comparison with On-Policy Distillation Variants}
\label{app:opd-variants}

Section~\ref{sec:main_res} compares \method against OPSD with the token-level KL objective of Equation~\ref{eq:opsd}. Here we compare alternative output divergences and distillation--GRPO schedules on Qwen3.5-9B with $64$ uniformly sampled student frames. All methods receive the same training data and student rollout budget, and all teacher-assisted objectives use the teacher view combining global context with local windows. Evaluation uses the ordinary student view without evidence annotations or teacher computation.

\paragraph{Compared objectives.}
We retain the standard OPSD objective as the reference and consider three variants:
\begin{itemize}[itemsep=2pt, leftmargin=*]
\item \textbf{OPSD-JSD} replaces the student-to-teacher KL in Equation~\ref{eq:opsd} with the generalized Jensen--Shannon divergence, using an interpolation coefficient of $\alpha=0.5$. This comparison changes the output divergence while preserving on-policy sampling from the student.
\item \textbf{OPSD+GRPO} combines the distillation loss with the GRPO objective of Equation~\ref{eq:rl}, using an equal weight. The GRPO term applies a response-level advantage uniformly across tokens; it does not use evidence-aware credit redistribution or hidden-state alignment.
\item \textbf{OPSD-Anneal} combines distillation with GRPO using equal initial weights for the two objectives, and linearly decays the distillation weight to zero, so that training ends in pure GRPO. This variant tests a schedule that uses teacher supervision early and leaves GRPO as the sole objective at the end.
\end{itemize}
These variants vary how output-distribution supervision is defined or combined with the response-level reward.

\begin{table}[h]
\centering
\caption{Accuracy (\%) of on-policy distillation variants and \method on Qwen3.5-9B at 64 student frames. All methods use the same teacher view, training data, and student rollout budget. Avg. is the unweighted mean over the five benchmarks. The best result in each column is shown in bold.}
\label{app:tab-opd-variants}
\resizebox{\textwidth}{!}{
\setlength{\tabcolsep}{4pt}
\begin{tabular}{lcccccc}
\toprule
Method & Video-MME & MVBench & LongVideoBench & Video-MME 2 & MLVU & Avg. \\
\midrule
OPSD & 66.1 & 68.7 & 62.9 & 26.7 & 53.3 & 55.5 \\
OPSD-JSD & 65.3 & 68.2 & 61.5 & 25.9 & 53.7 & 54.9 \\
OPSD+GRPO & 66.3 & 67.8 & 63.3 & 27.0 & 52.7 & 55.4 \\
OPSD-Anneal & 67.1 & 68.5 & 63.4 & 27.1 & 55.3 & 56.3 \\
\green \method & \green \textbf{70.7} & \green \textbf{72.8} & \green \textbf{65.2} & \green \textbf{28.9} & \green \textbf{61.6} & \green \textbf{59.8} \\
\bottomrule
\end{tabular}
}
\end{table}

\paragraph{Overall comparison.}
Table~\ref{app:tab-opd-variants} shows that the three modifications have different effects. OPSD-Anneal achieves the highest macro-average among the distillation configurations, at $56.3$, followed by standard OPSD at $55.5$, OPSD+GRPO at $55.4$, and OPSD-JSD at $54.9$. Thus, neither using the Jensen--Shannon divergence nor adding GRPO yields an aggregate improvement over standard OPSD in this setting. Annealing improves the macro-average, but \method remains ahead at $59.8$ and obtains the highest accuracy on every benchmark. All benchmark differences discussed below are absolute percentage-point differences.

\paragraph{Effect of the divergence choice.}
OPSD-JSD scores lower than standard reverse-KL OPSD on four of the five benchmarks: Video-MME decreases from $66.1$ to $65.3$, MVBench from $68.7$ to $68.2$, LongVideoBench from $62.9$ to $61.5$, and Video-MME~2 from $26.7$ to $25.9$. MLVU is the exception, increasing from $53.3$ to $53.7$. The resulting macro-average is $54.9$, compared with $55.5$ for standard OPSD.

\paragraph{Combining distillation with GRPO.}
OPSD+GRPO improves over standard OPSD on Video-MME, LongVideoBench, and Video-MME~2 by $0.2$, $0.4$, and $0.3$ points, respectively. These gains are offset by decreases of $0.9$ points on MVBench and $0.6$ points on MLVU, leaving the macro-average at $55.4$, close to the $55.5$ of standard OPSD. Adding the GRPO term changes which benchmarks benefit without improving the overall average in this run.

\paragraph{Annealing the distillation objective.}
OPSD-Anneal outperforms both OPSD-JSD and OPSD+GRPO on all five benchmarks and is the strongest distillation variant by macro-average. Relative to standard OPSD, it improves Video-MME by $1.0$ point, LongVideoBench by $0.5$, Video-MME~2 by $0.4$, and MLVU by $2.0$, while MVBench decreases slightly from $68.7$ to $68.5$. The macro-average rises from $55.5$ to $56.3$.

\paragraph{Comparison with \method.}
Against OPSD-Anneal, \method improves Video-MME, MVBench, LongVideoBench, Video-MME~2, and MLVU by $3.6$, $4.3$, $1.8$, $1.8$, and $6.3$ points, respectively. Its macro-average is $59.8$, compared with $56.3$ for OPSD-Anneal, for a $3.5$-point advantage.

\subsection{Teacher Parameterization}
\label{app:teacher-parameterization}

The teacher of \method shares the student's architecture and supplies token scores and hidden-state targets from the teacher view combining global context with local windows. We compare the default fixed teacher with an exponential moving average (EMA) teacher on Qwen3.5-9B with $64$ student frames. The two settings use the same teacher-view construction, training data, student rollout budget, and ECR and SSA objectives; only the teacher parameters differ.

\paragraph{Fixed and EMA teachers.}
The fixed teacher retains $\bar\theta=\theta_0$ throughout the entire training run, where $\theta_0$ is the common checkpoint used to initialize the teacher and student. The EMA variant instead maintains an exponential moving average of the student parameters with decay $0.99$:
\[
\bar\theta \leftarrow 0.99\,\bar\theta + 0.01\,\theta.
\]
In both settings, the teacher scores student-generated responses under the privileged view and supplies the teacher term in the evidence contrast of Equation~\ref{eq:ecr-gap} and the representation targets for SSA in Equation~\ref{eq:ssa}. The fixed teacher's scores and hidden states are recomputed for each student response, while its weights remain unchanged. Evaluation uses only the student with uniformly sampled frames.

\begin{table}[h]
\centering
\caption{Effect of the teacher parameterization on Qwen3.5-9B at 64 input frames. The fixed teacher retains the initial checkpoint throughout training, while the EMA teacher averages student parameters with decay $0.99$ under the same teacher view and training configuration. Scores are accuracies (\%), and Avg. is the unweighted mean over the five benchmarks.}
\label{app:tab-teacher-param}
\resizebox{\textwidth}{!}{
\setlength{\tabcolsep}{4pt}
\begin{tabular}{lcccccc}
\toprule
Teacher $\bar\theta$ & Video-MME & MVBench & LongVideoBench & Video-MME 2 & MLVU & Avg. \\
\midrule
EMA (decay $0.99$) & 69.3 & 70.5 & 64.7 & 27.9 & 60.8 & 58.6 \\
\green Fixed (default) &\green  \textbf{70.7} &\green  \textbf{72.8} & \green \textbf{65.2} & \green \textbf{28.9} &\green  \textbf{61.6} &\green  \textbf{59.8} \\
\bottomrule
\end{tabular}
}
\end{table}

\paragraph{Results across benchmarks.}
Table~\ref{app:tab-teacher-param} shows that the fixed teacher outperforms the EMA teacher on all five benchmarks. Its accuracies are $70.7$, $72.8$, $65.2$, $28.9$, and $61.6$ on Video-MME, MVBench, LongVideoBench, Video-MME~2, and MLVU, respectively, compared with $69.3$, $70.5$, $64.7$, $27.9$, and $60.8$ for EMA. The corresponding advantages are $1.4$, $2.3$, $0.5$, $1.0$, and $0.8$ percentage points. The largest difference occurs on MVBench, while LongVideoBench shows the smallest difference. The macro-average is $59.8$ for the fixed teacher and $58.6$ for EMA, a difference of $1.2$ points. Thus, retaining the initial teacher weights is favored across the evaluated tasks, with the magnitude of the advantage varying by benchmark.

\paragraph{Effect on the evidence contrast.}
With the fixed teacher, the first term in Equation~\ref{eq:ecr-gap} is evaluated under $\theta_0$ and the privileged view, while the second uses the evolving rollout-policy parameters $\theta_{\mathrm{old}}$ and the uniform view on the same question, response token, and prefix. At initialization the weights coincide; after student updates, the contrast includes both visual and parameter differences. With EMA, the teacher side also changes as student parameters are incorporated into the average, changing the token scores used by ECR.

\paragraph{Effect on representation targets.}
Teacher parameterization also affects SSA because the response-end target states are computed by the teacher. The fixed setting uses $\theta_0$ for these targets, whereas the EMA setting uses the evolving averaged parameters; both apply SSA only to verifier-accepted responses and backpropagate through the student.

\paragraph{Choice of the default teacher.}
The fixed teacher improves all five benchmarks over EMA with decay $0.99$, increasing the macro-average from $58.6$ to $59.8$. We use it as the default throughout the paper.

\subsection{Statistics of Advantage Signs and Highly Weighted Tokens}
\label{app:advantage-token-stats}

ECR redistributes only positive trajectory credit, so its reach depends on how often responses actually receive positive advantages, and its usefulness depends on which tokens end up carrying large weights. We measure both during the training run of Section~\ref{sec:main_res}.

\paragraph{Advantage signs.}
Table~\ref{app:tab-advantage-signs} reports, over four phases of training, the fraction of rollouts with positive, negative, and zero group-normalized advantages. With binary rewards and $G=8$ rollouts per prompt, zero advantages occur when all responses in a group receive the same reward. Verifier-accepted responses from these groups can still contribute to SSA.

The first phase contains $28.5\%$ positive-advantage rollouts, $7.2\%$ negative-advantage rollouts, and $64.3\%$ zero-advantage rollouts. By the final phase, these fractions become $4.7\%$, $2.2\%$, and $93.1\%$, respectively. Thus, the fraction carrying a nonzero group-normalized signal falls from $35.7\%$ to $6.9\%$, while the zero-advantage fraction increases by $28.8$ percentage points. The positive fraction, which is the part of the policy signal that ECR reweights through Equation~\ref{eq:ecr-profile}, decreases by $23.8$ percentage points across training; the negative fraction decreases in parallel by $5.0$ points. Averaging the four equally sized phases gives approximately $16.4\%$ positive, $4.8\%$ negative, and $78.9\%$ zero advantages over the analyzed run.

\begin{table}[h]
\centering
\caption{Fractions (\%) of rollouts with positive, negative, and zero advantages across training phases on Qwen3.5-9B at 64 input frames.}
\label{app:tab-advantage-signs}
\small
\begin{tabular}{lcccc}
\toprule
Advantage & Steps 0--250 & 250--500 & 500--750 & 750--1,000 \\
\midrule
$A^i>0$ & 28.5 & 21.8 & 10.6 & 4.7 \\
$A^i<0$ & 7.2 & 5.8 & 3.8 & 2.2 \\
$A^i=0$ & 64.3 & 72.4 & 85.6 & 93.1 \\
\bottomrule
\end{tabular}
\end{table}

\paragraph{Which tokens receive large weights.}
To characterize where the redistributed credit goes, we use a sample of $100$ question--video examples from the training split and evaluate the same diagnostic at three checkpoints. From each example, we retain the positively advantaged student responses and rank their valid response tokens by the redistributed weights $\widetilde A^i_t$. We compare the highest-weight and lowest-weight deciles and pool the selected token instances across the three checkpoints. Prompt tokens, padding, and special control tokens are excluded. The table therefore reports token-level composition, rather than the fraction of responses assigned to each category; longer responses contribute more selected token instances.

The category definitions are operational and mutually exclusive. After separating punctuation and formatting from lexical spans, we apply the following rules:
\begin{itemize}[itemsep=2pt, leftmargin=*]
\item \textbf{Answer-committing tokens} are option letters, answer numbers, or equivalent answer values that occur in an explicit answer context such as ``Answer: B'' or ``the correct option is B''. The same letter or number elsewhere in an explanation is not assigned to this category solely because of its surface form.
\item \textbf{Temporal markers} are words or multiword expressions that indicate order, duration, repetition, or change over time, including ``before'', ``after'', ``then'', ``first'', ``later'', ``again'', and ``at the end''.
\item \textbf{Event verbs and nouns} describe visually observable actions, interactions, changes, or processes, such as ``pours'', ``opens'', ``jumps'', ``collision'', and ``movement''. Auxiliary and modal verbs are excluded from this category.
\item \textbf{Objects, places, attributes} identify visible entities or their properties, including people, objects, locations, colors, shapes, counts, and visible states, such as ``kitchen'', ``container'', ``red'', and ``three''.
\item \textbf{Function words, connectives, and auxiliaries} provide grammatical or discourse structure rather than visual content, including articles, determiners, conjunctions, prepositions, pronouns, auxiliary and modal verbs, negation, and discourse connectives such as ``the'', ``is'', ``and'', ``because'', and ``therefore''.
\item \textbf{Punctuation and formatting} consist of punctuation marks, line breaks, whitespace-only pieces, and other formatting symbols without lexical content.
\item \textbf{Others} is the fallback category for proper names, generic adverbs, unmatched subwords, or tokens whose category cannot be determined reliably by the predefined rules.
\end{itemize}
When a lexical span matches more than one rule, the tagger uses the following priority: explicit answer context, temporal expression, function word or connective, event expression, visual entity or attribute, and finally \textit{Others}. BPE pieces belonging to the same decoded lexical span inherit that span's category. This priority makes the assignment deterministic and ensures that every selected token receives exactly one label.

We compare the highest- and lowest-weight token deciles from $100$ training examples at three checkpoints. Table~\ref{app:tab-token-categories} reports the category proportions after pooling the selected token instances across checkpoints.

The top decile contains $74.3\%$ tokens from the four evidence- or answer-oriented categories: temporal markers ($17.3\%$), event verbs and nouns ($20.7\%$), objects, places and attributes ($21.0\%$), and answer-committing tokens ($15.3\%$). The corresponding share in the bottom decile is $24.8\%$. Event verbs and nouns increase from $5.2\%$ in the bottom decile to $20.7\%$ in the top decile, while objects, places and attributes increase from $7.5\%$ to $21.0\%$.

The complementary pattern appears for linguistic scaffolding. Function words, connectives and auxiliaries account for $4.8\%$ of the top decile and $23.1\%$ of the bottom decile, while punctuation and formatting account for $3.1\%$ and $25.4\%$, respectively. The \textit{Others} category accounts for $17.8\%$ of the top decile and $26.7\%$ of the bottom decile.

The highest-weight decile contains more temporal, event, visual-entity, and answer tokens than the lowest-weight decile. The redistributed weight is computed from the centered evidence contrast in Equation~\ref{eq:ecr-profile}, so these proportions identify where ECR concentrates training weight.

\begin{table}[h]
\centering
\caption{Composition (\%) of tokens in the top and bottom weight deciles among positively advantaged responses. Each response token is assigned to one of seven mutually exclusive categories; tokens that do not match the predefined rules are assigned to \textit{Others}.}
\label{app:tab-token-categories}
\small
\setlength{\tabcolsep}{4pt}
\renewcommand{\arraystretch}{1.1}
\begin{tabular}{@{}
    >{\raggedright\arraybackslash}p{0.30\linewidth}
    >{\centering\arraybackslash}p{0.15\linewidth}
    >{\centering\arraybackslash}p{0.17\linewidth}
    >{\raggedright\arraybackslash}p{\dimexpr0.38\linewidth-6\tabcolsep\relax}
    @{}}
\toprule
Category & Top decile & Bottom decile & Examples \\
\midrule
Temporal markers & 17.3 & 6.2 & after, then, at the end \\
Event verbs and nouns & 20.7 & 5.2 & pours, jumps, collision \\
Objects, places, attributes & 21.0 & 7.5 & kitchen, red container, three \\
Answer-committing tokens & 15.3 & 5.9 & option letters, final numbers \\
Function words, connectives, and auxiliaries & 4.8 & 23.1 & the, is, and, therefore \\
Punctuation and formatting & 3.1 & 25.4 & commas, line breaks \\
Others & 17.8 & 26.7 & proper names, generic adverbs, unmatched subwords \\
\bottomrule
\end{tabular}
\end{table}

\subsection{Sensitivity to the SSA Weight}
\label{app:ssa-weight}

We vary the SSA weight $\lambda_{\mathrm{SSA}}$ in Equation~\ref{eq:owlet-objective} on Qwen3.5-9B in non-thinking mode with $64$ uniformly sampled student frames. The fixed teacher, privileged teacher view, ECR objective, and remaining training settings are held constant. Table~\ref{app:tab-ssa-weight} evaluates $\lambda_{\mathrm{SSA}}\in\{0, 0.01, 0.02, 0.05, 0.1\}$. A weight of $0$ removes the representation-alignment term while retaining ECR, and $0.05$ is the default setting. We report accuracy on each benchmark and the unweighted macro-average to examine both the overall sensitivity and differences across tasks.

\begin{table}[h]
\centering
\caption{Sensitivity to the SSA weight on Qwen3.5-9B in non-thinking mode with 64 uniformly sampled student frames. Scores are accuracies (\%), and Avg. is the unweighted mean over the five benchmarks. All other training settings are held constant; $\lambda_{\mathrm{SSA}}=0$ disables SSA while retaining ECR, and $0.05$ is the default weight.}
\label{app:tab-ssa-weight}
\small
\resizebox{\linewidth}{!}{
\setlength{\tabcolsep}{4pt}
\renewcommand{\arraystretch}{1.1}
\begin{tabular}{lcccccc}
\toprule
$\lambda_{\mathrm{SSA}}$ & \textbf{Video-MME} & \textbf{MVBench} & \textbf{LongVideoBench} & \textbf{Video-MME 2} & \textbf{MLVU} & \textbf{Avg.} \\
\midrule
$0$ (w/o SSA) & 69.6 & 71.1 & 64.6 & 28.4 & 59.9 & 58.7 \\
$0.01$ & 70.5 & 71.1 & 64.8 & 27.9 & 59.5 & 58.8 \\
$0.02$ & 70.3 & 71.5 & \textbf{65.3} & \textbf{28.9} & 59.9 & 59.2 \\
$0.05$ (default) & 70.7 & \textbf{72.8} & 65.2 & \textbf{28.9} & 61.6 & \textbf{59.8} \\
$0.1$ & \textbf{70.9} & 71.3 & 64.8 & 28.3 & \textbf{61.9} & 59.4 \\
\bottomrule
\end{tabular}
}
\end{table}

The macro-average accuracies are $58.7\%$, $58.8\%$, $59.2\%$, $59.8\%$, and $59.4\%$ as $\lambda_{\mathrm{SSA}}$ increases from $0$ to $0.1$. The default $0.05$ gives the highest average and improves all five benchmarks over disabling SSA, with gains of $1.1$, $1.7$, $0.6$, $0.5$, and $1.7$ points on Video-MME, MVBench, LongVideoBench, Video-MME~2, and MLVU.

The best individual scores occur at different weights: $0.02$ for LongVideoBench, $0.05$ for MVBench, $0.1$ for Video-MME and MLVU, and a tie between $0.02$ and $0.05$ for Video-MME~2. At $0.1$, the macro-average falls to $59.4\%$ as MVBench, LongVideoBench, and Video-MME~2 decline relative to the default. The default remains within $0.3$ points of the best score on every benchmark.

\subsection{Qualitative Comparison Across Post-Training Methods}
\label{app:view-case-study}

We compare the responses of the Vanilla checkpoint and the checkpoints obtained with SFT, GRPO, OPSD, and \method on two video questions. For each example, all five models receive the same video frames, question, and answer options under the ordinary student inference protocol. The comparison therefore examines how different post-training methods affect the answers produced from a common input. The privileged teacher view is used only during training; the \method checkpoint receives no additional evidence at inference.

\begin{figure}[tp]
    \centering
    \includegraphics[width=\linewidth]{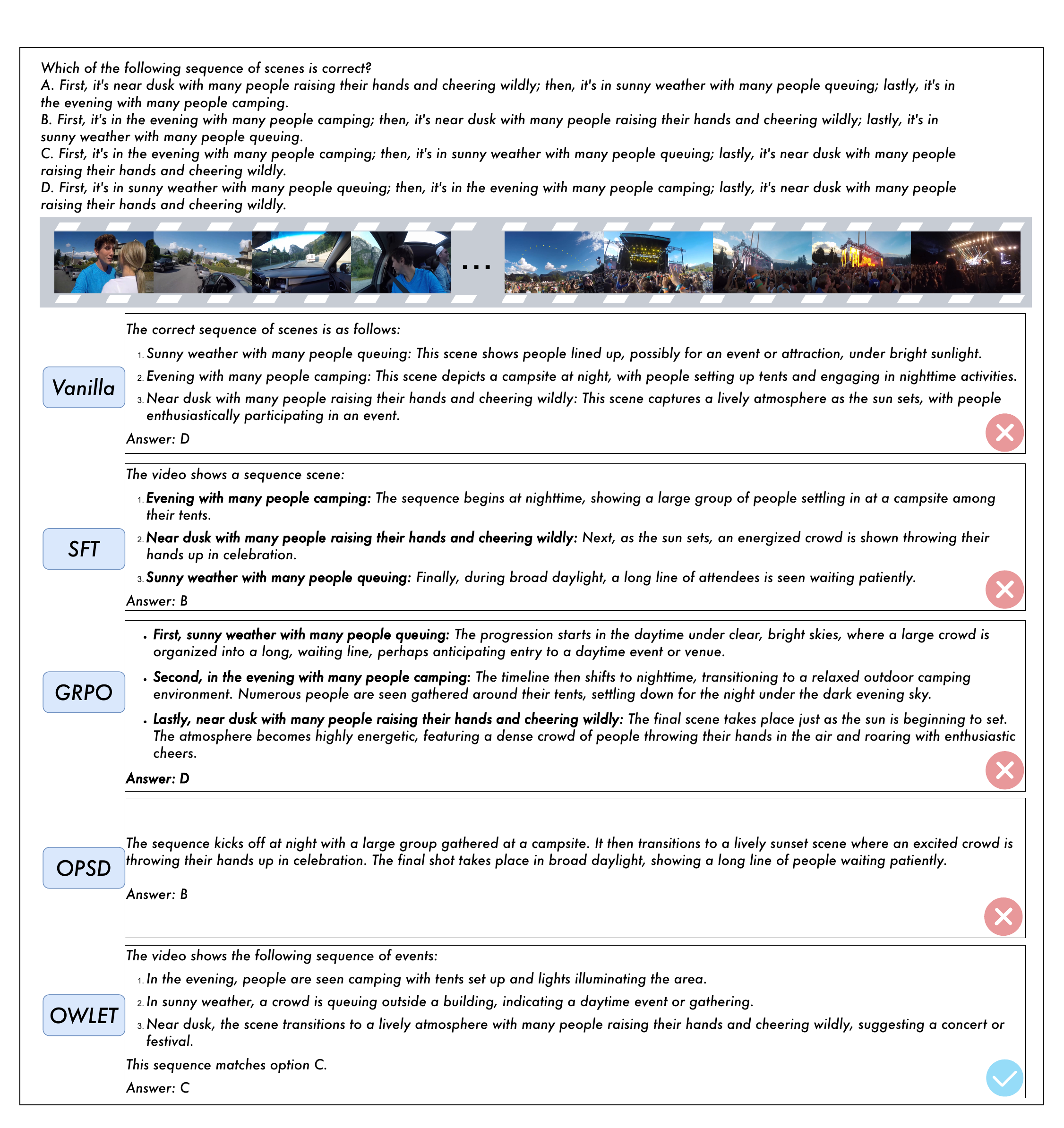}
    \caption{\textbf{Global-temporal case.} Responses from Vanilla and checkpoints trained with SFT, GRPO, OPSD, and \method to the same video input and question. Vanilla and GRPO select D, and SFT and OPSD select B; \method recovers the correct camping--queuing--cheering order and selects C.}
    \label{app:fig-view-case}
\end{figure}

\begin{figure}[tp]
    \centering
    \includegraphics[width=\linewidth]{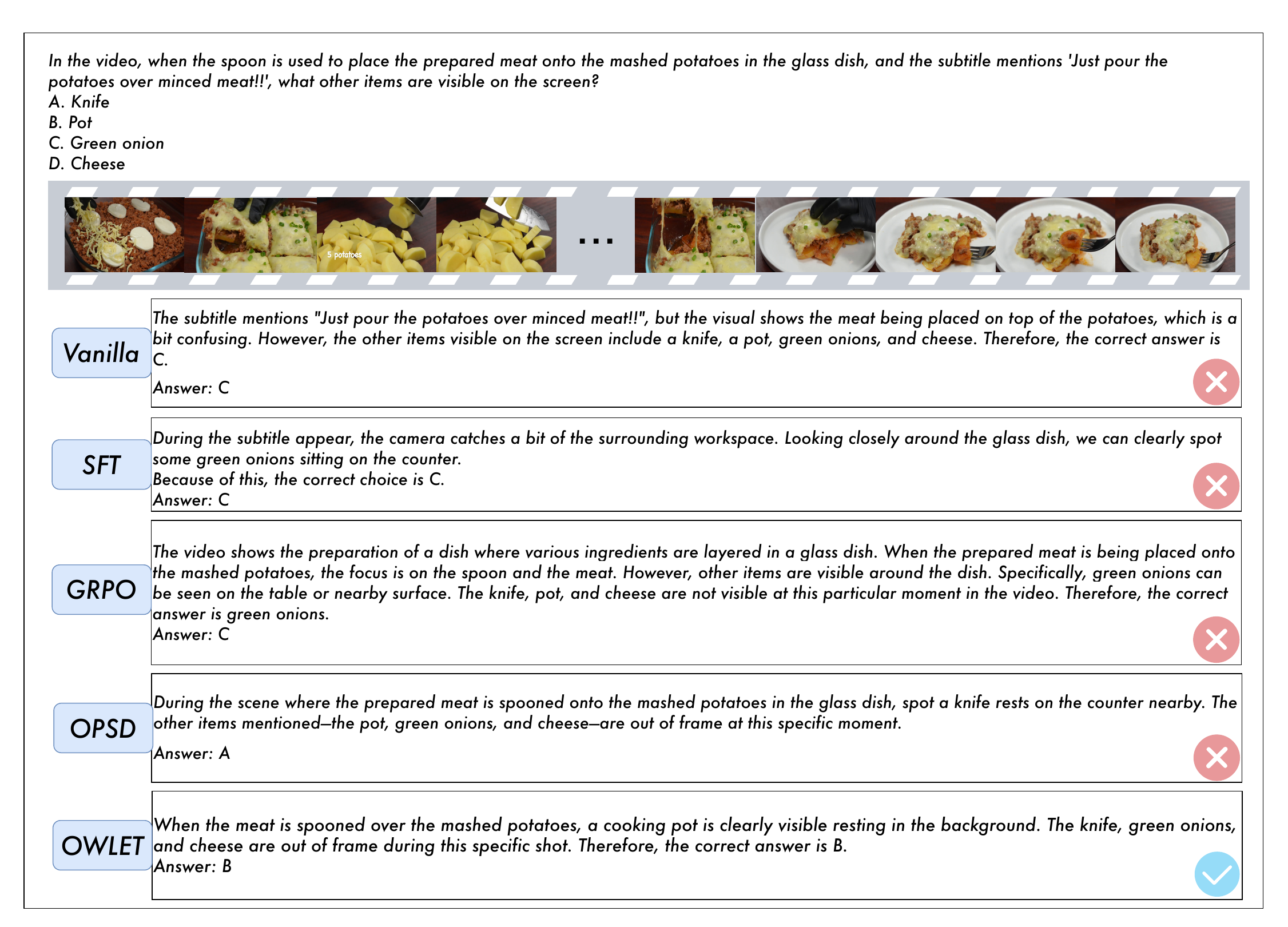}
    \caption{\textbf{Local-evidence case.} Responses to the same video input and question about the object visible at a specified cooking step. Vanilla, SFT, and GRPO select C (green onion), and OPSD selects A (knife); \method identifies the pot and selects the correct answer B.}
    \label{app:fig-local-evidence-case}
\end{figure}

The two examples cover the global-temporal and local-evidence question types discussed in Section~\ref{sec:motivation-teacher-view} and defined in Appendix~\ref{appendix:global-temporal-questions}. We select one example of each type on which \method answers correctly while Vanilla, SFT, GRPO, and OPSD answer incorrectly. Figures~\ref{app:fig-view-case} and~\ref{app:fig-local-evidence-case} show the questions, video overviews, and model responses. These selected cases show differences in the use of temporal and visual evidence; aggregate performance on the two question types is reported separately in Appendix~\ref{app:global-temporal-after}.

\paragraph{Case 1: Global temporal ordering.}
Figure~\ref{app:fig-view-case} asks which sequence of three scenes matches the video. The correct answer is C: people camping in the evening, followed by people queuing in sunny weather, and finally a crowd raising their hands and cheering near dusk. All four options contain the same three scenes and differ in their order. Answering therefore requires relating events across separated parts of the video; identifying the activities individually is insufficient to distinguish the alternatives. The lighting descriptions help identify the scenes, but the required order is their occurrence in the video.

The baseline responses exhibit two different ordering errors. Vanilla and GRPO choose D, placing the sunny queuing scene before the evening camping scene. SFT and OPSD choose B, placing the cheering scene before the queuing scene. All four responses mention the three target activities, yet arrange them into an incorrect sequence. Their explanations also elaborate on the individual scenes, describing tents, waiting attendees, or an excited crowd, without resolving the temporal relation that determines the answer. In comparison, \method gives the correct camping--queuing--cheering sequence and selects C. The example shows a global-temporal distinction: a response can describe the relevant events fluently while failing to preserve their order. The difference among the displayed answers lies in how these events are connected across time.

\paragraph{Case 2: Local evidence grounding.}
Figure~\ref{app:fig-local-evidence-case} asks which additional item is visible when prepared meat is spooned onto mashed potatoes in a glass dish, at the moment associated with the subtitle ``Just pour the potatoes over minced meat!!''. The options are a knife, a pot, green onion, and cheese, with B (pot) as the correct answer. The action and subtitle specify the local moment to inspect, while the requested object lies in the surrounding scene. Resolving this question requires associating an object with that particular moment. General knowledge of the recipe, or the presence of an ingredient elsewhere in the video, does not determine the answer.

Vanilla, SFT, and GRPO all select C (green onion), but their explanations differ. Vanilla claims that all four candidate items are visible and then chooses green onion, without explaining why it is the appropriate choice. SFT describes green onions on the counter, and GRPO makes the same identification while explicitly excluding the pot. OPSD instead chooses A (knife), claiming that the knife is visible nearby and that the remaining items are outside the frame. Thus, several responses refer directly to the requested cooking step and provide specific spatial descriptions, but still identify the wrong object. \method identifies a cooking pot in the background and selects B. The case shows the need to ground an object description in the question-specified temporal context.

\paragraph{Relation to evidence-guided post-training.}
The two cases show complementary evidence requirements: ordering separated events and identifying a detail at a specific moment. \method answers both correctly under the same student inputs used by the baselines.

\subsection{Training-Time Comparison}
\label{app:training-cost}

Table~\ref{app:tab-training-cost} compares the training cost of the three on-policy methods, GRPO, OPSD, and \method, on Qwen3.5-9B in non-thinking mode with $64$ uniformly sampled student frames. All three methods use eight H200 GPUs and a budget of $1{,}000$ updates. We report the average time per step for Rollout and Update, their sum, and the accumulated step time over this training budget.

\begin{table}[h]
\centering
\caption{On-policy training cost on eight H200 GPUs for Qwen3.5-9B at 64 student frames, averaged over the first $1{,}000$ updates. Update covers the non-rollout portion of each timed step. Total (s/step) is the sum of Rollout and Update; Total (h) gives the accumulated step time.}
\label{app:tab-training-cost}
\resizebox{0.8\linewidth}{!}{
\begin{tabular}{lcccc}
\toprule
Method & Rollout (s/step) & Update (s/step) & Total (s/step) & Total (h) \\
\midrule
GRPO & 406.99 & 507.30 & 914.29 & 254.0 \\
OPSD & 410.89 & 437.11 & 848.00 & 235.6 \\
\method & 434.31 & 514.24 & 948.55 & 263.5 \\
\bottomrule
\end{tabular}
}
\end{table}

\paragraph{Overall training cost.}
OPSD has the lowest measured cost at $848.00$ seconds per step, followed by GRPO at $914.29$ seconds and \method at $948.55$ seconds. Over $1{,}000$ updates, these correspond to $235.6$, $254.0$, and $263.5$ hours of accumulated step time, respectively. Relative to GRPO, \method adds $3.7\%$ (approximately $9.5$ hours); relative to OPSD, it adds $11.9\%$ (approximately $27.9$ hours).

\paragraph{Rollout and update costs.}
Relative to GRPO, \method spends $27.32$ more seconds per step on rollout and $6.94$ more seconds on update. Relative to OPSD, the rollout difference is $23.42$ seconds and the update difference is $77.13$ seconds.

\paragraph{Training and inference implications.}
The comparison quantifies the cost of running each method for the same number of updates. For \method, the fixed teacher supplies token scores and hidden-state targets only during training. At inference, the student uses the standard uniformly sampled view without invoking the teacher or constructing the privileged teacher view.

\subsection{Response Length during On-Policy Training}
\label{app:response-length}

\begin{figure}[t]
\centering
\includegraphics[width=\linewidth]{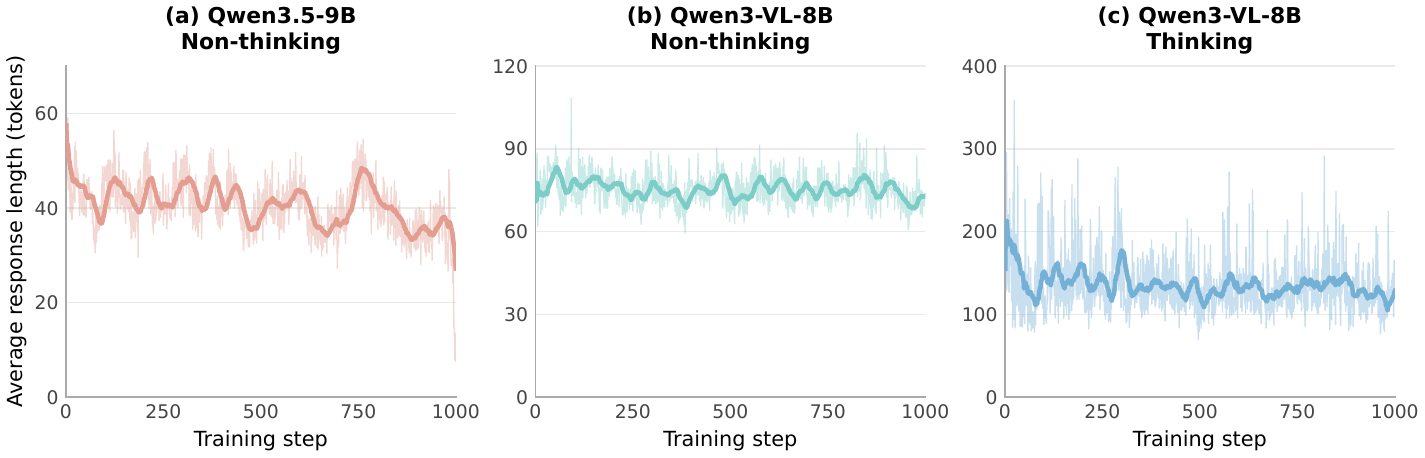}
\caption{Response length during the first $1{,}000$ training steps of \method with 64 student frames: (a) Qwen3.5-9B in non-thinking mode, (b) Qwen3-VL-8B-Instruct in non-thinking mode, and (c) Qwen3-VL-8B-Thinking. Thin, light curves show the mean number of non-padding generated tokens across all 128 responses at each step; thick curves show 25-step trailing moving averages, using the available history for the first 24 steps. Thinking lengths include the generated reasoning trace and final answer. The panels use separate vertical scales, and all per-step means are shown.}
\label{app:fig-response-length}
\end{figure}

We compare response-length dynamics across three \method runs: Qwen3.5-9B in non-thinking mode, Qwen3-VL-8B-Instruct in non-thinking mode, and Qwen3-VL-8B-Thinking. Each run uses $64$ uniformly sampled student frames, $16$ prompts per step, and $G=8$ responses per prompt. We analyze the $1{,}000$ training steps of each run, covering $128{,}000$ responses per run and $384{,}000$ responses in total. All generated responses are included, irrespective of verifier correctness. Every analyzed step is matched to a rollout file containing all $128$ decoded responses. For the thinking model, length includes the generated reasoning trace as well as the final answer. Figure~\ref{app:fig-response-length} shows the trajectories, and Table~\ref{app:tab-response-length-summary} summarizes their scales and changes.

\begin{table}[t]
\centering
\caption{Mean response length in generated tokens for the three runs in Figure~\ref{app:fig-response-length}. Overall averages all $1{,}000$ steps. Change is the percentage difference between the final and first 100-step means, computed before rounding.}
\label{app:tab-response-length-summary}
\small
\setlength{\tabcolsep}{5pt}
\begin{tabular}{llrrrr}
\toprule
Model & Mode & Steps 1--100 & Steps 901--1,000 & Overall & Change (\%) \\
\midrule
Qwen3.5-9B & Non-thinking & 42.4 & 33.9 & 40.4 & $-20.0$ \\
Qwen3-VL-8B & Non-thinking & 77.5 & 72.0 & 75.2 & $-7.1$ \\
Qwen3-VL-8B & Thinking & 145.6 & 123.7 & 135.2 & $-15.0$ \\
\bottomrule
\end{tabular}
\end{table}

\paragraph{Qwen3.5-9B: late shortening and direct answers.}
The Qwen3.5-9B run shows an early reduction followed by intermediate fluctuations and a sharper decrease near the end. Steps $1$--$50$ and $51$--$100$ average $44.7$ and $40.0$ tokens, respectively. The final ten steps average $17.2$ tokens, with step $1{,}000$ averaging $11.0$. The decoded rollouts connect this late shortening to a change in response format. We count a response as answer-only when its entire decoded text consists of \texttt{Answer: X}, where \texttt{X} is a single uppercase option letter, allowing whitespace and an optional terminal full stop. Such responses account for $22$ of $6{,}400$ rollouts ($0.34\%$) over steps $901$--$950$, but $821$ of $1{,}280$ rollouts ($64.1\%$) over the final ten steps. Thus, the late decrease coincides with a marked shift toward direct answer-only outputs.

\paragraph{Qwen3-VL-8B-Instruct: relatively stable response length.}
The non-thinking Qwen3-VL run maintains a more consistent response length, averaging $75.2$ tokens overall. Its consecutive $100$-step means range from $72.0$ to $77.5$ tokens, with the final block $7.1\%$ shorter than the first. The final ten steps average $70.9$ tokens, and no decoded response matches the answer-only pattern defined above.

\paragraph{Qwen3-VL-8B-Thinking: early compression with continued reasoning output.}
The thinking run produces longer responses, averaging $135.2$ generated tokens including the reasoning trace and final answer. Its mean falls from $157.0$ tokens over steps $1$--$50$ to $134.1$ over steps $51$--$100$, followed by a rebound to $146.5$ over steps $101$--$200$. Thereafter, the trajectory continues to fluctuate: the $100$-step averages over steps $401$--$1{,}000$ lie between $123.7$ and $136.5$ tokens. The final $100$ steps are $15.0\%$ shorter than the first $100$, but the final ten steps still average $133.8$ tokens. This run shows early shortening followed by fluctuations at a lower typical length, and differs from the sharp late decrease observed for Qwen3.5-9B.

\paragraph{Joint interpretation.}
Across the three runs, mean response length decreases between the first and final 100 steps. The trajectories differ: Qwen3-VL-Instruct remains relatively stable, Qwen3-VL-Thinking shortens early, and Qwen3.5-9B shifts toward answer-only responses near the end of training.

\subsection{Global-Temporal Accuracy after Post-Training}
\label{app:global-temporal-after}

Section~\ref{sec:motivation-teacher-view} shows that restricting the input to annotated evidence favors questions that can be answered from local moments, while broader temporal questions require additional context. We use the same diagnostic split to examine this trade-off after post-training. The Vanilla and Key Windows rows are input-view controls from the common initial Qwen3.5-9B checkpoint. GRPO, OPSD, and \method are evaluated at their final checkpoints with uniformly sampled student frames. We report overall accuracy together with the held-out global-temporal and local-evidence subsets defined in Appendix~\ref{appendix:global-temporal-questions}.

The two input-view controls reproduce the expected specialization trade-off. Vanilla obtains $50.5\%$ overall, with $51.8\%$ on global-temporal questions and $49.2\%$ on local-evidence questions. Key Windows increases the overall score to $57.8\%$ and the local-evidence score to $65.4\%$, but its global-temporal score is only $50.2\%$. Concentrating all frames around question-relevant moments is therefore effective for the local subset, yet it does not preserve the broader temporal information needed by the global subset.

 As shown in Table~\ref{app:tab-global-temporal-after}, all trained baselines improve over the Vanilla model on both subsets. GRPO reaches $55.8\%$ overall, $53.4\%$ on global-temporal questions, and $58.2\%$ on local-evidence questions. OPSD reaches $54.5\%$, $52.4\%$, and $56.6\%$, respectively. \method reaches $59.2\%$ overall, with $58.6\%$ global-temporal and $59.8\%$ local-evidence accuracy. Relative to GRPO, \method gains $3.4$ points overall, $5.2$ points on global-temporal questions, and $1.6$ points on local-evidence questions. Relative to OPSD, the corresponding gains are $4.7$, $6.2$, and $3.2$ points. The larger margins on the global-temporal subset indicate that the improvement is not limited to resolving the localized event; \method also retains the broader temporal information that the student needs under uniform sampling.

\method also provides a more balanced result than the Key Windows control. Its overall accuracy is $1.4$ points higher, and its global-temporal accuracy is $8.4$ points higher, while Key Windows remains $5.6$ points higher on local-evidence questions. Thus, the strongest local-evidence score in this table does not translate into the strongest overall score: the two subsets are equally represented in the diagnostic split, so the global-temporal improvement substantially affects the final average.

\begin{table}[h]
\centering
\caption{Accuracy (\%) on the held-out diagnostic split before and after post-training, overall and by question type, for Qwen3.5-9B at 64 input frames. The Vanilla and Key Windows rows use the same initial checkpoint with uniformly sampled and question-conditioned Key Windows views, respectively; GRPO, OPSD, and \method are evaluated at their final checkpoints under the uniformly sampled student view.}
\label{app:tab-global-temporal-after}
\small
\begin{tabular}{lccc}
\toprule
Method & Overall & Global-temporal & Local-evidence \\
\midrule
Vanilla & 50.5 & 51.8 & 49.2 \\
Key Windows & 57.8 & 50.2 & \textbf{65.4} \\
GRPO & 55.8 & 53.4 & 58.2 \\
OPSD & 54.5 & 52.4 & 56.6 \\
\green \method &\green  \textbf{59.2} &\green  \textbf{58.6} &\green  59.8 \\
\bottomrule
\end{tabular}
\end{table}

\subsection{Extended Training Schedules}
\label{app:longer-training}

The main experiments train for $1{,}000$ steps. We continue the Qwen3.5-9B run at $64$ input frames with the same training data, optimizer and learning rate, rollout budget of $16$ prompts with $G=8$ responses per step, clipping range, rule-based verifier, global-plus-local teacher view, and ECR and SSA settings. The teacher stays fixed at the initial checkpoint. We save checkpoints at $1{,}000$, $1{,}200$, $1{,}500$, $1{,}800$, $2{,}000$, $2{,}200$, $2{,}500$, $2{,}800$, and $3{,}000$ steps, and evaluate each with the main-text protocol on Video-MME, MVBench, LongVideoBench, Video-MME~2, and MLVU. The schedule ends at $3{,}000$ steps because of the available compute budget.

Table~\ref{app:tab-longer-training} reports the results. Macro-average accuracy increases from $59.8$ at $1{,}000$ steps to $61.3$ at $3{,}000$ steps, with the highest observed score of $61.5$ at $2{,}800$ steps. Between the initial and final checkpoints, LongVideoBench gains $2.3$ points, Video-MME $2.2$, MLVU $1.7$, Video-MME~2 $0.7$, and MVBench $0.6$; the five benchmarks reach their highest observed scores at different checkpoints. The mean absolute change between adjacent checkpoints is $0.30$ points for the macro-average and ranges from $0.41$ to $0.67$ points across benchmark columns.

The additional $2{,}000$ updates yield a $1.5$-point macro-average gain. Using the per-step cost measured in Appendix~\ref{app:training-cost}, they correspond to approximately $527$ hours on eight H200 GPUs. The main comparison uses a common $1{,}000$-update budget for GRPO, OPSD, and \method, and the main-text results continue to use the fixed $1{,}000$-step checkpoint for every backbone and frame budget.

\begin{table}[h]
\centering
\caption{Per-benchmark accuracy (\%) of \method checkpoints under an extended training schedule, for Qwen3.5-9B at 64 input frames. Checkpoints are saved at nine steps between the main-text budget of $1{,}000$ and $3{,}000$, spaced $200$--$300$ steps apart, and each is evaluated with the protocol of Table~\ref{tab:main} on the same five benchmarks; the $1{,}000$-step row reproduces the corresponding \method entry there. \textbf{Avg.} is the macro-average over the five benchmarks, computed before rounding. The last row gives the change from $1{,}000$ to $3{,}000$ steps.}
\label{app:tab-longer-training}
\resizebox{\textwidth}{!}{
\setlength{\tabcolsep}{4pt}
\renewcommand{\arraystretch}{1.1}
\begin{tabular}{lcccccc}
\toprule
\textbf{Training steps} & \textbf{Video-MME} & \textbf{MVBench} & \textbf{LongVideoBench} & \textbf{Video-MME 2} & \textbf{MLVU} & \textbf{Avg.} \\
\midrule
1,000 (main text) & 70.7 & 72.8 & 65.2 & 28.9 & 61.6 & 59.8 \\
1,200 & 70.5 & 73.1 & 65.5 & 28.4 & 60.8 & 59.7 \\
1,500 & 71.0 & 73.6 & 65.5 & 28.6 & 60.5 & 59.8 \\
1,800 & 72.3 & 73.0 & 64.8 & 29.5 & 61.0 & 60.1 \\
2,000 & 72.0 & 73.6 & 66.1 & 29.2 & 62.7 & 60.7 \\
2,200 & 70.9 & 73.3 & 66.9 & 29.1 & 62.9 & 60.6 \\
2,500 & 71.5 & 73.0 & 67.2 & 29.5 & 62.5 & 60.7 \\
2,800 & 72.8 & 73.9 & 67.7 & 30.0 & 63.1 & 61.5 \\
3,000 & 72.9 & 73.4 & 67.5 & 29.6 & 63.3 & 61.3 \\
\midrule
$\Delta$ ($3{,}000$ vs.\ $1{,}000$) & $+2.2$ & $+0.6$ & $+2.3$ & $+0.7$ & $+1.7$ & $+1.5$ \\
\bottomrule
\end{tabular}
}
\end{table}

\end{document}